\documentclass[runningheads]{llncs}

\usepackage{eccv}

\usepackage{eccvabbrv}

\usepackage{graphicx}
\usepackage{booktabs}

\usepackage[accsupp]{axessibility}  

\usepackage[pagebackref,breaklinks,colorlinks,citecolor=eccvblue]{hyperref}

\usepackage{orcidlink}
\definecolor{cvprblue}{rgb}{0.21,0.49,0.74}
\usepackage{mathabx}
\usepackage{multirow}
\usepackage{pifont}
\usepackage{subfloat}
\usepackage{enumitem}
\usepackage{caption}
\usepackage{graphicx}
\usepackage{xspace} 
\usepackage{xcolor,colortbl}
\usepackage{mathtools}
\usepackage{cite}
\usepackage{tikz}
\usepackage{lmodern}
\usepackage{arydshln}

\DeclareCaptionLabelFormat{andtable}{#1~#2  \&  \tablename~\thetable}
\newcommand{\Paragraph}[1]{\vspace{0.6mm} \noindent \textbf{#1}\hspace{0mm}}

\newcommand{\inv}{^{\text{-}1}}

\definecolor{app_color}{RGB}{240, 94, 35}
\definecolor{hough_red}{RGB}{191,30,45}
\definecolor{hough_green}{RGB}{5,104,57}
\definecolor{hough_matrix_orange}{RGB}{245,238,48}
\definecolor{hough_matrix_blue}{RGB}{113,155,210}

\title{Revisit Self-supervised Depth Estimation with Local Structure-from-Motion}

\author{Shengjie Zhu  \and Xiaoming Liu}

\institute{
Department of Computer Science and Engineering,\\
Michigan State University, East Lansing, MI, 48824 \\
\email{zhusheng@msu.edu}, \email{liuxm@cse.msu.edu}
}

\begin{document}
\setcounter{secnumdepth}{4}
\maketitle
\begin{abstract}
Both self-supervised depth estimation and Structure-from-Motion (SfM) recover scene depth from RGB videos.
Despite sharing a similar objective, the two approaches are disconnected.
Prior works of self-supervision backpropagate losses defined within immediate neighboring frames.
Instead of learning-through-loss, this work proposes an alternative scheme by performing local SfM.
First, with calibrated RGB or RGB-D images, we employ a depth and correspondence estimator to infer depthmaps and pair-wise correspondence maps.
Then, a novel bundle-RANSAC-adjustment algorithm jointly optimizes camera poses and one depth adjustment for each depthmap.
Finally, we fix camera poses and employ a NeRF, however, without a neural network, for dense triangulation and geometric verification.
Poses, depth adjustments, and triangulated sparse depths are our outputs.
For the first time, we show self-supervision within $5$ frames already benefits SoTA supervised depth and correspondence models.
Despite self-supervision, our pose algorithm has certified global optimality, outperforming optimization-based, learning-based, and NeRF-based prior arts.
The project page is held in the \href{https://shngjz.github.io/SSfM.github.io/}{link}.

\end{abstract}
    
\section{Introduction}
\label{sec:intro}

Monocular depth estimation~\cite{fu2018deep, lee2019big} infers depthmap from a single image.
It is an essential vision task with applications in AR/VR~\cite{niu2021making}, autonomous driving~\cite{geiger2012we}, and 3D reconstruction~\cite{butime20063d}.
Most methods~\cite{yuan2022new, bhat2023zoedepth, ranftl2020towards, piccinelli2023idisc} supervise the model with groundtruth collected from stereo cameras~\cite{zhang2012microsoft} or LiDAR~\cite{geiger2012we}.
Recently, self-supervised depth~\cite{gordon2019depth, godard2019digging, zhu2020edge} has drawn significant attention due to its potential to scale up depth learning from massive unlabeled RGB videos.

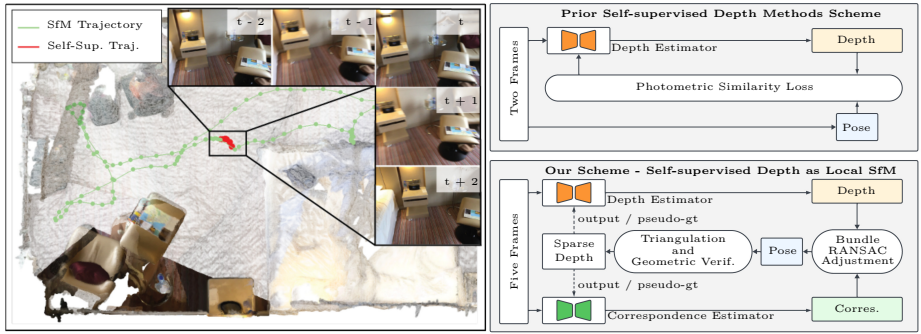
\begin{figure}[t!]
  \captionsetup{font=small}
  \centering
  \begin{tikzpicture}
  \draw (0, 0) node[inner sep=0] {\includegraphics[width=\linewidth]{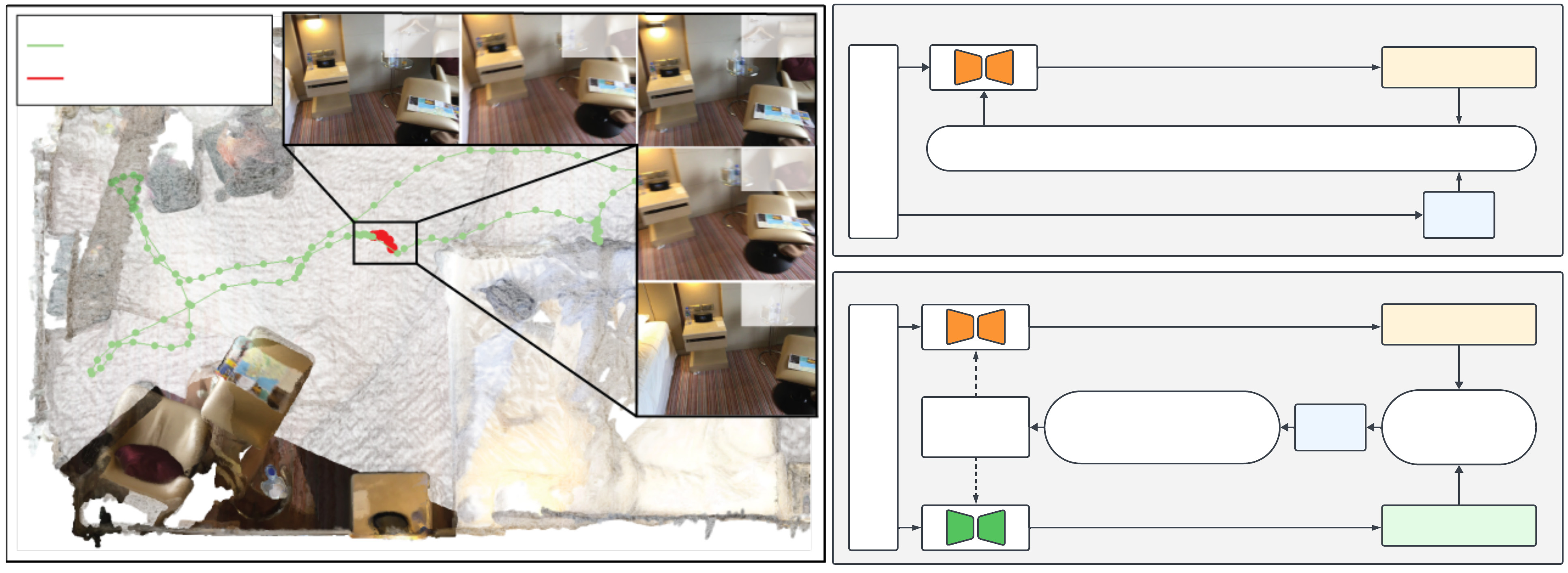}};
  \draw (-4.45-0.55+0.13, 1.8+0.1) node[inner sep=0] {\fontsize{4.0}{10}\selectfont SfM Trajectory};
  \draw (-4.29-0.55-0.05, 1.55+0.08) node[inner sep=0] {\fontsize{4.0}{10}\selectfont Self-Sup. Traj.};
  \draw (3.5, 1.08) node[inner sep=0] {\fontsize{4.0}{10}\selectfont Photometric Similarity Loss};
  \draw (5.25, 1.7) node[inner sep=0] {\fontsize{4.0}{10}\selectfont Depth};
  \draw (2.7, 1.6) node[inner sep=0] {\fontsize{4.0}{10}\selectfont Depth Estimator};
  \draw (5.25, 0.55) node[inner sep=0] {\fontsize{4.0}{10}\selectfont Pose};
  \draw (5.25, -0.3) node[inner sep=0] {\fontsize{4.0}{10}\selectfont Depth};
  \draw (5.26, -1.86) node[inner sep=0] {\fontsize{4.0}{10}\selectfont Corres.};
  \draw (4.87+0.4, -1.0+0.01+0.05) node[inner sep=0] {\fontsize{4.0}{10}\selectfont Bundle};
  \draw (4.87+0.4, -1.1+0.01) node[inner sep=0] {\fontsize{4.0}{10}\selectfont RANSAC};
  \draw (4.87+0.4, -1.2+0.01-0.05) node[inner sep=0] {\fontsize{4.0}{10}\selectfont Adjustment};
  \draw (4.27, -1.1+0.01) node[inner sep=0] {\fontsize{4.0}{10}\selectfont Pose};
  \draw (3.0-0.05, -1.1+0.11+0.05) node[inner sep=0] {\fontsize{4.0}{10}\selectfont Triangulation};
  \draw (3.0-0.05, -1.2+0.11) node[inner sep=0] {\fontsize{4.0}{10}\selectfont and};
  \draw (3.0-0.05, -1.3+0.11-0.05) node[inner sep=0] {\fontsize{4.0}{10}\selectfont Geometric Verif.};
  \draw (1.5, -1.2+0.15 + 0.04) node[inner sep=0] {\fontsize{4.0}{10}\selectfont Sparse};
  \draw (1.5, -1.3+0.15 - 0.04) node[inner sep=0] {\fontsize{4.0}{10}\selectfont Depth};
  \draw (2.64, -0.42) node[inner sep=0] {\fontsize{4.0}{10}\selectfont Depth Estimator};
  \draw (3.05, -1.97) node[inner sep=0] {\fontsize{4.0}{10}\selectfont Correspondence Estimator};
  \draw (0.7, -1.1) node[inner sep=0] {\fontsize{4.0}{10}\selectfont \rotatebox{90}{Five Frames}};
  \draw (0.7, 1.2) node[inner sep=0] {\fontsize{4.0}{10}\selectfont \rotatebox{90}{Two Frames}};
  \draw (0.0, 1.95) node[inner sep=0] {\fontsize{4.0}{10}\selectfont t};
  \draw (-1.35, 1.95) node[inner sep=0] {\fontsize{4.0}{10}\selectfont t - 1};
  \draw (-2.8, 1.95) node[inner sep=0] {\fontsize{4.0}{10}\selectfont t - 2};
  \draw (0.0, 0.9) node[inner sep=0] {\fontsize{4.0}{10}\selectfont t + 1};
  \draw (0.0, -0.18) node[inner sep=0] {\fontsize{4.0}{10}\selectfont t + 2};
  \draw (2.35, -0.68) node[inner sep=0] {\fontsize{4.0}{10}\selectfont output / pseudo-gt};
  \draw (2.35, -1.55) node[inner sep=0] {\fontsize{4.0}{10}\selectfont output / pseudo-gt};
  \draw (3.45, -0.03) node[inner sep=0] {\fontsize{4.0}{10}\selectfont {\textbf{Our Scheme - Self-supervised Depth as Local SfM}}};
  \draw (3.45, 2.05) node[inner sep=0] {\fontsize{4.0}{10}\selectfont {\textbf{Prior Self-supervised Depth Methods Scheme}}};
  \end{tikzpicture}
  \vspace{-4mm}
  \caption{
  \small 
  \textbf{Revisit Self-supervision with Local SfM.}
  The work proposes alternating the learning-through-loss with a local SfM pipeline for self-supervised depth estimation.
  We summarize our differences.
  On self-supervision:
  (1) Instead of using naive two-view camera poses, we propose a Bundle-RANSAC-Adjustment pose optimization algorithm with multi-view constraints.
  (2) Instead of backpropagating through a loss, we produce a sparse point cloud with explicit triangulation and geometric verification.
  The point cloud serves as either output or pseudo-groundtruth for self-supervision.
  On SfM:
  (1) Our local SfM is adapted to use estimated monocular depthmaps and automatically resolve their scale inconsistency between pairs of images.
  (2) We maintain accuracy under significant sparse view variations, \textit{e.g.}, red trajectories.
  We generalize SfM to as few as $5$ frames, similar to the number of images used to define self-supervision loss.
  \vspace{-4mm}
  \label{fig:teaser}
  }
\end{figure}

Classic SfM methods~\cite{schonberger2016structure, snavely2006photo, sarlin2023pixel, wu2013towards, agarwal2011building, crandall2011discrete} also reconstruct scene depth from unlabled RGB videos.
Despite its relevance, SfM is rarely applied to self-supervised depth learning.
We outline two potential reasons.
First, SfM is an off-the-shelf algorithm unrelated to the depth estimator.
Scale ambiguity renders SfM poses and depths at different scales compared to depth models.
Second, self-supervision has a well-defined training scheme to work with universal unlabeled videos.
It backpropagates through photometric loss computed within immediate neighboring frames, \textit{e.g.},  {red} trajectory in \cref{fig:teaser}.
In contrast, SfM is more selective to input videos. 
It requires images of diverse view variations (green trajectory in \cref{fig:teaser}), being inaccurate and unstable when applied to a small frame window.


\begin{figure*}[t!]
  \captionsetup{font=small}
  \centering
  \begin{tikzpicture}
  \draw (0, 0) node[inner sep=0] {\includegraphics[width=\linewidth]{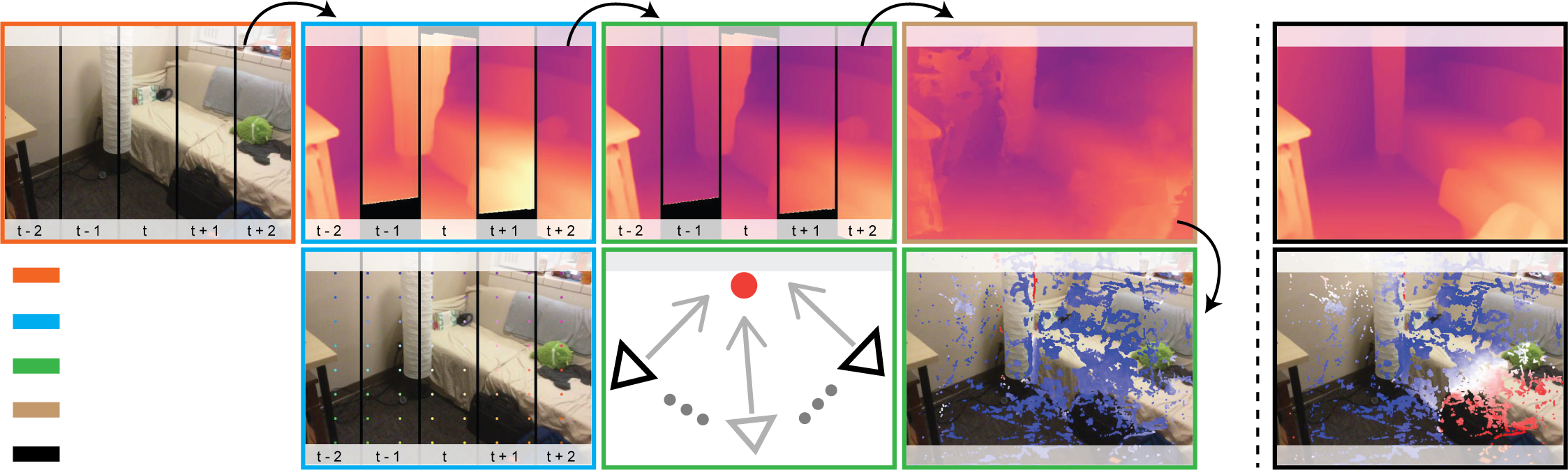}};
  \draw (-5.08, 1.55) node[inner sep=0] {\fontsize{3.5}{10}\selectfont RGB Neighbouring Frames};
  \draw (-2.8, 1.55) node[inner sep=0] {\fontsize{3.5}{10}\selectfont N Monocular Depthmaps};
  \draw (-2.75, -0.22) node[inner sep=0] {\fontsize{3.5}{10}\selectfont N x (N-$1$) Correspondence};
  \draw (-0.50, 1.55) node[inner sep=0] {\fontsize{3.5}{10}\selectfont N-$1$ Depth Adjustments};
  \draw (-0.68, -0.22) node[inner sep=0] {\fontsize{3.5}{10}\selectfont N-$1$ Camera Poses};
  \draw (1.85, 1.55) node[inner sep=0] {\fontsize{3.5}{10}\selectfont RF Intermediate Depth};
  \draw (1.97, -0.22) node[inner sep=0] {\fontsize{3.5}{10}\selectfont Triangulated Sparse Depth};
  \draw (4.91, -0.22) node[inner sep=0] {\fontsize{3.5}{10}\selectfont  Monocular~\cite{bhat2023zoedepth} Sparse Depth};
  \draw (4.85, 1.55) node[inner sep=0] {\fontsize{3.5}{10}\selectfont Moncular~\cite{bhat2023zoedepth} Input Depth};
  \draw (4.85, -1.72) node[inner sep=0] {\fontsize{3.5}{10}\selectfont Accuracy ($\delta$ < $0.5$): $\boldsymbol{82.4\%}$};
  \draw (1.95, -1.72) node[inner sep=0] {\fontsize{3.5}{10}\selectfont Accuracy ($\delta$ < $0.5$): $\boldsymbol{99.5\%}$};
  \draw (-4.9, -0.32) node[inner sep=0] {\fontsize{3.5}{10}\selectfont Input RGB Frames};
  \draw (-4.96, -0.68) node[inner sep=0] {\fontsize{3.5}{10}\selectfont Input Estimation};
  \draw (-5.31, -1.02) node[inner sep=0] {\fontsize{3.5}{10}\selectfont Outputs};
  \draw (-4.71, -1.35) node[inner sep=0] {\fontsize{3.5}{10}\selectfont Intermediate Estimation};
  \draw (-5.02, -1.71) node[inner sep=0] {\fontsize{3.5}{10}\selectfont For Comparison};
  \end{tikzpicture}
    \vspace{-4mm}
    \caption{\small 
    \textbf{Local Structure-from-Motion.}
    With $N$ neighboring frames, we extract monocular depthmaps and pairwise dense correspondence maps with methods, \textit{e.g.}, ZoeDepth~\cite{bhat2023zoedepth} and PDC-Net~\cite{truong2023pdc}.
    Next, skipping the root frame, we optimize the rest $N-1$ camera poses and depth adjustments.
    The depth adjustments render input depthmaps \textbf{temporally consistent}.
    Fixing poses and adjustments, we use the Radiance Field (RF) for triangulation and output a geometrically verified sparse root depthmap.
    Our local SfM applies \textbf{self-supervision} with only $5$ RGB frames.
    Yet, our sparse output already outperforms the input supervised depth with SoTA performance.
    }
    \vspace{-2mm}
    \label{fig:triang_depth}
\end{figure*}
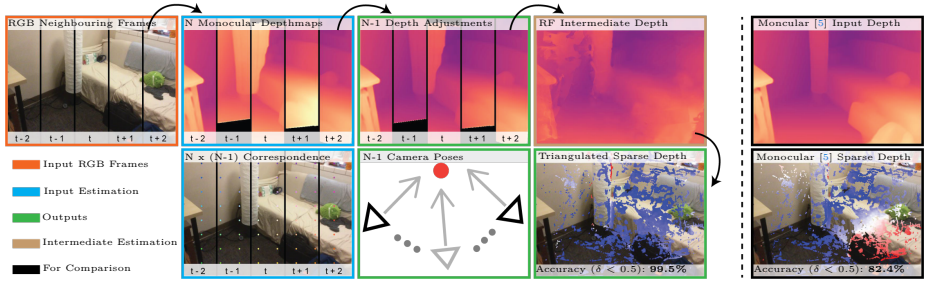

This work connects self-supervision with SfM.
We replace the self-supervision loss with a complete SfM pipeline that maintains robustness to a local window.
Shown in \cref{fig:triang_depth}, with $N$ frames as input, our algorithm outputs $N-1$ camera poses, $N-1$ depth adjustments, and the sparse triangulated point cloud.
In initialization, $N$ monocular depthmaps and $N\times (N-1)$ pairwise correspondence maps are inferred.
Next, we propose a Bundle-RANSAC-Adjustment pose estimation algorithm that retains accuracy for second-long videos.
The algorithm utilizes the 3D priors from mono-depthmap to compensate for the deficient camera views.
Correspondingly, we optimize $N-1$ depth adjustments to alleviate the depth scale ambiguity by temporally aligning to the root frame depth.

The Bundle-RANSAC-Adjustment extends two-view RANSAC 
with multi-view bundle-adjustment (BA). 
The algorithm has quadratic complexity,  designed for parallel GPU computation.
We \underline{RAN}domly \underline{SA}mple and hypothesize a set of normalized poses.
In \underline{C}onsensus checking, we apply BA to evaluate a robust inlier-counting scoring function over multi-view images.
Camera scales and depth adjustments are determined during BA to maximize the scoring function.


Next, we freeze the optimized poses and employ a Radiance Field (RF), \textit{i.e.}, a NeRFF~\cite{mildenhall2021nerf} without a neural network, for triangulation. 
We optimize RF to achieve multi-view depthmap and correspondence consistency within a shared 3D frustum volume.
For outputs, we apply geometric verification to extract multi-view consistent point cloud, \textit{i.e.}, a sparse root depthmap.

\cref{fig:teaser} contrasts our method with prior self-supervised depth and SfM methods.
To our best knowledge, there has not been prior work showing geometry-based self-supervised depth benefits supervised models.
However, self-supervision is supposed to augment supervised models with unlabeled data.
In \cref{fig:triang_depth}, our unique pipeline gives the \textbf{first} evident results, that self-supervision with \textbf{as few as} $5$ frames already benefits supervised models.

On top of depths, our multi-view RANSAC pose has {certified global optimality} under a robust scoring function.
It outperforms prior arts in optimization-based~\cite{schonberger2016structure, zhu2023lighteddepth}, learning-based~\cite{Teed2020DeepV2D, wang2023dust3r}, and NeRF-based~\cite{truong2023sparf} pose algorithms.


Beyond pose and depth,
our method has diverse applications.
The depth adjustments from our method provide empirically consistent depthmaps, important for AR image compositing.
When given RGB-D inputs, our method enables self-supervised correspondence estimation.
Our accurate pose estimation gives improved projective correspondence than the SoTA supervised correspondence input.
An example is in \cref{fig:projective_corres}.
We summarize our contributions as:
\begin{itemize}
    \item We propose a novel local SfM algorithm with Bundle-RANSAC-Adjustment.
    \item We show the \textbf{first} evident result that self-supervised depth with \textbf{as few as} $5$ frames already benefit SoTA supervised models.
    \item We achieve SoTA sparse-view pose estimation performance.
    \item We enable self-supervised temporally consistent depthmaps.
    \item We enable self-supervised correspondence estimation with $5$ RGB-D frames.
\end{itemize}

\section{Related Works}

\Paragraph{Structure-from-Motion.}
SfM is a comprehensive task~\cite{schonberger2016structure, wu2013towards}.
A typical pipeline is, correspondence extraction~\cite{lowe2004distinctive, tuytelaars2008local, brown2010discriminative}, two-view initialization~\cite{beder2006determining, li2006five, lepetit2009ep}, triangulation~\cite{li2007practical, olsson2010outlier}, and local \& global bundle-adjustment~\cite{schonberger2016structure, wu2013towards}.
Classic methods require diverse view variations for accurate reconstruction.
Our method compensates SfM on scarse camera views via introducing deep depth estimator.
Further, we suggest SfM itself is a self-supervised learning pipeline, as in \cref{fig:teaser}.
Finally, our SfM is not up-to-scale and shares the metric space as the input depthmap.


\Paragraph{Sparse Multi-view Pose Estimation.}
Estimating poses from sparse frames is  crucial for self-supervision~\cite{godard2019digging, zhou2017unsupervised, ranjan2019competitive, zhan2018unsupervised, clark2018ls}, video depth estimation~\cite{zhu2023lighteddepth, Teed2020DeepV2D, gu2023dro, ummenhofer2017demon}, and sparse-view NeRF~\cite{truong2023sparf, deng2022depth, jeong2021self, lin2021barf, niemeyer2022regnerf}.
Camera poses are estimated either by learning~\cite{godard2019digging,  clark2018ls, Teed2020DeepV2D, gu2023dro}, optimization~\cite{zhu2023lighteddepth, zhao2020towards} or together with NeRF~\cite{truong2023sparf,  lin2021barf}.
We propose an additional multi-view RANSAC pipeline with improved accuracy.

\Paragraph{Self-supervised Depth and Correspondence Estimation.}
Multiple works improve self-supervised depth in different ways, including learning loss~\cite{godard2019digging, watson2019self, pillai2019superdepth}, architecture~\cite{guizilini2022multi, zhou2023two}, camera pose~\cite{mahjourian2018unsupervised, zhao2020towards, chen2019self, bian2019unsupervised}, joint with semantics segmentation~\cite{zhu2020edge}, and using large-scale data~\cite{spencer2023kick, yang2024depth}.
Recently, \cite{spencer2023kick} shows self-supervision only performs on-par with supervised models under substantially more data.
\cite{yang2024depth} shows the benefit of self-supervision via exploiting non-geometry monocular semantic consistency.
Our method shows the first evident results where self-supervision benefits supervised models with only $5$ consecutive frames.

\Paragraph{Consistent Depth Estimation.}
AR applications necessitate temporally consistent depthmaps, \textit{i.e.}, depthmaps from different temporal frames reside in the same 3D space.
Recent works~\cite{zhang2021consistent, luo2020consistent} align depthmap according to the poses and points from the off-the-shelf COLMAP algorithm.
Our method seamlessly integrates SfM with monocular depthmaps, outputting consistent depth and poses.

\Paragraph{Test Time Refinement (TTR).} 
TTR aims to improve self-supervised / supervised depth estimators in testing time with RGB video~\cite{chen2019self, casser2019depth, watson2021temporal, shu2020feature, kuznietsov2021comoda}.
Methods~\cite{izquierdo2023sfm, tiwari2020pseudo} rely on off-the-shelf algorithms for pseudo depth and pose labels.
Recently, \cite{izquierdo2023sfm} first shows TTR improves supervised models.
TTR is our downstream application, which details strategies for utilizing noisy pseudo-labels.

\section{Methodology}
\begin{figure}[t!]
  \vspace{0mm}
  \captionsetup{font=small}
  \centering
  \begin{tikzpicture}
  \draw (0, 0) node[inner sep=0] {\includegraphics[width=\linewidth]{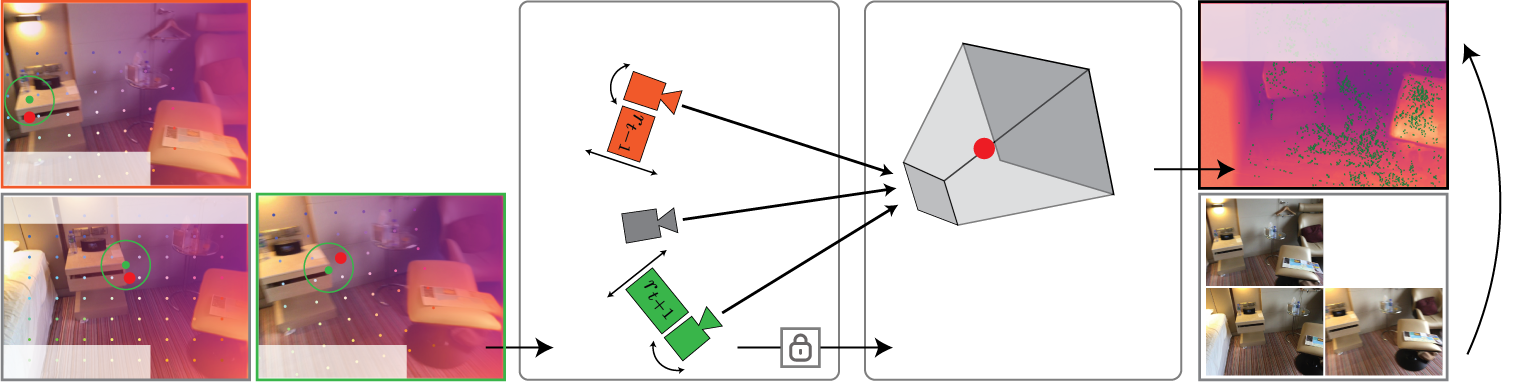}};
  \draw (-1.32-0.05, 1.5-0.15) node[inner sep=0] {\fontsize{4.0}{10}\selectfont (a). \cref{sec:pose_est} };
  \draw (-0.57-0.06, 1.35-0.15-0.05) node[inner sep=0] {\fontsize{4.0}{10}\selectfont Bundle-RANSAC-Adjustment};
  \draw (1.72-0.12, 1.5-2.8+0.05) node[inner sep=0] {\fontsize{4.0}{10}\selectfont (b). \cref{sec:triang} };
  \draw (2.2, -0.5) node[inner sep=0] {\fontsize{4.0}{10}\selectfont Frustum Radiance Field};
  \draw (2.34-0.12, 1.35-2.8) node[inner sep=0] {\fontsize{4.0}{10}\selectfont Frustum RF Triangulation};
  \draw (-5.13, -0.15) node[inner sep=0] {\fontsize{4.0}{10}\selectfont  RGB / Depth / Corr.};
  \draw (3.85+0.32-0.1, 1.5-0.15+0.035) node[inner sep=0] {\fontsize{4.0}{10}\selectfont (c). \cref{sec:geo_ver}};
  \draw (4.47+0.17-0.1, 1.35-0.15) node[inner sep=0] {\fontsize{4.0}{10}\selectfont  Geometric Verification};
  \draw (-5.5, -1.4) node[inner sep=0] {\fontsize{4.0}{10}\selectfont $\mathbf{I}_o$ at frame $t$ };
  \draw (-3.5, -1.4) node[inner sep=0] {\fontsize{4.0}{10}\selectfont Frame $t$+$1$ };
  \draw (-5.63, 0.17) node[inner sep=0] {\fontsize{4.0}{10}\selectfont Frame $t$-$1$ };
  \draw (6.1, -0.05) node[inner sep=0] {\fontsize{4.0}{10}\selectfont \rotatebox{90}{Self-supervision as Local SfM}  };
  \end{tikzpicture}
  \vspace{-4mm}
  \caption{
  \small 
  \textbf{Algorithm Overview.}
  After extracting monodepths and correspondence maps from inputs:
  (a) We apply Bundle-RANSAC-Adjustment to optimize $N-1$ camera poses $\mathcal{P}$ and $N - 1$ depth adjustments $\mathcal{R}$.
  (b) We fix poses and depth adjustments and optimize a frustum Radiance Field (RF) for triangulation.
  (c) We apply geometric verification to extract multi-view consistent 3D points via rendering with RF.
  We further detail step (a) in Figs~\ref{fig:pose_ba}, \ref{fig:hough}, and \ref{fig:ba_real}, and steps (b) and (c) in \cref{fig:ablation}.
  }
  \label{fig:framework}
  \vspace{-3mm}
\end{figure}
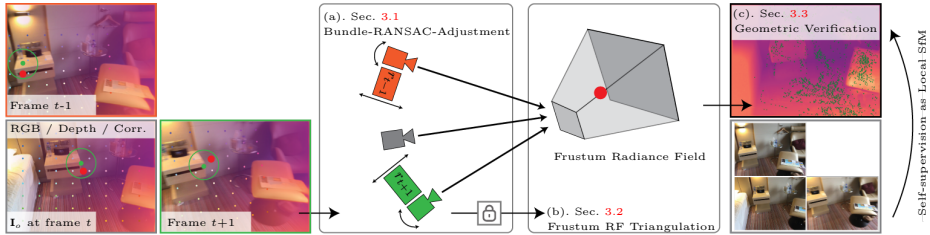

Our method runs sequentially.
From $N$ calibrated images $\mathcal{I}$, we extract $N$ monocular depthmaps $\mathcal{D}$ and $N\times(N-1)$ pair-wise dense correspondence $\mathcal{C}$.
We split the $N$ images into one root frame $\mathbf{I}_o$ in the center of the $N$-frame window where $o\!=\!\lfloor\frac{N+1}{2}\rfloor$, and $N-1$ support frames $\mathbf{I}_i$, where $i \in \mathbb{N}^+ = [1,N] \char`\\ \{o\}$. 
In~\cref{sec:pose_est}, after setting the root frame as identity pose, we use Bundle-RANSAC-Adjustment to optimize $N-1$ poses $\mathcal{P}$ and $N-1$ depth adjustments $\mathcal{R}$.
Next, in \cref{sec:triang}, we apply triangulation by optimizing a frustum Radiance Field (RF) $\mathbf{V}$, \textit{i.e.}, a NeRF without network.
Finally, in \cref{sec:geo_ver}, we apply geometric verification by rendering multi-view consistent 3D points from RF.
An overview is in \cref{fig:framework}.

\subsection{Bundle-RANSAC-Adjustment Pose Estimation}
\label{sec:pose_est}

We generalize two-view RANSAC with multi-view constraints through Bundle-Adjustment. 
\cref{sec:pipeline} describes our pipeline.
In \cref{sec:hough_ba_sec}, we propose Hough transform to accelerate computation.
We discuss the time complexity in \cref{sec:com_complex}.



\subsubsection{Optimization Pipeline}
\label{sec:pipeline}

~~~

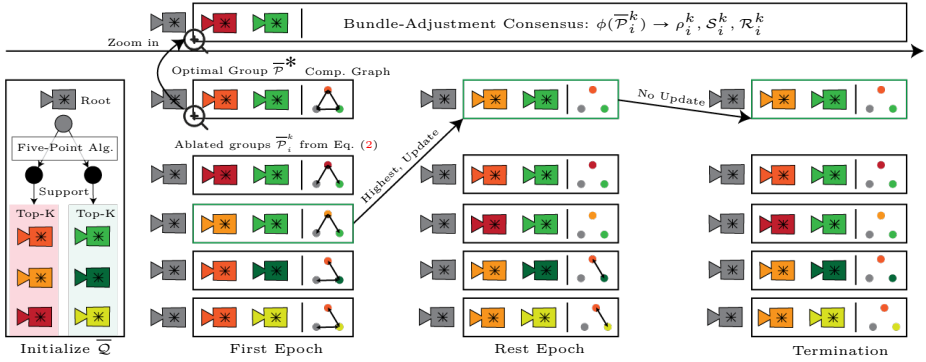
\begin{figure}[t!]
  \captionsetup{font=small}
  \centering
  \begin{tikzpicture}
  \draw (0, 0) node[inner sep=0] {\includegraphics[width=\linewidth]{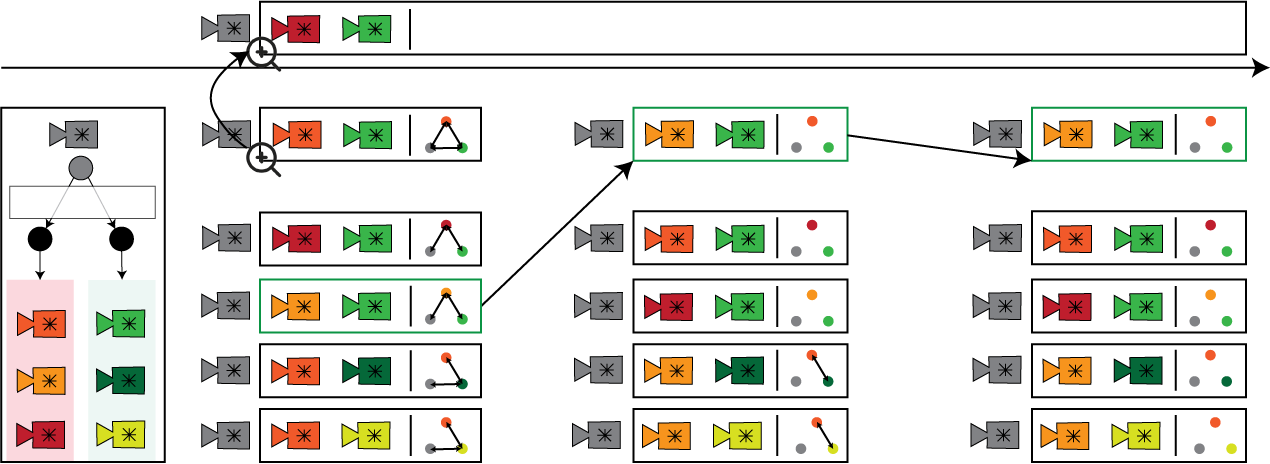}};
  \draw (-5.3, -2.35) node[inner sep=0] {\tiny Initialize $\overline{\mathcal{Q}}$};
  \draw (-4.4, 1.7) node[inner sep=0] {\fontsize{4.0}{10}\selectfont Zoom in};
  \draw (-2.5, -2.4) node[inner sep=0] {\tiny First Epoch};
  \draw (1.0, -2.4) node[inner sep=0] {\tiny Rest Epoch};
  \draw (5.0, -2.4) node[inner sep=0] {\tiny Termination};
  \draw (-2.95-0.1, 1.37) node[inner sep=0] {\fontsize{4.0}{10}\selectfont Optimal Group $\overline{\mathcal{P}}^*$};
  \draw (-1.75+0.2, 0.8+0.50) node[inner sep=0] {\fontsize{4.0}{10}\selectfont Comp. Graph};
  \draw (-2.5, 0.35) node[inner sep=0] {\fontsize{4.0}{10}\selectfont Ablated groups $\overline{\mathcal{P}}_{i}^k$ from \cref{eqn:pose_group}};
  \draw (1.2, 1.95) node[inner sep=0] {\tiny Bundle-Adjustment Consensus: $ \phi (\overline{\mathcal{P}}_{i}^k) \rightarrow \rho_{i}^k, \mathcal{S}_{i}^k, \mathcal{R}_{i}^k $};
  \draw (-5.3, 0.28) node[inner sep=0] {\fontsize{4.0}{10}\selectfont {Five-Point Alg.} };
  \draw (-4.9, 0.92) node[inner sep=0] {\fontsize{4.0}{10}\selectfont Root};
  \draw (-5.32, -0.30) node[inner sep=0] {\fontsize{4.0}{10}\selectfont Support};
  \draw (-5.7, -0.6) node[inner sep=0] {\fontsize{4.0}{10}\selectfont Top-K};
  \draw (-4.9, -0.6) node[inner sep=0] {\fontsize{4.0}{10}\selectfont Top-K};
  \draw (-0.85, 0.05) node[inner sep=0] {\fontsize{4.0}{10}\selectfont \rotatebox{44}{Highest, Update} };
  \draw (2.75, 0.95) node[inner sep=0] {\fontsize{4.0}{10}\selectfont \rotatebox{-8}{No Update} };
  \end{tikzpicture}
  \vspace{-4mm}
  \caption{
  \small 
  \textbf{Pose Optimization Pipeline.}
  We show a sample execution when $N\!=\!3$ and $K\!=\!3$.
  We initialize normalized pose candidates pool $\overline{\mathcal{Q}}$.
  Optimal group $\overline{\mathcal{P}}^*$ is set to top candidates within $\overline{\mathcal{Q}}$.
  In each epoch, \cref{eqn:pose_group} ablates pose group $\overline{\mathcal{P}}^k_{i}$.
  Each group is scored with \cref{eqn:overall_max} via BA with Hough Transform, detailed in \cref{sec:hough_ba_sec}.
  The optimal group with the highest score is updated with \cref{eqn:update}.
  Termination occurs when the maximum score stabilizes. 
  We maintain quadratic complexity by avoiding repetitive computation after the first epoch, shown with the Comp. Graph, detailed in \cref{sec:com_complex}.
  }
  \label{fig:pose_ba}
\end{figure}

\Paragraph{\underline{RAN}dom \underline{SA}mple.} 
We use five-point algorithm~\cite{li2006five} as the minimal solver.
We execute it between root and each support frame, extracting a pool of $(N-1) \times K$ normalized poses (\textit{i.e.}, pose of unit translation), $\overline{\mathcal{Q}} = \{\overline{\mathbf{P}}_{i}^k \mid i \in \mathbb{N}^+, k \in [1, K]\}$, where $\overline{\mathbf{P}}_{i}^k \in \mathbb{R}^{3\times4}$.
The $K$ is the number of normalized poses extracted per frame.
We term a set of $N-1$ normalized poses as a group $\overline{\mathcal{P}} \in \mathbb{R}^{(N-1)\times 3 \times 4}$.
Two-view RANSAC enumerates over single normalized pose $\overline{\mathbf{P}}$.
Our multi-view algorithm hence enumerates over normalized pose group $\overline{\mathcal{P}}$.
We initialize the optimal group $\overline{\mathcal{P}}^*$ as the top candidate from $K$ poses of  $\overline{\mathcal{Q}}$ for each frame.
See examples in \cref{fig:pose_ba}.

\Paragraph{Bundle-Adjustment \underline{C}onsensus.} 
While computing consensus counts, the camera scales $\mathcal{S}$ and depth adjustments $\mathcal{R}$ are automatically determined with bundle-adjustment to maximize a robust scoring function:
\begin{equation}
    \rho_{i} = \phi (\overline{\mathcal{P}}) =  \underset{\mathcal{S}, \mathcal{R}} {\max}  \; f (\mathcal{S}, \mathcal{R} \mid \overline{\mathcal{P}}, \mathcal{D}, \mathcal{C}).
    \label{eqn:overall_max}
\end{equation}

\Paragraph{Search for Optimal Group.} 
Our multi-view RANSAC has a significantly larger solution space than two-view RANSAC.
With $N$ view inputs, we determine the optimal group out of $K^{N-1}$ combinations.
Hence, we iteratively search for the optimal group with a greedy strategy.
For each epoch, we ablate $(N-1)(K-1)$ additional pose groups:
\begin{equation}
    \overline{\mathcal{P}}_{i}^k = \overline{\mathcal{P}}_i^* \; \char`\\ \; \{ \overline{\mathbf{P}}_i^* \} \;  \cup \; \{\overline{\mathbf{P}}_{i}^k\},
    \label{eqn:pose_group}
\end{equation}
where $i \in \mathbb{N}^+$ and $k \in [1, K]$.
Combine \cref{eqn:pose_group} and \cref{fig:pose_ba}, taking frame $i$ as an example, we replace the optimal pose $\overline{\mathbf{P}}_i^*$ by its $K-1$ other candidates $\overline{\mathbf{P}}_{i}^k$, generating $K-1$ groups.
For $N$ frames, we have $(N-1)(K-1) + 1$ groups.
We apply bundle-adjustment to each group to evaluate \cref{eqn:overall_max}.
As shown in \cref{fig:framework} and \cref{fig:pose_ba}, we select the normalized pose together with its optimized scales and depth adjustments that maximize the scores as the output,
\begin{equation}
    \mathcal{P}_i^* = b ( \overline{\mathcal{P}}_{i}^*, \mathcal{S}_{i}^*), \;  \mathcal{R}_i^* =  \mathcal{R}_{i}^k, \;  \text{where} \, k = \arg \max \{ \rho_{i}^k \}, \; \overline{\mathcal{P}}_i^* = \overline{\mathcal{P}}_{i}^k, \; \mathcal{S}_{i}^* = \mathcal{S}_{i}^k, \;
    \label{eqn:update}
\end{equation}
where $b(\cdot)$ combines normalized poses with scales.
\cref{fig:triang_depth} third column plots an adjusted temporal consistent depthmap after applying $\mathcal{R}^*$.
In \cref{fig:pose_ba}, the algorithm terminates when the maximum score stops increasing.

\Paragraph{Scoring Function.} 
Similar to other RANSAC methods, we adopt robust inlier-counting based scoring functions.
Expand \cref{eqn:overall_max} for a specific group $\overline{\mathcal{P}}$:
\begin{equation}
\small
    \phi (\overline{\mathcal{P}}) = \sum_{i, i\neq j} \sum_{j} f_{i, j} (s_{i}, s_{j}, r_{i}, r_{j} \mid \overline{\mathbf{P}}_{i}, \overline{\mathbf{P}}_{j}, \mathbf{D}_i, \mathbf{D}_j, \mathbf{C}_{i, j}),
    \label{eqn:overall_onegroup_max}
\end{equation}
where $i, j$ are frame index.
We set per-frame camera scale, depth, depth adjustment, and correspondence as $s \in \mathcal{S}, \, \mathbf{D} \in \mathcal{D}$, $r \in \mathcal{R}$, and $\mathbf{C} \in \mathcal{C}$.
The scoring function $f_{i, j}(\cdot)$ has various forms.
First, we describe a 2D scoring function:
\begin{equation}
\small
    f_{i, j}^{\text{2D}} (\cdot) = \sum_m \mathbf{1} \left( \| \pi(s_i, s_j, r_i \mid \overline{\mathbf{P}}_{i}, \overline{\mathbf{P}}_{j}, d_i^m) - \mathbf{c}_{i, j}^m \|_2 < \lambda^{\text{2D}} \right),
    \label{eqn:loss_2D}
\end{equation}
where $m \in [1, M]$ indexes sampled pixels per frame pair.
$f_{i, j}^{\text{2D}} (\cdot)$ measures the inlier count between depth projected correspondence and input correspondence. 
$\pi (\cdot)$ is projection process. 
Intrinsic is skipped.
$d$ and $\mathbf{c}$ are depth and correspondence sampled from $\mathbf{D}$ and $\mathbf{C}$.
An example is in \cref{fig:hough}. 
The $\mathbf{1}(\cdot)$ is the indicator function.
The projected pixel is an inlier if it resides within the circle of radius $\lambda^{\text{2D}}$ and center at correspondence $\mathbf{c}_{i, j}^m$ (denoted as $\mathbf{p}_j$ in \cref{fig:hough}). 
$\mathbf{c}_{i, j}^m$ is sampled from correspondence map $\mathbf{C}_{i, j}$.
\noindent Second, we introduce a 3D scoring function:
\begin{equation}
\small
    f_{i, j}^{\text{3D}} (\cdot)= \sum_m \mathbf{1}\left( \| \pi\inv (s_i \mid \overline{\mathbf{P}}_{i}, r_i, d_i^m) - \pi\inv (s_j \mid \overline{\mathbf{P}}_{j}, r_j, d_j^m) \|_2 < \lambda^{\text{3D}} \right).
    \label{eqn:loss_3D}
\end{equation}
Depth pair $d_i$ and $d_j$ is determined by correspondence.
Unlike the 2D one, the 3D function fixes depth adjustment $r$.
Function $\pi^{-1}(\cdot )$ back-projects 3D point.


\subsubsection{Hough Transform Acceleration}
\label{sec:hough_ba_sec}

~~

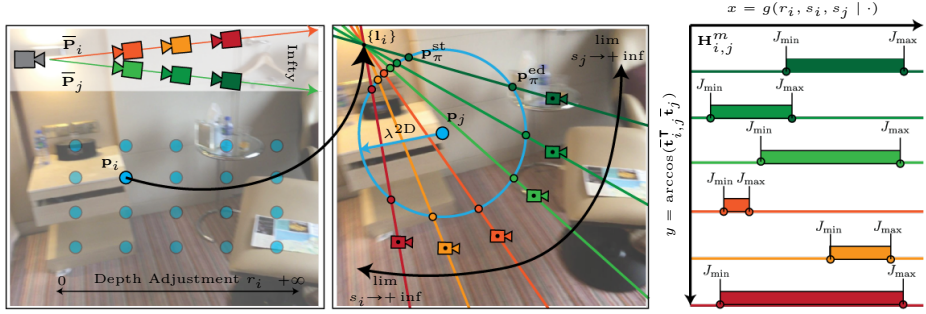
\begin{figure}[t!]
  \captionsetup{font=small}
  \centering
  \begin{tikzpicture}
  \draw (0, 0) node[inner sep=0] {\includegraphics[width=\linewidth]{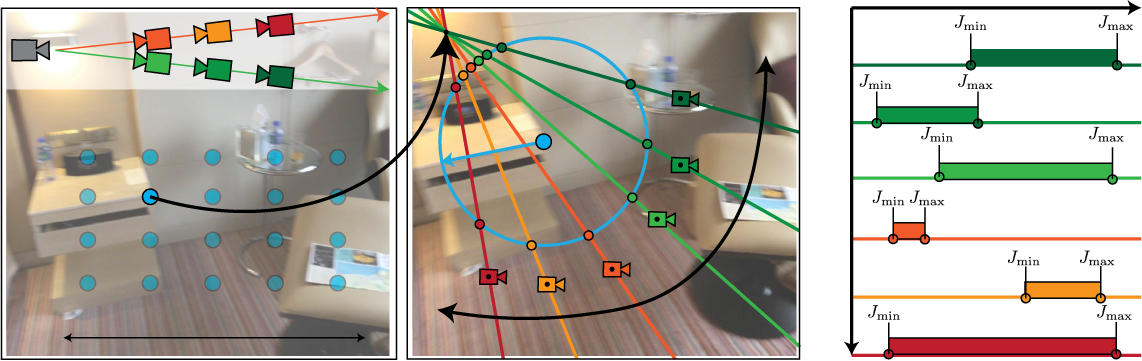}};
  \draw (4.50, 2.05) node[inner sep=0] {\tiny $x = g(r_i, s_i, s_j \mid \cdot)$};
  \draw (2.75, 0.0) node[inner sep=0] {\tiny \rotatebox{90}{$y = \arccos(\overline{\mathbf{t}}_{i, j}^\intercal \overline{\mathbf{t}}_{j})$}};
  \draw (-5.2, 1.6) node[inner sep=0] {\tiny $\overline{\mathbf{P}}_i$};
  \draw (-5.2, 1.1) node[inner sep=0] {\tiny $\overline{\mathbf{P}}_j$};
  \draw (-2.3, 1.35) node[inner sep=0] {\tiny \rotatebox{-90}{Infty}};
  \draw (-5.35, -1.55) node[inner sep=0] {\tiny $0$};
  \draw (-2.3, -1.55) node[inner sep=0] {\tiny $+\infty$};
  \draw (-3.8, -1.55) node[inner sep=0] {\tiny Depth Adjustment $r_i$};
  \draw (-4.7, -0.0) node[inner sep=0] {\tiny $\mathbf{p}_i$};
  \draw (-0.10, 0.60) node[inner sep=0] {\tiny $\mathbf{p}_j$};
  \draw (-0.85, 0.45) node[inner sep=0] {\tiny $\lambda^{\text{2D}}$};
  \draw (1.9, 1.5) node[inner sep=0] {\tiny $\underset{s_j \rightarrow +\inf}{\lim}$};
  \draw (-1.05, -1.7) node[inner sep=0] {\tiny $\underset{s_i \rightarrow +\inf}{\lim}$};
  \draw (-1.1, 1.70) node[inner sep=0] {\tiny $\{\mathbf{l}_i\}$};
  \draw (-0.4, 1.5) node[inner sep=0] {\tiny $\mathbf{p}_{\pi}^{\text{st}}$};
  \draw (0.9, 1.2) node[inner sep=0] {\tiny $\mathbf{p}_{\pi}^{\text{ed}}$};
  \draw (3.35, 1.6) node[inner sep=0] {\tiny $\mathbf{H}_{i, j}^m$};
  \end{tikzpicture}
  \vspace{-4mm}
  \caption{
  \small 
  \textbf{Hough Transform between Two Normalized Poses.} 
  With fixed normalized poses, there exists three variables, scales $s_i$  \& $s_j$  of $\overline{\mathbf{P}}_i$ \& $\overline{\mathbf{P}}_j$ and adjustment $r_i$.
  Pixel $\mathbf{p}_i$ and $\mathbf{p}_j$ are corresponded. 
  Ablating pose scales maps pixel $\mathbf{p}_i$ to a set of epipolar lines $\{\mathbf{l}_i\}$, however, bounded by \textcolor{hough_red}{Red} and \textcolor{hough_green}{Green} at infinite scales.
  We have three observations.
  First, with fixed normalized poses, epipolar lines $\mathbf{l}_i$ have limited possibilities.
  Second, scale $s$ and depth adjustment $d$ are equivalent, both adjusting projection on epipolar line.
  Third, per epipolar line, to be an inlier, the projection has to reside within the line-circle intersection, between $\mathbf{p}_{\pi}^{\text{st}}$ and $\mathbf{p}_{\pi}^{\text{ed}}$.
  The observations motivate us to discretize the solution space to a 2D matrix, \textit{i.e.}, Hough Transform.
  Right figure plots an example transformation $\mathbf{H}_{i, j}^m$ from frame $i$ to $j$ on the $m$th pixel $\mathbf{p}_i$.
  }
  \label{fig:hough}
\end{figure}

\noindent 
Maximizing \cref{eqn:overall_max} for each pose group is computationally prohibitive, as shown in \cref{fig:pose_ba}.
We propose Hough Transform for acceleration.
We use \cref{eqn:loss_2D}, the 2D function $f^{\text{2D}}(\cdot)$  as an example for illustration.
See our motivation in \cref{fig:hough}.

\Paragraph{Hough Transform.}
The relative pose between $\overline{\mathbf{P}}_i$ and $\overline{\mathbf{P}}_j$ is defined as:
\begin{equation}
\begin{aligned} 
    \mathbf{P}_{i, j} = \mathbf{P}_j \mathbf{P}_i\inv =
    \begin{bmatrix}
    \mathbf{R}_{i, j} & s_{i, j} \overline{\mathbf{t}}_{i, j}
    \end{bmatrix}
    =
    \begin{bmatrix}
    \mathbf{R}_j \mathbf{R}_i^{\text{-}1} & -s_i \mathbf{R}_j \mathbf{R}_i^{\text{-}1} \overline{\mathbf{t}}_i + s_j \overline{\mathbf{t}}_j
    \end{bmatrix},
    \label{eqn:two_view}
\end{aligned}
\end{equation}
where $\mathbf{R}, \overline{\mathbf{t}}$, and $s$ are rotation, normalized translation and pose scale.
From \cref{eqn:two_view} and \cref{fig:hough}, $\overline{\mathbf{t}}_{i, j}$ is controlled by the scale $s_i$ and $s_j$, and 
thus we have:
\begin{equation}
   \lim_{s_i \rightarrow +\inf}  \overline{\mathbf{t}}_{i, j} = -\mathbf{R}_j \mathbf{R}_i^{\text{-}1} \overline{\mathbf{t}}_i, \quad \lim_{s_j \rightarrow +\inf} \overline{\mathbf{t}}_{i, j} =    \overline{\mathbf{t}}_j .
    \label{eqn:lim}
\end{equation}
For a pixel $\mathbf{p}_i$ on frame $i$, its corresponding epipolar line $\mathbf{l}_i$ on frame $j$ is:
\begin{equation}
    \mathbf{l}_i =\mathbf{K}^{\text{-}\intercal} [\,\overline{\mathbf{t}}_{i, j}{]}_{\times} \mathbf{R}_{i, j}  \mathbf{K}\inv \mathbf{p}_i.
    \label{eqn:eppline}
\end{equation}
\cref{eqn:lim} and \cref{eqn:eppline} suggest the epipolar line has limited possibilities.
Operation $[ \cdot{]}_{\times}$ is the cross product in matrix form.
Further, as the depth re-projected pixel  $\mathbf{p}_{\pi}$ of $\mathbf{p}_i$ always locate on the epipolar line $\mathbf{l}_i$~\cite{hartley2003multiple}, we have:
\begin{equation}
  \mathbf{l}_i^\intercal \mathbf{p}_{\pi} = 0, \;  \mathbf{p}_{\pi} = \pi (s_i, s_j, r_i \mid \overline{\mathbf{P}}_i, \overline{\mathbf{P}}_j, d_i).
  \label{eqn:line}
\end{equation}
To be an inlier of the scoring function $f^\text{2D}(\cdot)$, we have:
\begin{equation}
    \| \mathbf{p}_{\pi} - \mathbf{p}_j\|_2 \leq \lambda^{\text{2D}}.
    \label{eqn:circle}
\end{equation}
Combining \cref{eqn:line}, \cref{eqn:circle} and \cref{fig:hough}, to be an inlier, the projected pixel $\mathbf{p}_{\pi}$ has to reside within the line segment, with two end-points computed by the line-circle intersection. 
The circle centers at corresponded pixel $\mathbf{p}_j$ on frame $j$ with a radius $\lambda^{\text{2D}}$.
We denote the two end-points $\mathbf{p}_{\pi}^{\text{st}}$ and $\mathbf{p}_{\pi}^{\text{ed}}$.
We add their calculation in Supp.
Function $J(\cdot)$ follows \cite{zhu2023lighteddepth} Supp. Eq.~($\color{red}{4}$), which maps a projected pixel $\mathbf{p}_{\pi}$ and adjusted depth $r_i d_i$ to camera scale $s_{i, j}$ as:
$
    s_{i, j} = J(\overline{\mathbf{P}}_{i, j}, r_i d_i, \mathbf{p}_{\pi}).
$

\begin{corollary}
\label{cor:corollary1}
A pixel is an inlier iff:
\begin{equation}
    J(\overline{\mathbf{P}}_{i, j}, r_i d_i, \mathbf{p}_{\pi}^{\text{st}}) \leq s_{i, j} \leq J(\overline{\mathbf{P}}_{i, j}, r_i d_i, \mathbf{p}_{\pi}^{\text{ed}}).
    \label{eqn:collary1}
\end{equation}
\end{corollary}

\begin{corollary}
\label{cor:corollary2}
Scale and depth are equivalent as;
\begin{equation}
    s_{i, j} = J(\overline{\mathbf{P}}_{i, j}, r_i d_i, \mathbf{p}_{\pi}) = r_i \cdot J(\overline{\mathbf{P}}_{i, j}, d_i, \mathbf{p}_{\pi}).
    \label{eqn:collary2}
\end{equation}
\end{corollary}

\noindent See \cref{fig:hough} and proof in Supp. 
Combine Eqs.~(\ref{eqn:collary1}) and (\ref{eqn:collary2}),
\begin{equation}
   J(\overline{\mathbf{P}}_{i, j}, d_i, \mathbf{p}_{\pi}^{\text{st}}) \leq \frac{s_{i, j}}{r_i} \leq J(\overline{\mathbf{P}}_{i, j}, d_i, \mathbf{p}_{\pi}^{\text{ed}}).
   \label{eqn:hough_x}
\end{equation}
Set $g(\cdot)$ maps the variables under optimization to intermediate term $\frac{s_{i, j}}{r_i}$:
\begin{equation}
\small
    J(\overline{\mathbf{P}}_{i, j}, d_i, \mathbf{p}_{\pi}^{\text{st}}) 
    \leq g(r_i, s_i, s_j \mid \overline{\mathbf{P}}_{i}, \overline{\mathbf{P}}_{j}) \leq 
    J(\overline{\mathbf{P}}_{i, j}, d_i, \mathbf{p}_{\pi}^{\text{ed}}).
    \label{eqn:hough}
\end{equation}
The $i$th pixel is an inlier if and only if its projection satisfies \cref{eqn:hough}.
Note, the value space of function $g(\cdot)$ is mapped to a 2D space $\mathbf{H}$ after Hough Transform:
\begin{equation}
x = g(r_i, s_i, s_j \mid \overline{\mathbf{P}}_{i}, \overline{\mathbf{P}}_{j}), \; y = \arccos(\overline{\mathbf{t}}_{i, j}^\intercal \overline{\mathbf{t}}_{j}),
\label{eqn:map_xy}
\end{equation}
where $x$ and $y$ are transformed coordinates.
From \cref{eqn:map_xy}, $x$ is a synthesized translation magnitude and $y$ is angular variable.
We then set $x \in [0, x_{\text{max}}]$, and $ y \in [0, \theta_{\text{max}}]$, where
$\theta_{\text{max}} = \arccos ( -\overline{\mathbf{t}}_{j}^\intercal  { \mathbf{R}_{i, j}  \overline{\mathbf{t}}_i})$.
Finally, the value of $\mathbf{H}$ is:
\begin{equation}
    \forall y \in [0, \theta_{\text{max}}], \; \mathbf{H}(x \mid y) = 1, \text{if} \;\; x \in [J_{\text{min}}  , J_{\text{max}}],
\end{equation}
where $J_{\text{min}}$ and $J_{\text{max}}$ are the two bounds from \cref{eqn:hough}.
The transformation over the scoring function $f_{i, j}^{\text{2D}}$ with all $M$ sampled pixels between frame $\mathbf{I}_i$ and $\mathbf{I}_j$:
\begin{equation}
\small
     \mathbf{H}_{i, j} = \sum_m \mathbf{H}^m_{i, j}, \; f_{i, j}^{\text{2D}} (s_i, s_j, r_i \mid \overline{\mathbf{P}}_i, \overline{\mathbf{P}}_j)  =  \mathbf{H}_{i, j} (x, y),
     \label{eqn:hough_index}
\end{equation}
where $x$ and $y$ are functions of $s_i, s_j, r_i$.
\cref{eqn:overall_max} becomes:
\begin{equation}
    \phi (\overline{\mathcal{P}}) =  \underset{\mathcal{S}, \mathcal{R}} {\max}  \; \sum_{i} \sum_{j, j\neq i} \mathbf{H}_{i, j}(x(\mathcal{S}, \mathcal{R}), y(\mathcal{S}, \mathcal{R})). 
    \label{eqn:hough_score_total}
\end{equation}
In our implementation, we discretize $\mathbf{H}_{i, j}$ to a 2D matrix.

\begin{figure}[t!]
  \captionsetup{font=small}
  \centering
  \includegraphics[width=\linewidth]{
  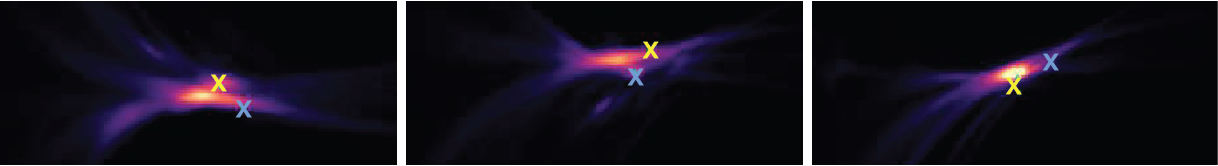
  }
  \vspace{-4mm}
  \caption{
  \small 
  \textbf{Visualize Hough Transform Matrix $\mathbf{H}_i^j$} from \cref{eqn:hough_index}.
  Area with higher intensity suggests more inlier counts.
  Given normalized pose group, for $N$ views, there exists $N\times(N-1)$ matrices $\mathbf{H}_i^j$, constraining $N-1$ scale and $N-1$ adjustments.
  We plot the \textcolor{hough_matrix_blue}{start} and \textcolor{orange}{end} points after optimizing \cref{eqn:hough_score_total} in the figure.
  }
  \label{fig:ba_real}
  \vspace{-2mm}
\end{figure}

\Paragraph{Accelerate Bundle-Adjustment Consensus.} 
The BA determines $N-1$ camera scales and $N-1$ depth adjustments to maximize the scoring function $\phi(\cdot)$ in \cref{eqn:hough_score_total}.
With Hough transform, BA maximizes the summarized intensity via \textbf{indexing} $N\times(N-1)$ Hough transform matrices $\mathbf{H}$.
It avoids BA repetitively enumerating all sampled pixels.
\cref{fig:ba_real} shows an example optimization process.

\Paragraph{Certified Global Optimality} of robust inlier-counts scoring function \cref{eqn:loss_2D} and \cref{eqn:loss_3D} are achieved after optimization.
See \cref{fig:corres_curve} for more analysis.

\Paragraph{Optimization with RGB-D.}
With GT depthmap, the algorithm switches to the 3D scoring function $f_{i, j}^{\text{3D}}(\cdot)$.
The depth adjustment is fixed to $1$ and the 2D line-circle intersection becomes 3D line-sphere intersection. 
See Supp.~for details.

\subsubsection{Computational Complexity}
\label{sec:com_complex}

~~~

\Paragraph{Naive Time Complexity.} From \cref{eqn:pose_group} and \cref{fig:pose_ba}, in each epoch, we evaluate $(N-1) (K-1)$ pose groups with Hough Transform Acceleration.
Suppose each group takes $T$ iterations to optimize \cref{eqn:hough_score_total}, the time complexity is:
\begin{equation}
     \mathcal{O}((N-1) (K-1) \cdot N (N-1) \cdot(M + T)),
\end{equation}
where each group computes $N (N-1)$ Hough matrices $\mathbf{H}$.
Each matrix enumerates $M$ sampled pixels, see \cref{eqn:hough_index}.
Maximizing \cref{eqn:hough_score_total} becomes indexing $\mathbf{H}$, hence has a constant time complexity $T$, where $T << M$.

\Paragraph{Counting Unique Hough Matrices.}
Most computation is spent on Hough matrices.
In \cref{fig:pose_ba}, each connection in the computation graph suggests two unique Hough matrices.
We minimize time complexity by only computing \textbf{unique} Hough matrices.
In \cref{fig:pose_ba} first epoch, the initial optimal group $\overline{\mathcal{P}}^*$ has $N(N-1)$ matrices.
Each ablated group only differs by one pose, hence introducing $2(N-1)(N-1)(K-1)$ matrices.
The first-epoch complexity is then:
\begin{equation}
\small
    \mathcal{O}^{H}( N (N-1) M + 2(N-1)^2 (K-1) M) + \mathcal{O}^{\text{BA}}( N (N-1) (K-1)T).
    \label{eqn:init}
\end{equation}
Only the Hough transform is accelerated. 
As $T << M$, the  complexity of BA is neglectable.
After the first epoch, $\overline{\mathcal{P}}^*$ only updates one pose per epoch, hence introducing $2(N-2)(K-1)$ matrices.
The complexity for the rest epochs is, 
\begin{equation}
\small
    \mathcal{O}^{H}(2(N-2)(K-1) M) + \mathcal{O}^{\text{BA}}( N (N-1) (K-1)T).
    \label{eqn:rest}
\end{equation}
While \cref{eqn:rest} has linear complexity, our method only updates one pose per epoch.
Updating poses in all frames like other SfM methods is still quadratic.

\begin{figure}[t!]
    \captionsetup{font=small}
    \centering
    \subfloat[Triangulation]{\includegraphics[width=0.32\linewidth]{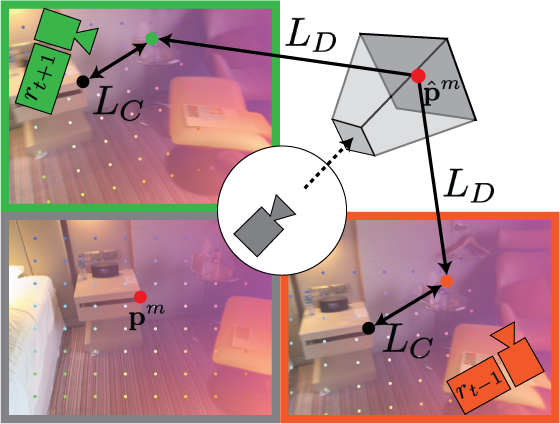}} \quad \quad
    \subfloat[Geometric Verification ]{\includegraphics[width=0.32\linewidth]{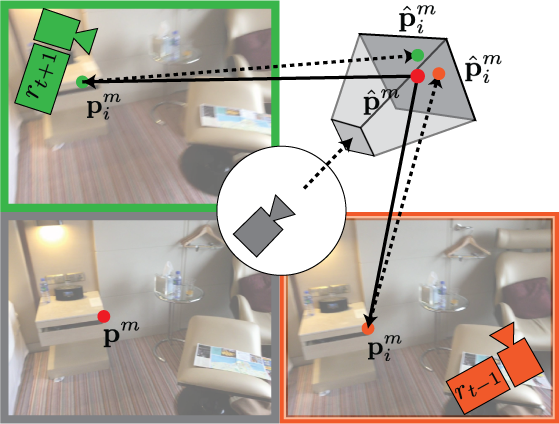}}
    \vspace{-2mm}
    \caption{\small 
    \textbf{Triangulation} optimizes frustum RF for multiview consistency \textit{w.r.t.} depth and correspondence.
    \textbf{Geometric Verification} inferences RF for sparse multiview consistent 3D points.
    For simplicity, in (a), we only plot $L_c$ defined from the root frame.
    }
    \label{fig:ablation}
    \vspace{-3mm}
\end{figure}

\subsection{Frustum Radiance Field Triangulation}
\label{sec:triang}
\Paragraph{Frustum Radiance Field.}
Now, we fix the optimized pose $\mathcal{P}^*$.
Then we employ a frustum radiance field $\mathbf{V}$ of size $H \times W \times D$ for dense triangulation.
Field $\mathbf{V}$ is defined over the root frame $\mathbf{I}_o$ and shares similarity with the categorical depthmap~\cite{fu2018deep, bhat2021adabins}.
We follow \cite{wang2021nerf, truong2023sparf} in rendering the depth $d$.
The RGB estimation is skipped as unrelated.
A 3D ray originated from pixel $\mathbf{p}_i$ at frame $i$ is discretized into a set of 3D points and depth labels.
With slight abuse of notation, we denote $ \{\hat{\mathbf{p}}_{i,t} = \mathbf{o} + d_t \mathbf{r} \mid t \in [1, T]\}$, where $\hat{\mathbf{p}}$ is a 3D point, $d_t$ is depth label and $\mathbf{r}$ is ray direction. 
Set integration interval $\delta_t = d_{t+1} - d_t$, depth $d$ is:
\begin{equation}
    d({\mathbf{p}_i}) = \sum_{t} \alpha_t d_t, \;
    \alpha_t = T_t  (1-\exp \left(-\sigma_t \delta_t\right)), \; T_t=\exp (-\sum_{t^{\prime} \in [1, t]} \sigma_{t^{\prime}} \delta_{t^{\prime}}).
    \label{eqn:nerf}
\end{equation}
We set the camera origin of frame $i$ as $\mathbf{o}$.
Instead of regressing occupancy $\delta$ with MLP~\cite{wang2021nerf, truong2023sparf}, 
we directly interpolate the radiance field $\mathbf{V}$:
\begin{equation}
    \delta_t = \mathbf{V} (u, v, w), \text{where} \; \begin{bmatrix}
        u & v & w
    \end{bmatrix}^\intercal = \pi(\mathbf{E}, \hat{\mathbf{p}}_{i, t}).
\end{equation}
Matrix $\mathbf{E}$ is the identity matrix. 
Function $\pi(\cdot)$ is projection function.
Compared to using the MLP, frustum radiance field $\mathbf{V}$ is more computationally efficient~\cite{fridovich2022plenoxels}.

\Paragraph{Triangulation.}
Classic triangulation method~\cite{schonberger2016structure} operates on a single 3D point.
The RF provides additional constraints where all optimized points share a canonical 3D volume. 
In \cref{fig:ablation}, we supervise $\mathbf{V}$ for multi-view consistency between dense depthmap $\mathcal{D}$ and correspondence map $\mathcal{C}$.
On depth:
\begin{equation}
\small
    L_D = \frac{1}{NM} \sum_i \sum_m \|\pi (\mathbf{P}_i, \hat{\mathbf{p}}^{m}) - d_{i}^m \|_1.
\end{equation}
Here, $\hat{\mathbf{p}}^{m}$ is rendered from the root frame, following depth computed with \cref{eqn:nerf}.
To apply correspondence consistency, we have:
\begin{equation}
\small
L_C = \frac{1}{N(N-1)M} \sum_i \sum_{j, j\neq i} \sum_m \|\pi (\mathbf{P}_j, \hat{\mathbf{p}}_{i}^m) - \mathbf{q}_{i, j}^m \|_1,
\label{eqn:pixel_loc}
\end{equation}
where $ \hat{\mathbf{p}}_{i}^m = \pi\inv (\mathbf{P}_i, \mathbf{p}_{i}^m, d(\mathbf{p}_{i}^m)), \mathbf{p}_{i}^m = \pi(\mathbf{P}_i, \hat{\mathbf{p}}^m).$
With slight abuse of notation, function $\pi (\cdot)$ returns depth for $L_D$, and location for $L_C$.
We \textbf{always} first render from the root frame and subsequently project to $N$ frames. 
From there, we project to other supported frames again, forming $N(N-1)$ pairs.

\subsection{Geometric Verification}
With the RF optimized, we apply geometric verification to acquire sparse multi-view consistent 3D points, as in \cref{fig:ablation}:
\begin{equation}
\small
   \mathcal{C} = \{ \sum_{i, i\neq o} c_{i}^m \geq n^{\text{c}} \}, \; c_{i}^m = 1 \; \text{if} \sum_{i, i\neq o} \| \hat{\mathbf{p}}_{i}^m - \hat{\mathbf{p}}^m \|_2 \leq \lambda^c.
\end{equation}
We follow the same rendering process as training, where $\hat{\mathbf{p}}_{i}^m$ is computed with \cref{eqn:pixel_loc}.
First, we render 3D points from the root frame, project them to other views, and render 3D points from there again.
A point is valid if a minimum of $n^{\text{c}}$ views are consistent with the root.
\label{sec:geo_ver}

\section{Experiments}

\begin{table*}[t!]
  \centering
  \captionsetup{font=small}
  \caption{
  \small
  \textbf{
  Self-Supervised Depth Estimation.
  }
  We apply self-supervision with $5$ frames via executing the local SfM.
  We output improved sparse depthmaps over SoTA supervised inputs.
  The evaluation is conducted over the root frame.
  }
  \resizebox{ 0.95\linewidth}{!}{%
  \begin{tabular}{c|lc|cc|ccccc}
    \hline
    \textbf{Dataset} & \textbf{Method} & \textbf{Density}  & \cellcolor[RGB]{155, 187, 228} $\mathbf{\delta_{0.5}}$ & \cellcolor[RGB]{155, 187, 228} $\mathbf{\delta_1}$ & \cellcolor[RGB]{222, 164, 151} $\textbf{SI}_{\textbf{log}}$ & \cellcolor[RGB]{222, 164, 151} \textbf{A.Rel} & \cellcolor[RGB]{222, 164, 151} \textbf{S.Rel}& \cellcolor[RGB]{222, 164, 151} \textbf{RMS} & \cellcolor[RGB]{222, 164, 151} $\textbf{RMS}_{\textbf{log}}$ \\
    \hline
    \multirow{6}{*}{ScanNet~\cite{dai2017scannet}} 
    & ZoeDepth~\cite{bhat2023zoedepth} & \multirow{2}{*}{ $9.1 \%$}   & $0.877$ & $0.963$      & $6.655$          & $0.056$          & $0.016$          & $0.154$          & $0.075$  \\
    & $\drsh$ Ours &  & $\mathbf{0.902}$ & $\mathbf{0.976}$ &  $\mathbf{5.901}$ & $\mathbf{0.050}$ & $\mathbf{0.014}$ & $\mathbf{0.149}$ & $\mathbf{0.070}$   \\
    \cdashline{2-10}
    & ZeroDepth~\cite{liu2023zero} & \multirow{2}{*}{ $5.6 \%$}   & $0.641$ & $0.834$      & $12.860$          & $0.124$          & $0.086$          & $0.337$          & $0.152$  \\
    & $\drsh$ Ours &  & $\mathbf{0.686}$ & $\mathbf{0.877}$ &  $\mathbf{9.463}$ & $\mathbf{0.106}$ & $\mathbf{0.067}$ & $\mathbf{0.295}$ & $\mathbf{0.133}$   \\
    \cdashline{2-10}
    & Metric3D~\cite{yin2023metric3d} & \multirow{2}{*}{ $2.6 \%$}   & $0.804$ & $0.946$      & $6.708$          & $0.067$          & $0.020$          & $0.150$          & $0.084$  \\
    & $\drsh$ Ours & & $\mathbf{0.854}$ & $\mathbf{0.968}$ &  $\mathbf{4.170}$ & $\mathbf{0.055}$ & $\mathbf{0.014}$ & $\mathbf{0.125}$ & $\mathbf{0.068}$   \\
    \bottomrule
    \multirow{6}{*}{KITTI360~\cite{liao2022kitti}} 
    & ZoeDepth~\cite{bhat2023zoedepth} & 
    \multirow{2}{*}{ $4.0 \%$}   & $0.677$ & $0.899$      & $14.154$          & $0.103$          & $0.490$          & $3.521$          & $0.153$  
    \\
    & $\drsh$ Ours & & 
    $\mathbf{0.719}$ & $\mathbf{0.910}$ &  $\mathbf{13.220}$ & $\mathbf{0.094}$ & $\mathbf{0.474}$ & $\mathbf{3.499}$ & $\mathbf{0.145}$   
    \\
    \cdashline{2-10}
    & ZeroDepth~\cite{liu2023zero} & 
    \multirow{2}{*}{ $4.5 \%$}   & $0.584$ & $0.844$  & $16.468$          & $0.132$  & $0.819$      & $3.486$          & $0.183$  
    \\
    & $\drsh$ Ours & & 
    $\mathbf{0.654}$ & $\mathbf{0.877}$ &  $\mathbf{13.881}$ & $\mathbf{0.115}$ & $\mathbf{0.772}$ & $\mathbf{3.395}$ & $\mathbf{0.164}$   
    \\
    \cdashline{2-10}
    & Metric3D~\cite{yin2023metric3d} & 
    \multirow{2}{*}{ $3.2 \%$}   & $0.846$ & $0.958$      & $9.226$          & $0.072$          & $0.508$          & $2.194$          & $0.104$  
    \\
    & $\drsh$ Ours &  & 
    $\mathbf{0.860}$ & $\mathbf{0.963}$ &  $\mathbf{8.896}$ & $\mathbf{0.068}$ & $\mathbf{0.487}$ & $\mathbf{2.139}$ & $\mathbf{0.101}$   
    \\
    \bottomrule
  \end{tabular}
  }
  \vspace{2mm}
  \label{tab:self_sup_depth}
\end{table*}

\begin{table*}[t!]
\vspace{-2mm}
  \centering
  \captionsetup{font=small}
  \caption{
  \small
  \textbf{
  Consistent Depth Estimation.
  }
  We measure the numerical improvement by aligning the support frame depthmaps to the root frame with our depth adjustment scalars.
  The evaluation is conducted on support frames on ScanNet~\cite{dai2017scannet}.
  }
  \resizebox{ 0.75\linewidth}{!}{%
  \begin{tabular}{l|cc|ccccc}
    \hline
    \textbf{Method} & \cellcolor[RGB]{155, 187, 228} $\mathbf{\delta_{0.5}}$ & \cellcolor[RGB]{155, 187, 228} $\mathbf{\delta_1}$ & \cellcolor[RGB]{222, 164, 151} $\textbf{SI}_{\textbf{log}}$ & \cellcolor[RGB]{222, 164, 151} \textbf{A.Rel} & \cellcolor[RGB]{222, 164, 151} \textbf{S.Rel}& \cellcolor[RGB]{222, 164, 151} \textbf{RMS} & \cellcolor[RGB]{222, 164, 151} $\textbf{RMS}_{\textbf{log}}$ \\
    \hline
    ZoeDepth~\cite{bhat2023zoedepth} & $0.658$ & $0.894$     & $\mathbf{9.242}$ & $0.104$          & $0.039$          & $0.255$          & $0.128$   \\
    $\drsh$ Ours  & $\mathbf{0.793}$ & $\mathbf{0.942}$  & $\mathbf{9.242}$ & $\mathbf{0.079}$ & $\mathbf{0.024}$ & $\mathbf{0.203}$ & $\mathbf{0.105}$   \\
    \cdashline{1-8}
    ZeroDepth~\cite{liu2023zero}  & $0.351$ & $0.589$     & $\mathbf{20.145}$ & $0.254$          & $0.223$          & $0.565$          & $0.287$   \\
    $\drsh$ Ours  & $\mathbf{0.490}$ & $\mathbf{0.725}$  & $\mathbf{20.145}$ & $\mathbf{0.199}$ & $\mathbf{0.156}$ & $\mathbf{0.457}$ & $\mathbf{0.237}$ \\
    \cdashline{1-8}
    Metric3D~\cite{yin2023metric3d} & $0.533$ & $0.753$      & $\mathbf{12.425}$          & $0.216$          & $0.339$          & $0.495$          & $0.228$  \\
    $\drsh$ Ours & $\mathbf{0.664}$ & $\mathbf{0.838}$ &  $\mathbf{12.425}$ & $\mathbf{0.137}$ & $\mathbf{0.126}$ & $\mathbf{0.345}$ & $\mathbf{0.175}$   \\
    \bottomrule
  \end{tabular}
  }
  \vspace{2mm}
  \label{tab:consist_depth}
\end{table*}

\begin{table}
  \centering
  \captionsetup{font=small}
  \caption{
  \small
  \textbf{Self-Supervised Correspondence Estimation.} 
  We improve correspondence with RGB-D  inputs, using metrics from~\cite{truong2023pdc}.
  The entry train and test are training and testing datasets of correspondence estimators.
  [Key: M=MegaDepth, S=ScanNet]
  }
  \resizebox{0.73\linewidth}{!}{%
  \begin{tabular}{lcc|ccccccccc}
    \hline
    \textbf{Method} &  \textbf{Train} & \textbf{Test}  & \cellcolor[RGB]{155, 187, 228} $\textbf{PCK-1}$ & \cellcolor[RGB]{155, 187, 228} $\textbf{PCK-3}$ & \cellcolor[RGB]{155, 187, 228} $\textbf{PCK-5}$ & \cellcolor[RGB]{222, 164, 151} $\textbf{AEPE}$ \\
    \hline
    PDC-Net~\cite{truong2023pdc}  & \multirow{3}{*}{M} & \multirow{3}{*}{S}   & $0.119$ & $0.511$ & $0.743$   & $4.612$   \\
    $\drsh$ LightedDepth~\cite{zhu2023lighteddepth}  &&& $0.061$ & $0.341$ & $0.563$ & $6.590$ \\
    $\drsh$ $\text{Ours}$ && & $\mathbf{0.178}$ & $\mathbf{0.658}$ & $\mathbf{0.866}$ & $\mathbf{2.898}$\\
    \hline
    RoMa~\cite{edstedt2023roma} & \multirow{3}{*}{S} & \multirow{3}{*}{S}   & $0.144$ & $0.583$ & $0.815$  & $3.333$ \\
    $\drsh$ LightedDepth~\cite{zhu2023lighteddepth}  && & $0.066$ & $0.359$ & $0.588$ & $5.974$\\
    $\drsh$ $\text{Ours}$  && & $\mathbf{0.183}$ & $\mathbf{0.638}$ & $\mathbf{0.844}$ & $\mathbf{3.067}$\\
    \bottomrule
  \end{tabular}
  }
  \label{tab:corres}
\end{table}

\subsection{Self-supervised Depth Estimation}

We benchmark whether self-supervision benefits supervised depth in unseen test data.
For the correspondence estimator, we use PDC-Net~\cite{truong2023pdc}.
For depth estimators, we adopt recently published in-the-wild depth estimator, including ZoeDepth~\cite{bhat2023zoedepth}, ZeroDepth~\cite{liu2023zero}, and Metric3D~\cite{yin2023metric3d}.
We evaluate with ScanNet~\cite{dai2017scannet} and KITTI360~\cite{liao2022kitti} where all models perform zero-shot prediction.

\Paragraph{Test Data.}
In dense correspondence estimation, methods~\cite{zhu2023pmatch, truong2023sparf, truong2023pdc} output confidence score per correspondence.
We follow~\cite{truong2023sparf, truong2023pdc} to set a minimum threshold of $0.95$.
We run on ScanNet test split and it returns $92$ sequences with sufficient correspondence.
We form our test split by sampling $5$ neighboring frames per valid sequence.
Similarly, we run on KITTI360 data and randomly select $100 \times 5$ test split, \textit{i.e.}, $100$ sequences with $5$ frames each.
We consider it a comprehensive experiment.
Similar to SPARF~\cite{truong2023sparf}, our triangulation trains a NeRF-like structure.
For reference, SPARF experiment on DTU dataset~\cite{jensen2014large} includes only $15$ sequences each with $3$ images.
In comparison, we include around $100$ sequences.

\Paragraph{Evaluation Protocols.}
We evaluate on \textbf{root} frame.
We remove the scale ambiguity in the local SfM system to correctly reflect depth improvement.
Specifically, we adjust all $5$ depthmaps by an identical scalar computed between estimated root and GT depthmap, \textit{i.e.}, the median scaling~\cite{gordon2019depth}.
This eliminates scale ambiguity in the root frame while preserving it in support frames.

\Paragraph{Results.}
In~\cref{tab:self_sup_depth}, our point cloud has a density of $2.6\% - 9.1\%$, which amounts to $\mathbf{10} - \mathbf{30}$k points on a $480 \times 640$ image.
On accuracy, we have \textbf{unanimous} improvement over all supervised models of both datasets.
Especially, we outperform strong baselines of ZoeDepth on ScanNet and Metric3D on KITTI360.


\label{sec:selfsuperviseddepth}


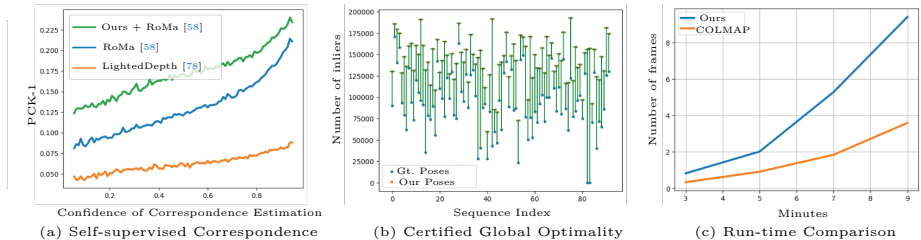
\begin{figure}[t!]
  \vspace{-1mm}
  \captionsetup{font=small}
  \centering
  \begin{tikzpicture}
  \draw (0, 0) node[inner sep=0] {\includegraphics[width=\linewidth]{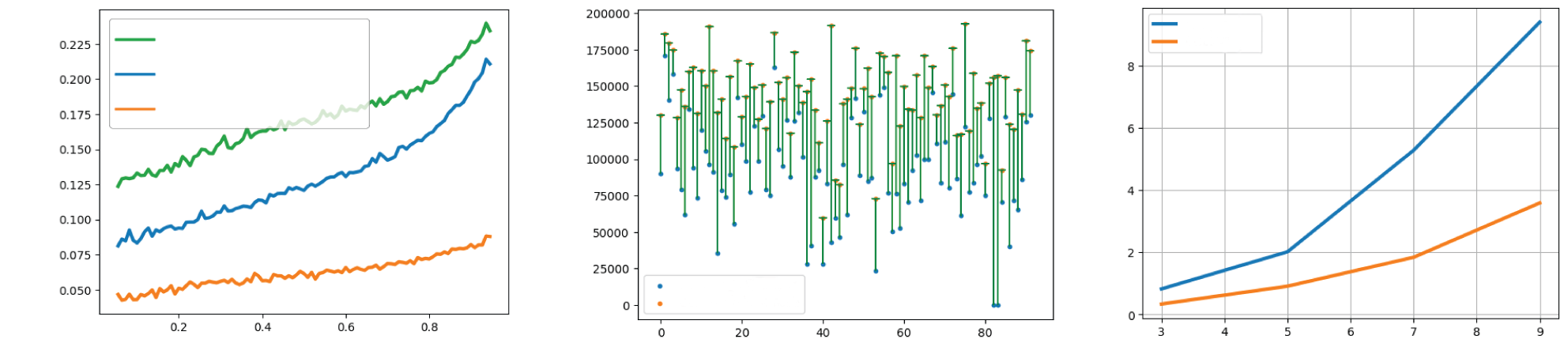}};
  \draw (-1.0 + 2.7 -5.85, 0.68 + 0.35) node[inner sep=0] {\fontsize{3.5}{10}\selectfont Ours +  RoMa~\cite{truong2023pdc}};
  \draw (-1.25 + 2.67 -5.85, 0.41 + 0.35) node[inner sep=0] {\fontsize{3.5}{10}\selectfont RoMa~\cite{truong2023pdc}};
  \draw (-1.1 + 2.85 -5.9, 0.15 + 0.35) node[inner sep=0] {\fontsize{3.5}{10}\selectfont LightedDepth~\cite{zhu2023lighteddepth}};
  \draw (-3.65, -1.45) node[inner sep=0] {\fontsize{4.0}{10}\selectfont Confidence of Correspondence Estimation};
  \draw (0.5, -1.45) node[inner sep=0] {\fontsize{4.0}{10}\selectfont Sequence Index};
  \draw (4.5, -1.45) node[inner sep=0] {\fontsize{4.0}{10}\selectfont Minutes };
  \draw (-5.75, 0.0) node[inner sep=0] {\fontsize{4.0}{10}\selectfont \rotatebox{90}{PCK-1} };
  \draw (-3.8, -1.5 -0.2) node[inner sep=0] {\tiny (a) Self-supervised Correspondence };
  \draw (0.45, -1.5 -0.2) node[inner sep=0] {\tiny (b) Certified Global Optimality };
  \draw (4.4, -1.5 -0.2) node[inner sep=0] {\tiny (c) Run-time Comparison};
  \draw (2.5, 0.2) node[inner sep=0] {\fontsize{4.0}{10}\selectfont \rotatebox{90}{Number of frames} };
  \draw (-1.7, 0.2) node[inner sep=0] {\fontsize{4.0}{10}\selectfont \rotatebox{90}{Number of inliers} };
  \draw (3.25 + 0.05, 1.15) node[inner sep=0] {\fontsize{3.5}{10}\selectfont Ours };
  \draw (3.39 + 0.05, 1.00) node[inner sep=0] {\fontsize{3.5}{10}\selectfont  COLMAP };
  \draw (-0.6+0.05, -0.85-0.03) node[inner sep=0] {\fontsize{3.5}{10}\selectfont Gt. Poses };
  \draw (-0.55+0.05, -1.0-0.03) node[inner sep=0] {\fontsize{3.5}{10}\selectfont  Our Poses };
  \end{tikzpicture}
  \vspace{-5.5mm}
  \caption{
  \small 
  \textbf{Ablation Studies} on the ScanNet.
  }
  \label{fig:corres_curve}
\end{figure}

\subsection{Consistent Depth Estimation}

We evaluate on ScanNet.
We follow \cref{sec:selfsuperviseddepth} data split but evaluate the \textbf{support} frames.
Temporal consistent depth is essential for AR applications~\cite{luo2020consistent}.
\cref{tab:consist_depth} reflects the performance gain by aligning support frames to root with adjustments, which are jointly estimated with camera poses, see \cref{fig:triang_depth} and \cref{fig:framework}.

\begin{figure*}[t!]
\vspace{-2mm}
  \captionsetup{font=small}
  \centering
  \begin{tikzpicture}
  \draw (0, 0) node[inner sep=0] {\includegraphics[width=\linewidth]{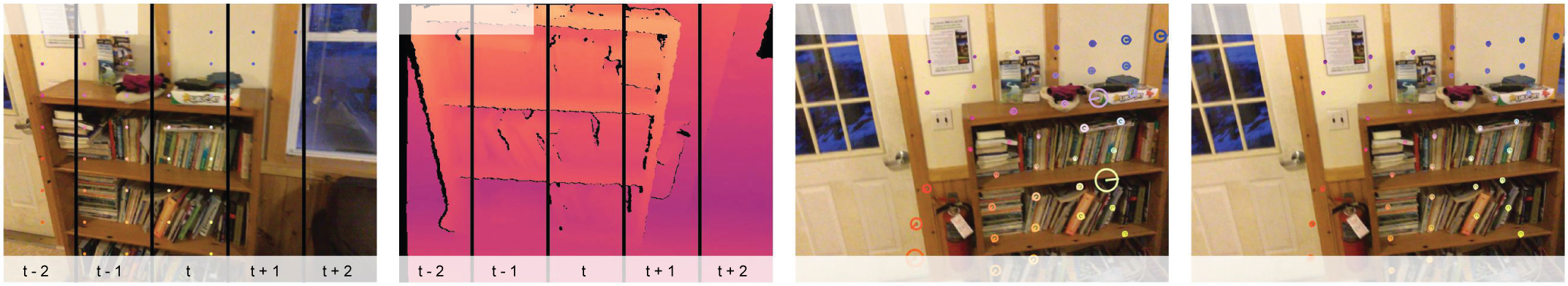}};
  \draw (-5.4, 0.95) node[inner sep=0] {\fontsize{4}{10}\selectfont RGB-D Inputs};
  \draw (-2.33, 0.95) node[inner sep=0] {\fontsize{4}{10}\selectfont RGB-D Inputs};
  \draw (0.6, 0.95) node[inner sep=0] {\fontsize{4}{10}\selectfont RoMa~\cite{edstedt2023roma} };
  \draw (1.4, -1.0) node[inner sep=0] {\fontsize{4}{10}\selectfont Correspondence from t to t+$1$};
  \draw (3.8, 0.95) node[inner sep=0] {\fontsize{4.0}{10}\selectfont RoMa + Ours };
  \draw (4.48, -1.0) node[inner sep=0] {\fontsize{4.0}{10}\selectfont Correspondence from  t to t+$1$};
  \end{tikzpicture}
    \vspace{-4mm}
    \caption{\small 
    \textbf{Self-supervised Correspondence Estimation} enabled by our method with RGB-D inputs. 
    The correspondence error is marked by the radius of the circle.
    }
    \vspace{-3mm}
    \label{fig:projective_corres}
\end{figure*}
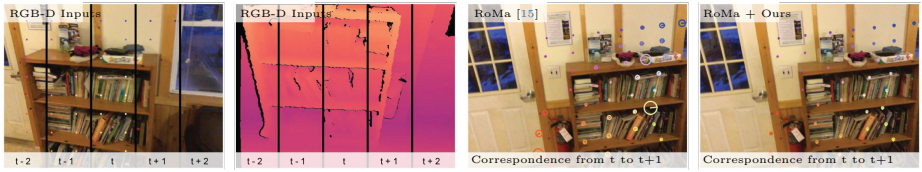

\subsection{Self-supervised Correspondence Estimation}
\label{sec:corres_exp}
Real-world image correspondence label is expensive, {\it e.g.}~KITTI provides only $200$ optical flow labels.
Existing datasets, such as MegaDepth and ScanNet, require large-scale 3D reconstruction with manual verification.
Hence, correspondence estimators can not fine-tune on general RGB-D datasets like NYUv2~\cite{silberman2012indoor} or KITTI~\cite{geiger2013vision}.
But our method enables self-supervised correspondence estimation on RGB-D data when using 3D scoring function \cref{eqn:loss_3D}.
The camera poses are optimized with the point cloud specified by depthmap and correspondence.
The accurate pose in turn improves projective correspondence.
In \cref{tab:corres}, with $5$ RGB-D frames, our method improves projective correspondence over inputs.
We use the same test split as \cref{sec:selfsuperviseddepth}.
The evaluation accumulates correspondence of each frame pair.
\cref{fig:corres_curve}a shows our improvement is \textbf{unanimous} over both confident and unconfident estimation.
A visual example is in \cref{fig:projective_corres}.

\subsection{Sparse-view Pose Estimation}
\Paragraph{Comparison with Optimization-based and Learning-based Poses.}
Previous studies either evaluate two-view pose~\cite{Teed2020DeepV2D, gu2023dro}, or SLAM-like odometry~\cite{wang2023dust3r}.
For more comparison, following \cref{sec:selfsuperviseddepth} ScanNet split, we keep root frame and gradually add neighboring frames.
In \cref{tab:scannet_camera_pose}, LightedDepth~\cite{zhu2023lighteddepth} and ours both use PDC-Net~\cite{truong2023pdc} correspondence and ZoeDepth~\cite{bhat2023zoedepth} mono-depth.
COLMAP~\cite{schonberger2016structure} uses PDC-Net correspondence.
In evaluation, we follow \cite{truong2023sparf} in aligning to GT poses.
In \cref{tab:scannet_camera_pose}, our \textbf{zero-shot} pose accuracy significantly outperforms all prior arts, including \cite{wang2023dust3r, gu2023dro, Teed2020DeepV2D} with ScanNet~\cite{dai2017scannet} or ScanNet++~\cite{yeshwanth2023scannet++} in their training set.
See Supp. Tab. $\color{red}{4}$ for complete comparison from $3$ to $9$ frames.
In \cref{fig:corres_curve}, we attribute our superiority to certified global optimality over robust measurements. 

\begin{table}[t!]
  \centering
  \captionsetup{font=small}
  \caption{
  \small
  \textbf{Sparse-view Pose Comparison} with optimization-based and learning-based methods. 
  We only compare against COLMAP on its success sequences.
  Please see the complete comparison from $3$ to $9$ frames in Supp. Tab. $\color{red}{1}$.
  Our method performs \textbf{zero-shot} testing on ScanNet while outperforming DeepV2D~\cite{Teed2020DeepV2D}, DRO~\cite{gu2023dro} with ScaNet~\cite{dai2017scannet} in training set.
  DUSt3R~\cite{wang2023dust3r} trains on a similar dataset ScanNet++~\cite{yeshwanth2023scannet++}.
  }
  \resizebox{0.9\linewidth}{!}{%
  \begin{tabular}{l|l|c|c|ccccccc}
    \hline
    \textbf{Frames} & \textbf{Method} & \textbf{Zero-shot} & \cellcolor[RGB]{155, 187, 228} $\textbf{Suc. (\%)}$ & \cellcolor[RGB]{155, 187, 228} $\textbf{PCK-3}$  & \cellcolor[RGB]{155, 187, 228} $\textbf{C3D-3}$ & \cellcolor[RGB]{222, 164, 151} $\textbf{Rot.}$ & \cellcolor[RGB]{222, 164, 151} $\textbf{Trans.}$ \\
    \hline
    \multirow{11}{*}{5} & COLMAP~\cite{schonberger2016structure} & \textcolor{ForestGreen}{\ding{51}} & $36.7$& $0.584$  & $0.863$  & $0.577$ & $1.296$  \\
    & Ours & \textcolor{ForestGreen}{\ding{51}} & ${100.0}$  & $\mathbf{0.727}$ & $\mathbf{0.904}$ & $\mathbf{0.422}$ & $\mathbf{1.062}$  \\
    \cdashline{2-8}
    & DeepV2D~\cite{Teed2020DeepV2D} - ScanNet & \textcolor{RedOrange}{\ding{55}} & \multirow{9}{*}{100.0} & $0.526$ & $0.805$ & $0.945$ & $1.496$ \\
    & DeepV2D~\cite{Teed2020DeepV2D} - NYUv2& \textcolor{ForestGreen}{\ding{51}}  & & $0.530$ & $0.771$ & $1.041$ & $1.568$\\
    & DeepV2D~\cite{Teed2020DeepV2D} - KITTI& \textcolor{ForestGreen}{\ding{51}}  & & $0.125$ & $0.387$ & $4.908$ & $4.231$\\
    & LightedDepth~\cite{zhu2023lighteddepth} & \textcolor{ForestGreen}{\ding{51}}  & & $0.651$ & $0.832$ & $ 0.469$ & $1.550$ \\
    & DRO~\cite{gu2023dro} - ScanNet & \textcolor{RedOrange}{\ding{55}} & & $0.656$ & $0.853$ & $0.385$ & $1.200$ \\
    & DRO~\cite{gu2023dro} - KITTI& \textcolor{ForestGreen}{\ding{51}}  & & $0.003$ & $0.211$ & $3.610$ & $5.469$ \\
    & DUSt3R~\cite{wang2023dust3r} \textit{w.o.} Intrinsic& \textcolor{ForestGreen}{\ding{51}}  & & $0.364$ & $0.705$   & $0.487$ & $2.074$ \\
    & DUSt3R~\cite{wang2023dust3r} \textit{w.t.} Intrinsic & \textcolor{ForestGreen}{\ding{51}} & & $0.594$ & $0.824$ & $0.570$ & $1.759$ \\
    & Ours & \textcolor{ForestGreen}{\ding{51}} &  & $\mathbf{0.799}$ & $\mathbf{0.900}$ & $\mathbf{0.368}$ & $ \mathbf{1.120}$\\ 
    \bottomrule
  \end{tabular}
  }
  \label{tab:scannet_camera_pose}
\end{table}

\begin{table}[t!]
  \centering
  \captionsetup{font=small}
  \caption{
  \small
  \textbf{Sparse-view Pose Comparison} with NeRF-based methods following \cite{truong2023sparf}. 
  }
  \resizebox{0.6\linewidth}{!}{%
  \begin{tabular}{l|c|cc|cc}
    \hline
    \multirow{2}{*}{\textbf{Method}}& \multirow{2}{*}{\textbf{Frames}}&  \multicolumn{2}{c|}{LLFF~\cite{shafiei2021learning}} &  \multicolumn{2}{c}{Replica~\cite{replica19arxiv}} \\
     &  & \cellcolor[RGB]{222, 164, 151} $\textbf{Rot.}$ & \cellcolor[RGB]{222, 164, 151} $\textbf{Trans.}$ & \cellcolor[RGB]{222, 164, 151} $\textbf{Rot.}$ & \cellcolor[RGB]{222, 164, 151} $\textbf{Trans.}$\\
    \hline
    BARF~\cite{lin2021barf} & \multirow{6}{*}{3}  & $2.04$ & $11.6$ & $3.35$ & $16.96$\\
    RegBARF~\cite{lin2021barf, niemeyer2022regnerf} & & $1.52$ & $5.0$ & $3.66$ & $20.87$\\
    DistBARF~\cite{lin2021barf, barron2022mip} & & $5.59$ & $26.5$ & $2.36$ & $7.73$\\
    SCNeRF~\cite{jeong2021self} & & $1.93$ & $11.4$ & ${0.65}$ & ${4.12}$\\
    SPARF~\cite{truong2023sparf} & & $\underline{0.53}$ & $\underline{2.8}$ & $\textbf{0.15}$ & $\textbf{0.76}$ \\
    Ours  & & $\textbf{0.46}$ & $\textbf{1.9}$ & $\underline{0.52}$ & $\underline{4.09}$\\
    \bottomrule
  \end{tabular}
  }
  \vspace{-2mm}
  \label{tab:sparse_render}
\end{table}

\Paragraph{Comparison with NeRF-based Poses.}
Sparse view NeRF methods optimize NeRF jointly with camera poses, mandating a sophisticated and time-consuming optimization scheme.
\textit{E.g.}, SPARF~\cite{truong2023sparf}, takes one day to optimize the pose and NeRF.
Typically, their poses are initialized with COLMAP.
Our method provides an alternative initialization with superior performance.
In \cref{tab:sparse_render}, our initialization achieves better or on-par pose performance than SoTA~\cite{truong2023sparf} while only taking $\sim$$3$ minutes (\cref{fig:ablation}).
Our lower performance on Replica dataset might be due to ZoeDepth not being trained on synthetic data.
Our work suggests the straightforward ``first-pose-then-NeRF'' scheme also applies to short videos.




\Paragraph{Certified Global Optimality}.
In \cref{fig:corres_curve}b, our Bundle-RANSAC-Adjustment \textbf{always} finds more inliers than groundtruth poses.
To our best knowledge, we are the \textbf{first} work that extends RANSAC to a multi-view system.

\Paragraph{Run-time.}
In \cref{fig:corres_curve}c, we run approximately $3\times$  slower than COLMAP.
But both have quadratic complexity.
With $3/5/7/9$ frames, we take $0.8/2.0/5.3/9.4$ minutes on RTX 2080 Ti GPU, while
COLMAP uses $0.3/0.9/1.8/3.6$ minutes on Intel Xeon 4216 CPU.
COLMAP runs sequentially.
But our method is highly parallelized. 
Our core operation Hougn Transform scales up with more GPUs.

\section{Conclusion}
By revisiting self-supervision with local SfM, we first show self-supervised depth benefits SoTA supervised model with only $5$ frames.
We have SoTA sparse-view pose accuracy, applicable to NeRF rendering.
We have diverse applications including self-supervised correspondence and consistent depth estimation.

\Paragraph{Limitation.}
The NeRF-like triangulation constrains our method from applying to large-scale self-supervised learning.
Its efficiency requires improvement.

\clearpage

\setcounter{page}{1}
\title{======Supplementary======\\ Revisit Self-supervised Depth Estimation with Local Structure-from-Motion}

\author{Shengjie Zhu  \and Xiaoming Liu}

\institute{
Department of Computer Science and Engineering,\\
Michigan State University, East Lansing, MI, 48824 \\
\email{zhusheng@msu.edu}, \email{liuxm@cse.msu.edu}
}

\maketitle

\setcounter{section}{0}

\section{Extended Methodology}


\subsection{Hough Transform with 3D Scoring Function}
In \cref{sec:corres_exp} and \cref{fig:projective_corres}, with RGB-D inputs, we utilize the 3D scoring function \cref{eqn:loss_3D}. 
For a pixel $\mathbf{p}_i$ on frame $i$, denote its backprojected 3D ray on the image coordinate system of frame $j$ as $\hat{\mathbf{l}}_i$. 
The ray $\hat{\mathbf{l}}_i$ is determined by two 3D points:
\begin{equation}
    \hat{\mathbf{p}}_1 = \mathbf{R}_{i, j}  \mathbf{K}\inv \mathbf{p}_i + \mathbf{t}_{i, j}, \quad \hat{\mathbf{p}}_2 = \mathbf{t}_{i, j}.
\end{equation}
The depth of 3D point $\hat{\mathbf{p}}_1$ can be set to arbitrary values. 
We set it to $1$ for convenience. 
3D point $\hat{\mathbf{p}}_2$ is the frame $i$ camera origin on frame $j$.
Correspondingly, the backprojected 3D point $\hat{\mathbf{p}}_{\pi}$ is:
\begin{equation}
  \hat{\mathbf{l}}_i^\intercal \hat{\mathbf{p}}_{\pi} = 0, \;  \hat{\mathbf{p}}_{\pi} = \pi^{-1} (s_i, s_j, r_i \mid \overline{\mathbf{P}}_i, \overline{\mathbf{P}}_j, d_i).
\end{equation}
To be an inlier of the scoring function $f^\text{3D}(\cdot)$, we have:
\begin{equation}
\small
    \| \hat{\mathbf{p}}_{\pi} - \hat{\mathbf{p}}_j\|_2 \leq \lambda^{\text{3D}}, \; \hat{\mathbf{p}}_{j} = \pi^{-1} (s_i, s_j, r_j \mid \overline{\mathbf{P}}_i, \overline{\mathbf{P}}_j, d_j).
    \label{eqn:sphere}
\end{equation}
\cref{eqn:sphere} suggests a 3D line and sphere intersection, with the two end-points denoted as $\hat{\mathbf{p}}_{\pi}^{\text{st}}$ and $\hat{\mathbf{p}}_{\pi}^{\text{ed}}$.
\begin{figure}[t!]
  \captionsetup{font=small}
  \centering
  \begin{tikzpicture}
  \draw (0, 0) node[inner sep=0] {\includegraphics[width=\linewidth]{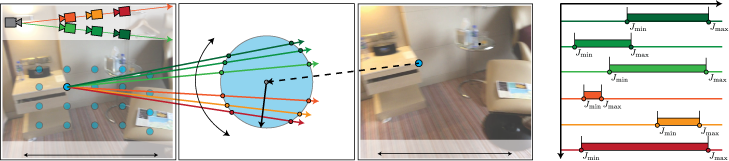}};
  \draw (4.7, 1.45) node[inner sep=0] {\fontsize{4.0}{10}\selectfont $x = g(r_i, r_j, s_i, s_j \mid \cdot)$};
  \draw (3.1, 0.0) node[inner sep=0] {\fontsize{4.0}{10}\selectfont \rotatebox{90}{$y = \arccos(\overline{\mathbf{t}}_{i, j}^\intercal \overline{\mathbf{t}}_{j})$}};
  \draw (-5.3, 1.0+0.15-0.05) node[inner sep=0] {\fontsize{4.0}{10}\selectfont $\mathbf{P}_i$};
  \draw (-5.3, 1.0-0.15-0.05) node[inner sep=0] {\fontsize{4.0}{10}\selectfont $\mathbf{P}_j$};
  \draw (-3.4, 0.95) node[inner sep=0] {\fontsize{4.0}{10}\selectfont \rotatebox{-90}{Infty}};
  \draw (-4.6-1.1, -1.15) node[inner sep=0] {\fontsize{4.0}{10}\selectfont $0$};
  \draw (-4.6+1.15, -1.15) node[inner sep=0] {\fontsize{4.0}{10}\selectfont +$\infty$};
  \draw (-4.6, -1.15) node[inner sep=0] {\fontsize{4.0}{10}\selectfont Depth Adjustment $r_i$};
  \draw (1.35, -1.12) node[inner sep=0] {\fontsize{4.0}{10}\selectfont Depth Adjustment $r_j$};
  \draw (1.45-1.2, -1.12) node[inner sep=0] {\fontsize{4.0}{10}\selectfont $0$};
  \draw (1.45+1.0, -1.12) node[inner sep=0] {\fontsize{4.0}{10}\selectfont +$\infty$};
  \draw (-5.05, 0.05) node[inner sep=0] {\fontsize{4.0}{10}\selectfont $\mathbf{p}_i$};
  \draw (1.0, 0.1) node[inner sep=0] {\fontsize{4.0}{10}\selectfont $\mathbf{p}_j$};
  \draw (-1.4, -0.2) node[inner sep=0] {\fontsize{4.0}{10}\selectfont $\lambda^{\text{3D}}$};
  \draw (-1.1-1.4, 0.9) node[inner sep=0] {\fontsize{4.0}{10}\selectfont $\underset{s_i \rightarrow +\inf}{\lim}$};
  \draw (-1.1-1.0, -1.15) node[inner sep=0] {\fontsize{4.0}{10}\selectfont $\underset{s_j \rightarrow +\inf}{\lim}$};
  \draw (-4.7, -0.3) node[inner sep=0] {\fontsize{4.0}{10}\selectfont $\{\hat{\mathbf{l}}_i\}$};
  \draw (-2.0, 0.6) node[inner sep=0] {\fontsize{4.0}{10}\selectfont $\hat{\mathbf{p}}_{\pi}^{\text{st}}$};
  \draw (-1.15, 0.8) node[inner sep=0] {\fontsize{4.0}{10}\selectfont $\hat{\mathbf{p}}_{\pi}^{\text{ed}}$};
  \draw (3.5, 1.12) node[inner sep=0] {\fontsize{4.0}{10}\selectfont $\mathbf{H}_{i, j}^m$};
  \draw (-0.9, 1.1) node[inner sep=0] {\fontsize{4.0}{10}\selectfont Backprj. to 3D};
  \end{tikzpicture}
  \vspace{-4mm}
  \caption{
  \small 
  \textbf{Two-view Hough Transform on 3D Scoring Function.} 
  Pixels $\mathbf{p}_i$ and $\mathbf{p}_j$ are corresponded. 
  Similar to \cref{fig:hough}, ablating pose scales map pixel $\mathbf{p}_i$ to a set of 3D rays originated from camera origin, denoted as $\{\hat{\mathbf{l}}_i\}$.
  To be an inlier, a backprojected 3D point $\hat{\mathbf{p}}_{\pi}$ has to reside within a sphere centered with $\hat{\mathbf{p}}_j$ with a radius $\lambda^{\text{3D}}$, \textit{i.e.}, between 3D segments $\hat{\mathbf{p}}_{\pi}^{\text{st}}$ and $\hat{\mathbf{p}}_{\pi}^{\text{ed}}$.
  Different from \cref{fig:hough}, with fixed normalized poses, there exists \textbf{four} variables to optimize, including the additional frame $j$ depth adjustment $r_j$.
  }
  \label{fig:hough_3D_supp}
\end{figure}
Compare 2D intersection condition \cref{eqn:circle} with 3D intersection condition \cref{eqn:sphere}, we find \cref{eqn:sphere} further relates to the depth adjustment $r_j$ on frame $j$.
Similarly, compare \cref{fig:hough} with \cref{fig:hough_3D_supp}, the circle center of 2D scoring function is independent of depth adjustments $r_j$.
The sphere center of the 3D scoring function, however, changes \textit{w.r.t.}~depth adjustments.
This introduces one additional variable in Hough Transform, \textit{i.e.}, the depth adjustment $r_j$ at frame $j$, also shown in \cref{fig:hough_3D_supp}.
The additional dimensionality creates an excessively high time and space complexity excessively for Hough Transform.
Hence, we are unable to optimize depth adjustment in 3D scoring function.
One viable solution is to first execute with 2D scoring function, fix the adjustments, and then employ the 3D scoring function.
We did not experiment with this strategy since we already had competitive performance.
When with RGB-D inputs, we use 3D scoring function as groundtruth depth does not require any scale adjustment. 
The rest is similar to \cref{sec:hough_ba_sec} and skipped.

\subsection{Proof of \cref{eqn:lim}.}
From \cref{eqn:two_view}, the relative pose translation magnitude $s_{i, j}$ is:
\begin{equation}
s_{i, j}^2 = s_i^2 + s_j^2 + k_{\theta} \cdot s_i s_j , \; k_{\theta} = 2\overline{\mathbf{t}}_j^\intercal \mathbf{R}_j \mathbf{R}_i^{\text{-}1} \overline{\mathbf{t}}_i.
\end{equation}
The relative pose normalized translation vector $\overline{\mathbf{t}}_{i, j}$ is:
\begin{equation}
    \overline{\mathbf{t}}_{i, j} = -s_i / s_{i, j} \mathbf{R}_j \mathbf{R}_i^{\text{-}1} \overline{\mathbf{t}}_i + s_j / s_{i, j} \overline{\mathbf{t}}_j.
\end{equation}
Illustrate with $s_i$, when $s_i \rightarrow +\infty$, we have:
\begin{equation}
    \lim_{s_i \rightarrow +\inf}  s_i / s_{i, j} = 1, \; \lim_{s_i \rightarrow +\inf}  s_j / s_{i, j} = 0.
\end{equation}
Combined, we have:
\begin{equation}
    \lim_{s_i \rightarrow +\inf} \overline{\mathbf{t}}_{i, j} = -\mathbf{R}_j \mathbf{R}_i^{\text{-}1} \overline{\mathbf{t}}_i.
\end{equation}
The other case of \cref{eqn:lim} follows a similar logic.

\subsection{Proof of \cref{eqn:collary1} and \cref{eqn:collary2}.}
\paragraph{Notations.} For simplicity, the relative pose $\mathbf{P}_{i, j}$ between frame $i$ and $j$ is defined as $\mathbf{P}_{i, j} = \begin{bmatrix}
    \mathbf{R} & s \cdot \overline{\mathbf{t}}
\end{bmatrix}.$
Denote the 2D pixel locations on frame $i$ as $\mathbf{p}$. 
Its projection is set to $\mathbf{q}$, \textit{i.e.}, the variable $\mathbf{p}_\pi$ in \cref{sec:hough_ba_sec}.
The projection is:
\begin{equation}
    d' \mathbf{q}=d' 
    \begin{bmatrix}
    q_x & q_y & 1
    \end{bmatrix}^\intercal 
    = r\cdot d \mathbf{K} \mathbf{R} \mathbf{K}^{\text{-}1} \mathbf{p}+ s \mathbf{K} \overline{\mathbf{t}}.
\end{equation}
Variable $r$ is the depth adjustment on frame $i$. 
Variable $d$ is the depth of pixel $\mathbf{p}$.
The pixel location $q_x$ and $q_y$ are:
\begin{equation}
\small
    q_x = \frac{r \cdot d \mathbf{m}_1^\intercal \mathbf{p} + s\cdot x}{r \cdot d \mathbf{m}_3^\intercal \mathbf{p} + s\cdot z}, \;  q_y = \frac{r \cdot d \mathbf{m}_2^\intercal \mathbf{p} + s\cdot y}{r \cdot d \mathbf{m}_3^\intercal \mathbf{p} + s\cdot z}.
    \label{eqn:prjprocess}
\end{equation}
The remaining variables are defined as follows:
\begin{equation}
\small
    \begin{bmatrix}
    \mathbf{m}_1 & \mathbf{m}_2 & \mathbf{m}_3
    \end{bmatrix}^\intercal =  \mathbf{K} \mathbf{R} \mathbf{K}^{\text{-}1}, \;
    s \cdot \begin{bmatrix}
        x & y & z
    \end{bmatrix}^\intercal =  s \cdot \mathbf{K} \overline{\mathbf{t}}.
\end{equation}
Here, we slightly abuse the notation of $x$ and $y$ without implying the Hough Transform locations.
The mapping function $J(\cdot)$ from projection to camera scale is hence:
\begin{equation}
    s = J(\overline{\mathbf{P}}, r\cdot d, \mathbf{q}) = r \cdot \frac{ d (\mathbf{m}_1^\intercal \mathbf{p} - q_x\mathbf{m}_3^\intercal \mathbf{p})}{z \cdot q_x - x}.
    \label{eqn:funcj}
\end{equation}
There exists another analytical mapping from the axis-$y$ direction.
We skip this for simplicity.

\Paragraph{Proof of Corollary \ref{cor:corollary2}.} From \cref{eqn:funcj}, it is obvious that:
\begin{equation}
    s = J(\overline{\mathbf{P}}, r\cdot d, \mathbf{q}) = r\cdot J(\overline{\mathbf{P}}, d, \mathbf{q}).
\end{equation}

\Paragraph{Proof of Corollary \ref{cor:corollary1}.}
We decompose the relative camera pose $\mathbf{P}$ from frame $i$ and $j$ into a rotation movement $\mathbf{P}^r$ followed by a translation movement $\mathbf{P}^t$:
\begin{equation}
\small
    \mathbf{P}^r = \begin{bmatrix}
    \mathbf{R} & \mathbf{0}
\end{bmatrix}, \;
    \mathbf{P}^t = \begin{bmatrix}
    \mathbf{E} & s \cdot \mathbf{R} \overline{\mathbf{t}}
\end{bmatrix}, \;
\mathbf{P} = \mathbf{P}^t \mathbf{P}^r.
\end{equation}
Correspondingly, we introduce an intermediate pixel $\mathbf{q}^r$ as a consequence of the pure rotation movement.
That is to say, between pixels $\mathbf{p}$ and $\mathbf{q}^r$ are a pure rotation movement. 
And between pixels $\mathbf{q}^r$ and $\mathbf{q}$ are a pure translation movement.
The pixels $\mathbf{q}^r$ and $\mathbf{q}$ have the following relationship:
\begin{equation}
    \mathbf{q}^r = \lim_{s \rightarrow 0} \mathbf{q} = \lim_{r\cdot d \rightarrow +\infty} \mathbf{q}.
    \label{eqn:pure_rot}
\end{equation}
\cref{eqn:pure_rot} is obvious. 
We skip its proof.
We introduce it to illustrate two facts.
First, from \cref{fig:hough}, the intersection point of all epipolar lines $\{\mathbf{l}_i\}$ is $\mathbf{q}^r$, \textit{i.e.}, the projected pixel from pure rotation.
Second, the pixel $\mathbf{q}^r$ and $\mathbf{q}$ formulates a pure translation movement. 
It simplifies our proof if we replace the pixel $\mathbf{p}$ to pixel $\mathbf{q}^r$ in \cref{eqn:prjprocess}.
Due to a pure translation movement, the $ \mathbf{K} \mathbf{R} \mathbf{K}^{\text{-}1}$ is an identity matrix.
We have:
\begin{equation}
    q_x = \frac{r \cdot d^r \cdot q_x^r + s \cdot x}{r \cdot d^r + s \cdot z}, \; q_y = \frac{r \cdot d^r \cdot q_y^r + s \cdot y}{r \cdot d^r + s \cdot z}.
\end{equation}
We compute the squared magnitude of the vector $\|\overrightarrow{\mathbf{q}^r \mathbf{q}}\|_2^2$:
\begin{equation}
\begin{aligned}
    \|\overrightarrow{\mathbf{q}^r \mathbf{q}}\|_2^2 &= (q_x -q_x^r)^2 + (q_y - q_y^r)^2 \\
    &=\frac{(s\cdot x)^2 + (s\cdot y)^2}{(r \cdot d^r + s \cdot z)^2} = \frac{x^2 +  y^2}{(r \cdot d^r / s + z)^2}.
    \label{eqn:mag}
\end{aligned}
\end{equation}
Meanwhile, we underlyingly require the pixel $\mathbf{q}$ to be visible in both frames.
This requires a positive depth:
\begin{equation}
    r \cdot d^r / s + z > 0.
    \label{eqn:mag_pos}
\end{equation}
Combining \cref{eqn:mag} and \cref{eqn:mag_pos}, we conclude that the vector $\|\overrightarrow{\mathbf{q}^r \mathbf{q}}\|_2^2$ monotonously increases as translation magnitude $s$ increases.
It fits the intuition that the projected pixel $\mathbf{q}$ moves further as the camera translation magnitude increases.
To prove the corollary \ref{cor:corollary1}, we have:
\begin{equation}
   \| \overrightarrow{\mathbf{q}^r \mathbf{q}^{\text{st}}}\|^2_2 \leq \|\overrightarrow{\mathbf{q}^r \mathbf{q}} \|_2^2\leq \|\overrightarrow{\mathbf{q}^r \mathbf{q}^{\text{ed}}}\|_2^2.
\end{equation}
The $s$ monotonously increases with vector magnitude, then:
\begin{equation}
   J(\overline{\mathbf{P}}, r\cdot d, \mathbf{q}^{\text{st}}) \leq J(\overline{\mathbf{P}}, r\cdot d, \mathbf{q}) \leq J(\overline{\mathbf{P}}, r\cdot d, \mathbf{q}^\text{ed}).
\end{equation}
We apologize for the abuse of the notation.
The variable $\mathbf{q}^\text{ed}$ refers to $\mathbf{p}_\pi$ in \ref{sec:hough_ba_sec}.
We next address some corner cases unattended in main paper due to space issues.
\begin{itemize}
    \item The Epipolar line and the circle do not intersect: \\
    We exclude Hough Transform under this condition since there are no inlier projected pixels.
    \item \cref{eqn:mag_pos} does not hold:\\
    To be visible in both views, \cref{eqn:mag_pos} has to hold. Hence it provides an additional bound to $J_{\text{min}}$ and $J_{\text{max}}$, if needed.
\end{itemize}

\Paragraph{Compute Start and End Intersected Points.}
To compute $\mathbf{p}_\pi^{\text{st}}$ and $\mathbf{p}_\pi^{\text{ed}}$ in Sec.~\ref{sec:hough_ba_sec}, we follow the analytical line-circle intersection solution.

\begin{table}[t!]
  \centering
  \captionsetup{font=small}
  \caption{
  \small
  \textbf{Extended Sparse-view Pose Comparison} with optimization-based and learning-based methods on ScanNet.
  We extend main paper \cref{tab:scannet_camera_pose} across $3$ to  $9$ frames.
  }
  \resizebox{0.85\linewidth}{!}{%
  \begin{tabular}{l|l|c|ccccccc}
    \hline
    \textbf{Frames} & \textbf{Method} & \textbf{Zero-shot} & \cellcolor[RGB]{155, 187, 228} $\textbf{Suc. (\%)}$ & \cellcolor[RGB]{155, 187, 228} $\textbf{PCK-3}$  & \cellcolor[RGB]{155, 187, 228} $\textbf{C3D-3}$ & \cellcolor[RGB]{222, 164, 151} $\textbf{Rot.}$ & \cellcolor[RGB]{222, 164, 151} $\textbf{Trans.}$ \\
    \hline
    \multirow{10}{*}{3} & COLMAP~\cite{schonberger2016structure} & \textcolor{ForestGreen}{\ding{51}} &  $10.0$ & $0.379$ & $0.686$ & $1.055$ & $1.765$\\
    & Ours & \textcolor{ForestGreen}{\ding{51}}  & ${100.0}$ & $\mathbf{0.750}$ & $\mathbf{0.890}$ & $\mathbf{0.210}$ & $\mathbf{0.624}$\\
    \cdashline{2-8}
    & DeepV2D~\cite{Teed2020DeepV2D} - ScanNet & \textcolor{RedOrange}{\ding{55}} & \multirow{8}{*}{100.0} & $0.591$ & $0.863$  & $0.319$ & $0.907$ \\
    & DeepV2D~\cite{Teed2020DeepV2D} - NYUv2 & \textcolor{ForestGreen}{\ding{51}} & & $0.619$ & $0.839$ & $0.381$ & $0.946$ \\
    & LightedDepth~\cite{zhu2023lighteddepth} & \textcolor{ForestGreen}{\ding{51}} & & $0.742$ & $0.874$ & $0.380$ & $1.127$ \\
    & DRO~\cite{gu2023dro} - ScanNet & \textcolor{RedOrange}{\ding{55}} & & $0.757$ & $ 0.883$ & $0.368$  &  $0.881$\\
    & DRO~\cite{gu2023dro} - KITTI & \textcolor{ForestGreen}{\ding{51}} & & $0.006$ & $0.258$ & $2.043$ & $3.435$ \\
    & DUSt3R~\cite{wang2023dust3r} \textit{w.o.} Intrinsic & \textcolor{ForestGreen}{\ding{51}}  & & $0.409$ & $0.770$ & $0.356$ & $1.355$\\
    & DUSt3R~\cite{wang2023dust3r} \textit{w.t.} Intrinsic & \textcolor{ForestGreen}{\ding{51}} & & $0.574$ & $0.827$ & $0.467$ & $1.245$\\
    & Ours & \textcolor{ForestGreen}{\ding{51}} & & $\mathbf{0.858}$ & $\mathbf{0.918}$ & $\mathbf{0.275}$ & $\mathbf{0.820}$\\
    \hline
    \multirow{12}{*}{5} & COLMAP~\cite{schonberger2016structure} & \textcolor{ForestGreen}{\ding{51}} & $36.7$& $0.584$  & $0.863$  & $0.577$ & $1.296$  \\
    & Ours & \textcolor{ForestGreen}{\ding{51}} & \multirow{1}{*}{100.0}  & ${0.727}$ & ${0.904}$ & ${0.422}$ & ${1.062}$  \\
    \cdashline{2-8}
    & DeepV2D~\cite{Teed2020DeepV2D} - ScanNet & \textcolor{RedOrange}{\ding{55}} & \multirow{10}{*}{100.0} & $0.526$ & $0.805$ & $0.945$ & $1.496$ \\
    & DeepV2D~\cite{Teed2020DeepV2D} - NYUv2 & \textcolor{ForestGreen}{\ding{51}} & & $0.530$ & $0.771$ & $1.041$ & $1.568$\\
    & DeepV2D~\cite{Teed2020DeepV2D} - KITTI & \textcolor{ForestGreen}{\ding{51}} & & $0.125$ & $0.387$ & $4.908$ & $4.231$\\
    & LightedDepth~\cite{zhu2023lighteddepth} & \textcolor{RedOrange}{\ding{55}} & & $0.651$ & $0.832$ & $ 0.469$ & $1.550$ \\
    & DRO~\cite{gu2023dro} - ScanNet & \textcolor{RedOrange}{\ding{55}} & & $0.656$ & $0.853$ & $0.385$ & $1.200$ \\
    & DRO~\cite{gu2023dro} - KITTI & \textcolor{ForestGreen}{\ding{51}} & & $0.003$ & $0.211$ & $3.610$ & $5.469$ \\
    & DUSt3R~\cite{wang2023dust3r} \textit{w.o.} Intrinsic & \textcolor{ForestGreen}{\ding{51}} & & $0.364$ & $0.705$   & $0.487$ & $2.074$ \\
    & DUSt3R~\cite{wang2023dust3r} \textit{w.t.} Intrinsic & \textcolor{ForestGreen}{\ding{51}} & & $0.594$ & $0.824$ & $0.570$ & $1.759$ \\
    & Ours & \textcolor{ForestGreen}{\ding{51}} &  & $\mathbf{0.799}$ & $\mathbf{0.900}$ & $\mathbf{0.368}$ & $ \mathbf{1.120}$ \\
    \hline
    \multirow{10}{*}{7} &  COLMAP~\cite{schonberger2016structure} & \textcolor{ForestGreen}{\ding{51}}  & $70.6$  & $0.571$ & $0.852$ &  $0.596$&$1.521$ \\
    & Ours  & \textcolor{ForestGreen}{\ding{51}} & $100.0$  & $\mathbf{0.723}$ & $\mathbf{0.881}$ & $\mathbf{0.470}$ & $\mathbf{1.383}$ \\
    \cdashline{2-8}
    & DeepV2D~\cite{Teed2020DeepV2D} - ScanNet & \textcolor{RedOrange}{\ding{55}} & \multirow{8}{*}{100.0} & $0.395$ & $0.667$ & $1.719$ & $2.673$ \\
    & DeepV2D~\cite{Teed2020DeepV2D} - NYUv2 & \textcolor{ForestGreen}{\ding{51}} & & $0.434$ & $0.674$ & $1.830$ & $2.740$ \\
    & LightedDepth~\cite{zhu2023lighteddepth} & \textcolor{ForestGreen}{\ding{51}} & & $0.560$ & $0.801$ & $ 0.551 $ & $1.909$ \\
    & DRO~\cite{gu2023dro} - ScanNet & \textcolor{RedOrange}{\ding{55}} & & $0.567$ & $0.790$ & $0.619$ & $1.744$  \\
    & DRO~\cite{gu2023dro} - KITTI & \textcolor{ForestGreen}{\ding{51}} && $0.002$ & $0.191$ & $5.040$ & $7.375$\\
    & DUSt3R~\cite{wang2023dust3r} \textit{w.o.} Intrinsic & \textcolor{ForestGreen}{\ding{51}} & & $ 0.343$ & $0.659$ & $0.558$ & $ 2.428$\\
    & DUSt3R~\cite{wang2023dust3r} \textit{w.t.} Intrinsic & \textcolor{ForestGreen}{\ding{51}} & & $0.558$ & $0.804$ & $0.561$ & $2.017$ \\
    & Ours & \textcolor{ForestGreen}{\ding{51}} & & $\mathbf{0.739}$ & $\mathbf{0.873}$ & $\mathbf{0.458}$ & $\mathbf{1.428}$\\
    \hline
    \multirow{10}{*}{9} & COLMAP~\cite{schonberger2016structure} & \textcolor{ForestGreen}{\ding{51}} &  $86.7$  & $0.558$ & $0.812$ & $0.756$ & $2.281$  \\
    & Ours & \textcolor{ForestGreen}{\ding{51}} & ${100.0}$& $\mathbf{0.687}$ & $\mathbf{0.852}$   & $\mathbf{0.538}$ & $\mathbf{1.675}$  \\
    \cdashline{2-8}
    & DeepV2D~\cite{Teed2020DeepV2D} - ScanNet & \textcolor{RedOrange}{\ding{55}} & \multirow{8}{*}{100.0} & $0.303$ & $0.569$ & $4.279$ & $4.573$\\
    & DeepV2D~\cite{Teed2020DeepV2D} - NYUv2 & \textcolor{ForestGreen}{\ding{51}} & & $0.349$ & $0.590$ & $5.509$ & $4.654$ \\
    & LightedDepth~\cite{zhu2023lighteddepth} & \textcolor{ForestGreen}{\ding{51}} & & $0.487$ & $0.766$ & $0.631$ & $ 2.322$\\
    & DRO~\cite{gu2023dro} - ScanNet & \textcolor{RedOrange}{\ding{55}} & & $0.483$ & $0.710$ & $1.313$ & $3.093$\\
    & DRO~\cite{gu2023dro} - KITTI & \textcolor{ForestGreen}{\ding{51}} & & $0.002$ & $0.174$ & $6.397$ & $9.142$ \\
    & DUSt3R~\cite{wang2023dust3r} \textit{w.o.} Intrinsic & \textcolor{ForestGreen}{\ding{51}} & & $0.333$ & $0.631$& $0.615$& $2.740$ \\
    & DUSt3R~\cite{wang2023dust3r} \textit{w.t.} Intrinsic & \textcolor{ForestGreen}{\ding{51}} & & $0.568$ & $0.801$& $0.584$&$2.278$\\
    & Ours & \textcolor{ForestGreen}{\ding{51}} & & $\mathbf{0.695}$ & $\mathbf{0.850}$ & $\mathbf{0.535}$ & $\mathbf{1.689}$ \\
    \bottomrule
  \end{tabular}
  }
  \label{tab:scannet_camera_pose_supp}
\end{table}

\section{Extended Experiments}
\subsection{Implementation Details}
\Paragraph{\cref{sec:pose_est} Camera Pose Estimation.} 
The proposed pose estimation algorithm runs on GPU devices. 
Per frame pair, we extract the top $K=128$ normalized poses to formulate the candidate pool $\overline{\mathcal{Q}}$.
We sample $M=10,000$ points per frame pair.
We exclude correspondence with a confidence lower than $0.2$.
We set $\lambda^{\text{2D}}$ to $2$ pixels and $\lambda^{\text{3D}}$ to $0.025$ meter. 
The Hough matrix $\mathbf{H}$ resolution is set to $100 \times 200$.
The maximum Hough transform on axis-$x$, \textit{i.e.}, the $x_{\text{max}}$ is set to $1$ meter.
The BA optimization iterations are set to $T=200$ at a learning rate of $5e^{-4}$ with an Adam optimizer~\cite{kingma2017adam}.

\Paragraph{\cref{sec:triang} Triangulation.}  
For the frustum radiance field $\mathbf{V}$, we configure its resolution as $240 \times 320 \times 128$ given a ScanNet image resolution of $480 \times 640$.
We scale the depth consistent loss $L_D$ by a factor of $0.01$.
This is due to the small magnitude of correspondence consistent loss $L_C$ which is defined over normalized pixel coordinates, i.e., pixels divided by focal length.
In triangulation, we optimize the Radiance Field $\mathbf{V}$ using the Adam optimizer with a learning rate of $1e^{-4}$ over $80,000$ iterations.
In testing, we set the geometric verification threshold $\lambda^c$ to $0.01$ meter.

\Paragraph{\cref{sec:geo_ver} Geometric Verification.}
The consistent threshold $\lambda^c$ is set to $0.01$ meters, and the minimum consistent view number $n^c$ is set to $2$ frames.

\begin{figure}[t!]
    \vspace{-1mm}
    \captionsetup{font=small}
    \centering
    \subfloat[Inlier score $f^{\text{2D}}(\cdot)$ / ZoeDepth~\cite{bhat2023zoedepth} / PDC-Net~\cite{truong2023pdc}]{\includegraphics[width=0.55\linewidth]{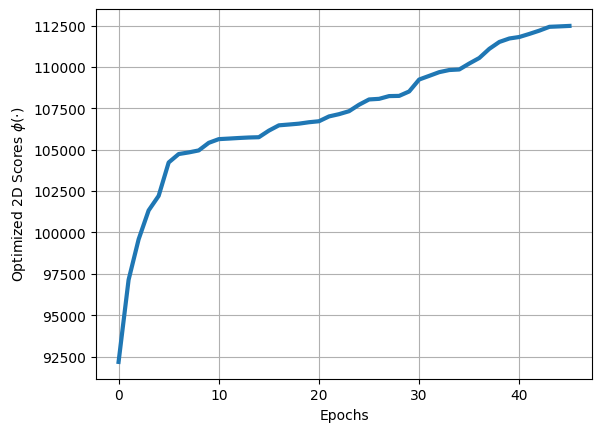}}
    \vspace{-2mm}
    \caption{\small 
    \textbf{Scores \textit{w.r.t} Optimization Epochs.} 
    The inlier scores always rise throughout the optimization.
    With $5$ frames, our algorithm terminates on average at $23.5$ epochs.
    [Key: Inlier Score / Monodepth Estimator / Correspondence Estimator]
    }
    \label{fig:inlier_opt}
\end{figure}

\subsection{Additional Comparisons}

Prior public benchmarks are inapplicable to measuring sparse-view pose estimation performance.
DeepV2D~\cite{Teed2020DeepV2D}, DRO~\cite{gu2023dro}, and LightedDepth~\cite{zhu2023lighteddepth} only report two-view relative pose performance.
DUSt3R~\cite{wang2023dust3r} reports on classic SfM and SLAM benchmarks with significantly more view variations.
Only NeRF-based~\cite{truong2023sparf} methods benchmark sparse-view pose performance and have been included in the main paper \cref{tab:sparse_render}.
For a comprehensive comparison, we additionally experiment on the ScanNet with other SoTA methods.

In \cref{tab:scannet_camera_pose_supp}, we divide the comparisons into two groups. 
The group above the dashed lines only evaluates the COLMAP successfully executed sequences.
The group below the dashed lines includes all ScanNet test sequences.
For two-view methods LightedDepth~\cite{zhu2023lighteddepth} and DRO~\cite{gu2023dro}, we repetitively apply two-view pose estimation between each support and root frame. 
For DUSt3R~\cite{wang2023dust3r} with intrinsic, we set the intrinsic and disable its update in optimization.
In \cref{tab:scannet_camera_pose_supp}, we achieve \textbf{unanimous} improvement on all frame numbers with \textbf{substantial} margin.

\subsection{Additional Ablation Studies}

\Paragraph{Scores \textit{w.r.t} Optimization Epochs.}
As shown in \cref{fig:inlier_opt}, inlier scores consistently rise during optimization.
Despite optimizing over discrete local optimal sets, \textit{i.e.}, predefined candidate pool $\overline{\mathcal{Q}}$, the convergence curve exhibits similar trends with smooth manifold optimization.
We attribute this to a sufficiently large candidate pool size, set at $K=128$ per frame in our experiment.

\begin{figure}[t!]
    \vspace{-1mm}
    \captionsetup{font=small}
    \centering
    \subfloat[Accuracy $\delta < 0.5$ / ZoeDepth~\cite{bhat2023zoedepth} / PDC-Net~\cite{truong2023pdc}]{\includegraphics[width=0.6\linewidth]{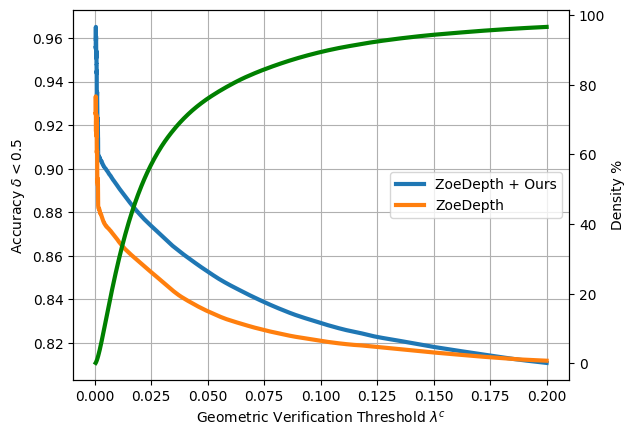}}
    \vspace{-2mm}
    \caption{\small 
    \textbf{Depth Density \& Accuracy \textit{w.r.t} Geometric Verification Threshold $\lambda^c$.} 
    \cref{tab:self_sup_depth} reports triangulated accuracy and density at $\lambda^c = 0.01$.
    From the plot, if set the threshold $\lambda^c$ to a larger value,  we maintain significant improvement while preserving a significantly larger density, \textit{e.g.}, $\lambda^c = 0.05$. The unit is in meters.
    [Key: Accuracy $\delta < 0.5$ / Monodepth Estimator / Correspondence Estimator]
    }
    \label{fig:geo_curve}
\end{figure}

\Paragraph{Geometric Verification Threshold.}
In \cref{tab:self_sup_depth_supp}, we benchmark the triangulated depth without geometric verification, \textit{i.e.}, depth performance without filtering.
On ZeroDepth~\cite{liu2023zero}, we outperform the supervised inputs even without geometric verification.
On ZoeDepth~\cite{bhat2023zoedepth} and Metric3D~\cite{yin2023metric3d}, we outperform the supervised inputs after the geometric verification, suggesting its necessity.
Across all cases, geometric verification improves performance.
In \cref{fig:geo_curve}, we ablate different triangulation threshold $\lambda^c$ values.
From \cref{fig:geo_curve}, our method maintains improvement at a much higher density than reported in the main paper \cref{tab:self_sup_depth}.

\begin{table*}[t!]
  \centering
  \captionsetup{font=small}
  \caption{
  \small
  \textbf{
  Ablation on Geometric Verification
  } on ScanNet dataset.
  We additionally report the main paper \cref{tab:self_sup_depth} without applying the \cref{sec:geo_ver} Geometric Verification, \textit{i.e.}, the radiance field intermediate depthmap with full density.
  See visual examples in the main paper \cref{fig:triang_depth} and supplementary \cref{fig:supp_sdepth}.
  }
  \resizebox{ 0.85\linewidth}{!}{%
  \begin{tabular}{lcc|cc|ccccc}
    \hline
    \textbf{Method} & \textbf{Geo. Ver.} & \textbf{Density}  & \cellcolor[RGB]{155, 187, 228} $\mathbf{\delta_{0.5}}$ & \cellcolor[RGB]{155, 187, 228} $\mathbf{\delta_1}$ & \cellcolor[RGB]{222, 164, 151} $\textbf{SI}_{\textbf{log}}$ & \cellcolor[RGB]{222, 164, 151} \textbf{A.Rel} & \cellcolor[RGB]{222, 164, 151} \textbf{S.Rel}& \cellcolor[RGB]{222, 164, 151} \textbf{RMS} & \cellcolor[RGB]{222, 164, 151} $\textbf{RMS}_{\textbf{log}}$ \\
    \hline
    ZoeDepth~\cite{bhat2023zoedepth} & \multirow{2}{*}{\textcolor{red}{\textbf{\ding{55}}}} & \multirow{2}{*}{Dense} & $\mathbf{0.808}$ & $\mathbf{0.937}$    & $\mathbf{9.592}$  & $\mathbf{0.074}$ & $\mathbf{0.028}$ & $\mathbf{0.211}$ & $\mathbf{0.102}$  \\
    $\drsh$ Ours & & & $0.800$ & $0.925$ & $11.006$ & $0.082$ & $0.032$ & $0.227$ & $0.118$  \\
    \cdashline{1-10}
    ZoeDepth~\cite{bhat2023zoedepth} & \multirow{2}{*}{\textcolor{ForestGreen}{\textbf{\ding{51}}}} & \multirow{2}{*}{ $9.1 \%$}   & $0.877$ & $0.963$      & $6.655$          & $0.056$          & $0.016$          & $0.154$          & $0.075$  \\
    $\drsh$ Ours & & & $\mathbf{0.902}$ & $\mathbf{0.976}$ &  $\mathbf{5.901}$ & $\mathbf{0.050}$ & $\mathbf{0.014}$ & $\mathbf{0.149}$ & $\mathbf{0.070}$   \\
    \cline{1-10}
    ZeroDepth~\cite{liu2023zero} & \multirow{2}{*}{\textcolor{red}{\textbf{\ding{55}}}} & \multirow{2}{*}{ Dense}   & $0.557$ & $0.771$      & $20.124$          & $0.166$          & $0.118$          & $0.427$          & $0.211$  \\
    $\drsh$ Ours & & & $\mathbf{0.595}$ & $\mathbf{0.809}$ &  $\mathbf{17.744}$ & $\mathbf{0.147}$ & $\mathbf{0.094}$ & $\mathbf{0.388}$ & $\mathbf{0.190}$   \\
    \cdashline{1-10}
    ZeroDepth~\cite{liu2023zero} & \multirow{2}{*}{\textcolor{ForestGreen}{\textbf{\ding{51}}}} & \multirow{2}{*}{ $5.6 \%$}   & $0.641$ & $0.834$      & $12.860$          & $0.124$          & $0.086$          & $0.337$          & $0.152$  \\
    $\drsh$ Ours & & & $\mathbf{0.686}$ & $\mathbf{0.877}$ &  $\mathbf{9.463}$ & $\mathbf{0.106}$ & $\mathbf{0.067}$ & $\mathbf{0.295}$ & $\mathbf{0.133}$   \\
    \cline{1-10}
    Metric3D~\cite{yin2023metric3d} & \multirow{2}{*}{\textcolor{red}{\textbf{\ding{55}}}} & \multirow{2}{*}{ Dense}   & $\mathbf{0.747}$ & $\mathbf{0.905}$      & ${12.325}$          & $\mathbf{0.089}$          & $\mathbf{0.043}$          & $\mathbf{0.259}$          & $\mathbf{0.129}$  \\
    $\drsh$ Ours & & & ${0.746}$ & ${0.883}$ &  $\mathbf{12.273}$ & ${0.109}$ & ${0.089}$ & ${0.306}$ & ${0.147}$   \\
    \cdashline{1-10}
    Metric3D~\cite{yin2023metric3d} & \multirow{2}{*}{\textcolor{ForestGreen}{\textbf{\ding{51}}}} & \multirow{2}{*}{ $2.6 \%$}   & $0.804$ & $0.946$      & $6.708$          & $0.067$          & $0.020$          & $0.150$          & $0.084$  \\
    $\drsh$ Ours & & & $\mathbf{0.854}$ & $\mathbf{0.968}$ &  $\mathbf{4.170}$ & $\mathbf{0.055}$ & $\mathbf{0.014}$ & $\mathbf{0.125}$ & $\mathbf{0.068}$   \\
    \bottomrule
  \end{tabular}
  }
  \vspace{2mm}
  \label{tab:self_sup_depth_supp}
\end{table*}

\subsection{Evaluation Metrics}
\Paragraph{Depth Related Metrics.} 
In \cref{tab:self_sup_depth}, we follow \cite{bhat2021adabins} for most evaluation metrics.
As the accuracy metric $\delta_{1}$ saturates, \cite{piccinelli2023idisc} introduces a stricter counterpart, $\delta_{0.5}$.
The term $\text{SI}_{\text{log}}$ is a scale-invariant metric~\cite{piccinelli2023idisc}.
As a result, the scale adjustment on support frames does not impact metric $\text{SI}_{\text{log}}$, as in main paper \cref{tab:self_sup_depth} row $5$ and $6$.

\Paragraph{Correspondence Related Metrics.} 
In \cref{tab:corres}, we follow \cite{truong2023pdc} in evaluation metrics.
The metric \text{AEPE} refers to average end-point-error, \textit{i.e.}, the L2 norm between estimated and groundtruth correspondence.
\text{PCK-}${K}$ refers to the percentage of correct correspondence if its end-point-error is smaller than a given pixel threshold ${K}$.

\Paragraph{Pose Related Metrics.}
In \cref{tab:scannet_camera_pose} and \cref{tab:sparse_render}, we follow \cite{truong2023sparf} in using $\text{Rot.}$ and $\text{Trans.}$.
The metric $\text{Rot.}$ measures the average difference in degree after alignment.
Similarly, metric $\text{Trans.}$ measures the translation magnitude after multiplied by $\times 100$.
We additionally report \text{PCK-}${K}$ and \text{C3D-}${K}$.
The metric \text{PCK-}${K}$ compares the projected correspondence using the groundtruth depthmaps.
Its accuracy is only related to the estimated camera poses.
The threshold is set to $3$ pixels
Similarly, the metric \text{C3D-}${K}$ compares the backprojected 3D points using the groundtruth depthmap with estimated poses.
The threshold is set to $5$ centimeters.

\subsection{Visualization}
\begin{figure*}[t!]
    \vspace{-5mm}
    \captionsetup{font=small}
    \centering
    \subfloat[]{\includegraphics[width=0.46\linewidth]{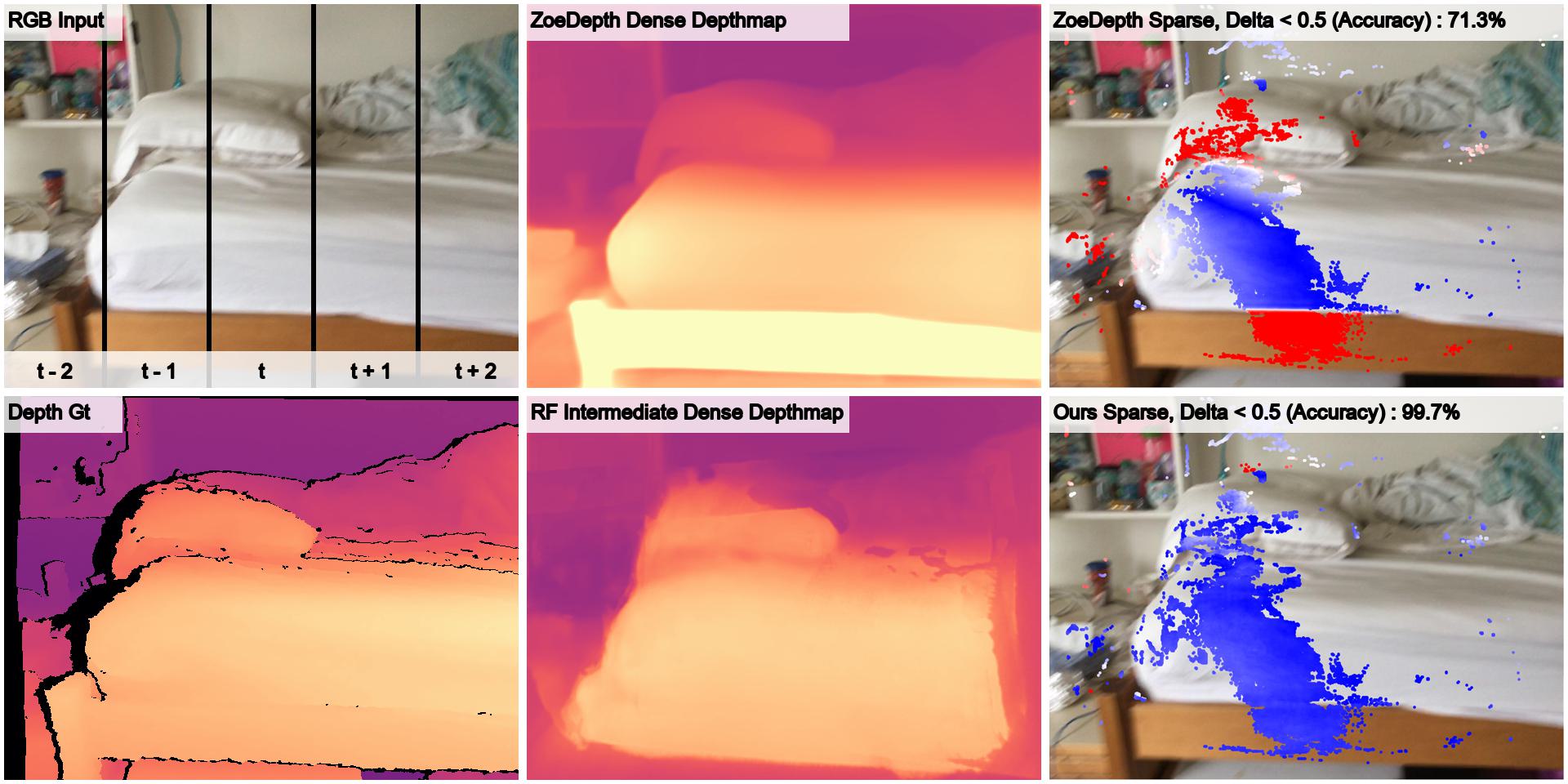}} \, 
    \subfloat[]{\includegraphics[width=0.46\linewidth]{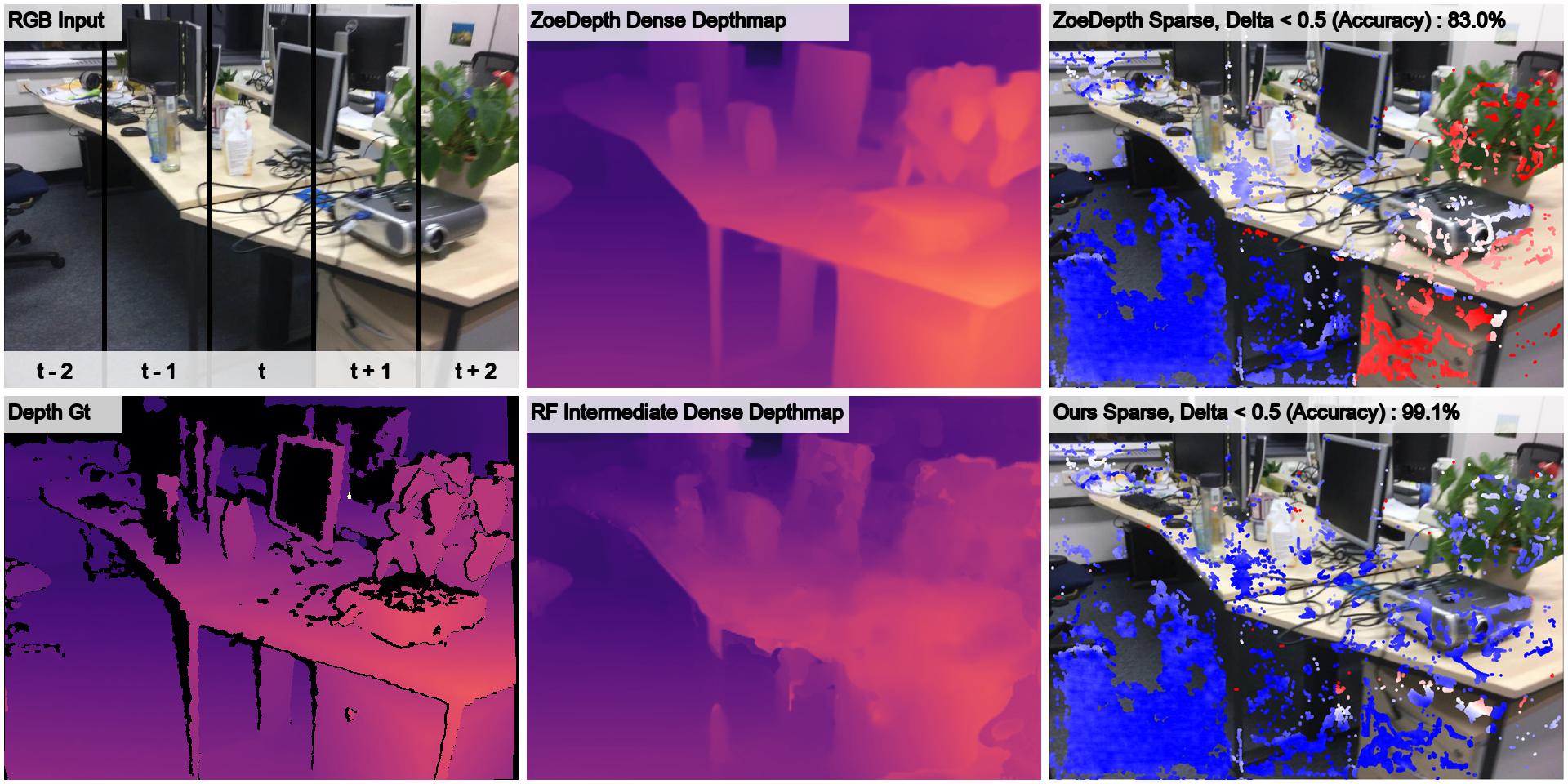}} \\
    \subfloat[]{\includegraphics[width=0.46\linewidth]{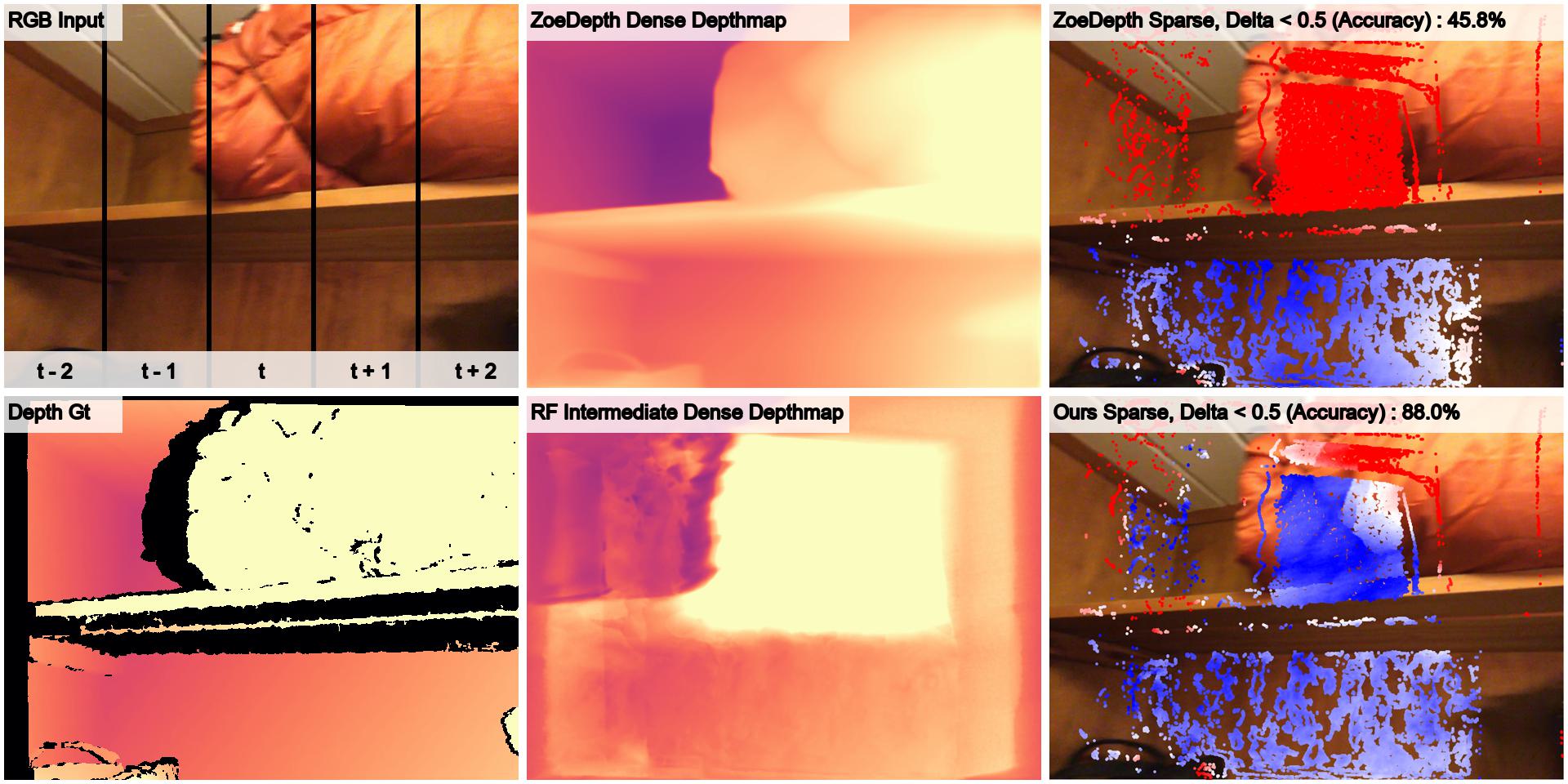}} \, 
    \subfloat[]{\includegraphics[width=0.46\linewidth]{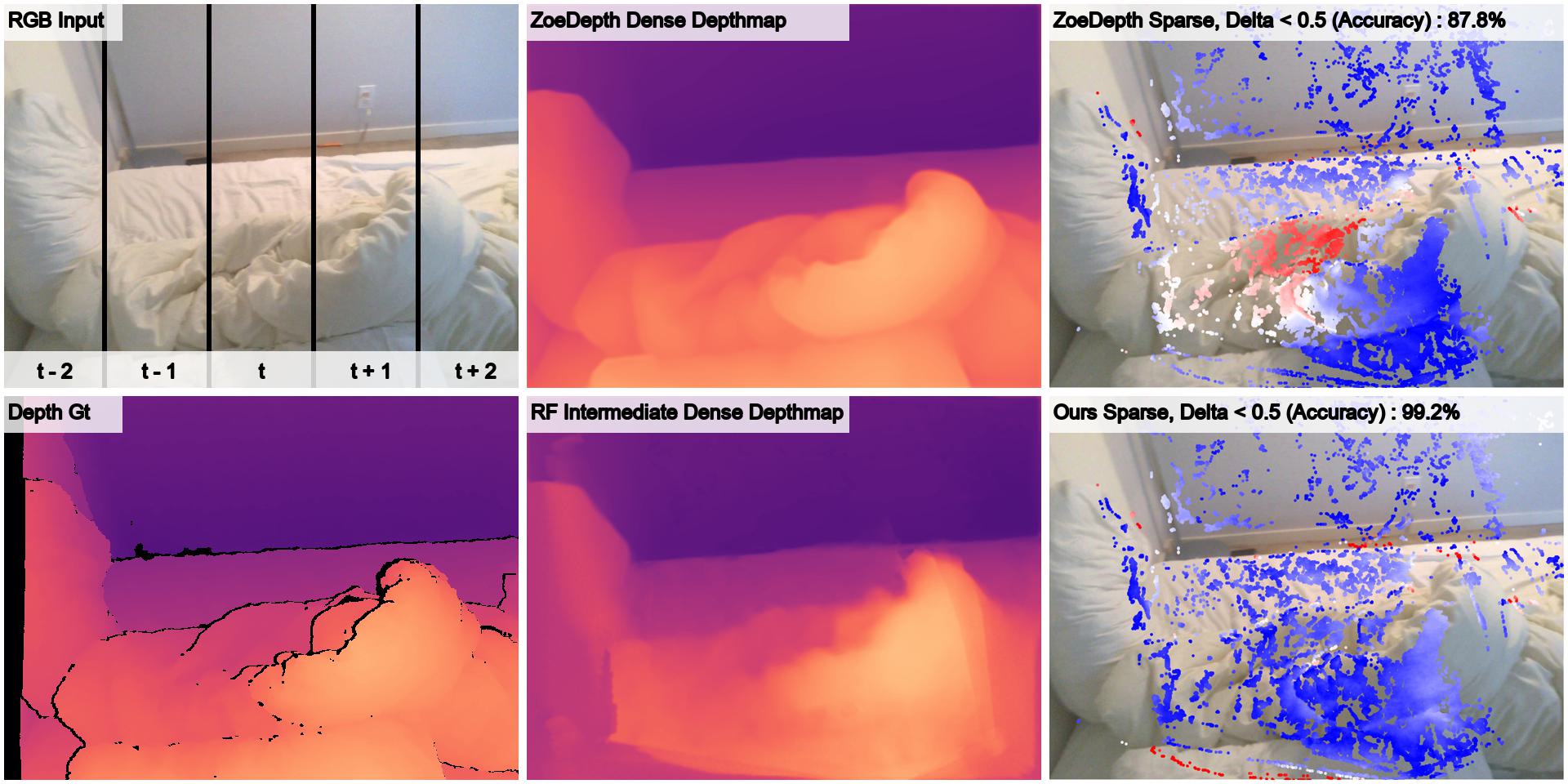}} \\
    \subfloat[]{\includegraphics[width=0.46\linewidth]{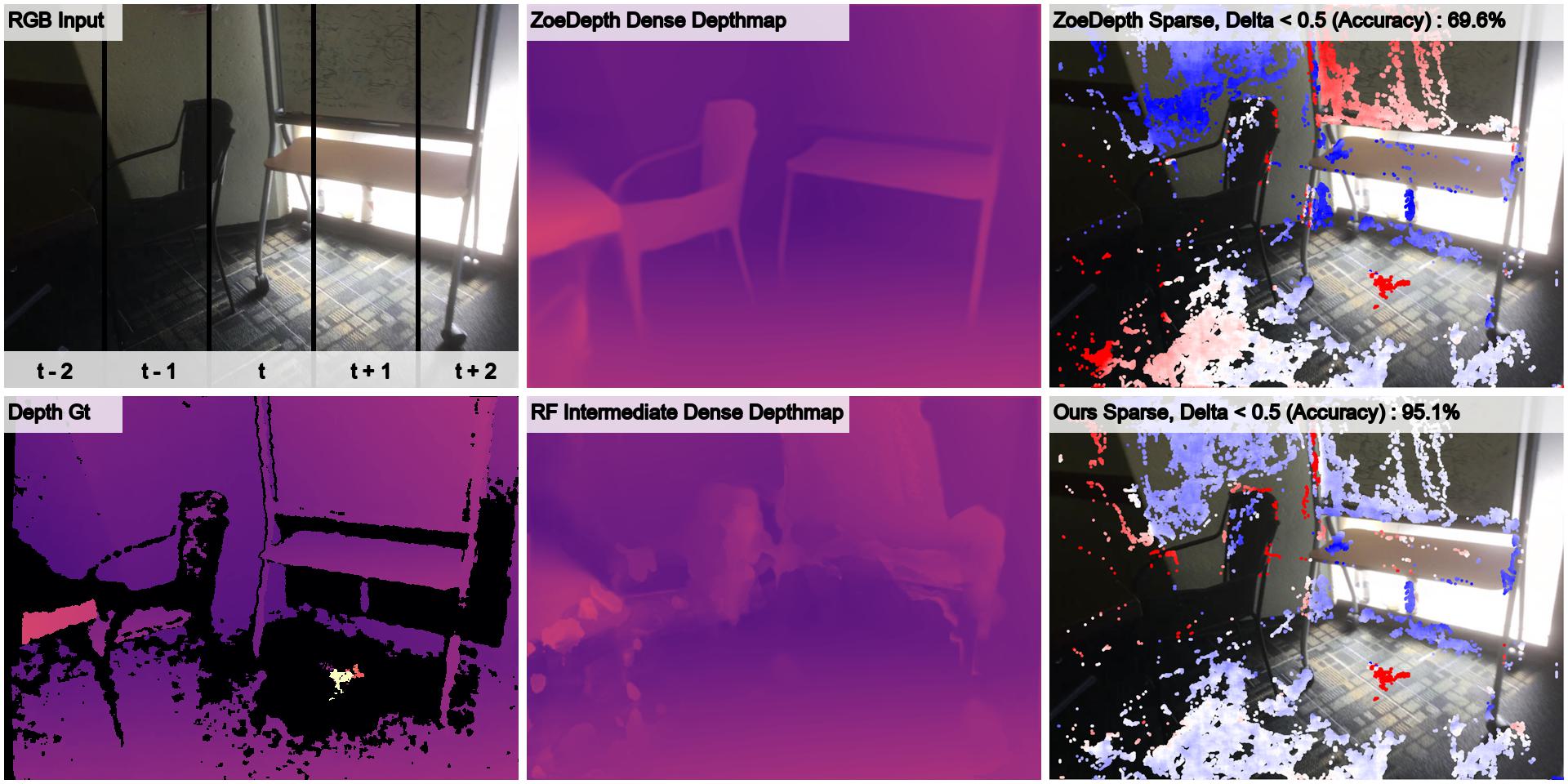}} \, 
    \subfloat[]{\includegraphics[width=0.46\linewidth]{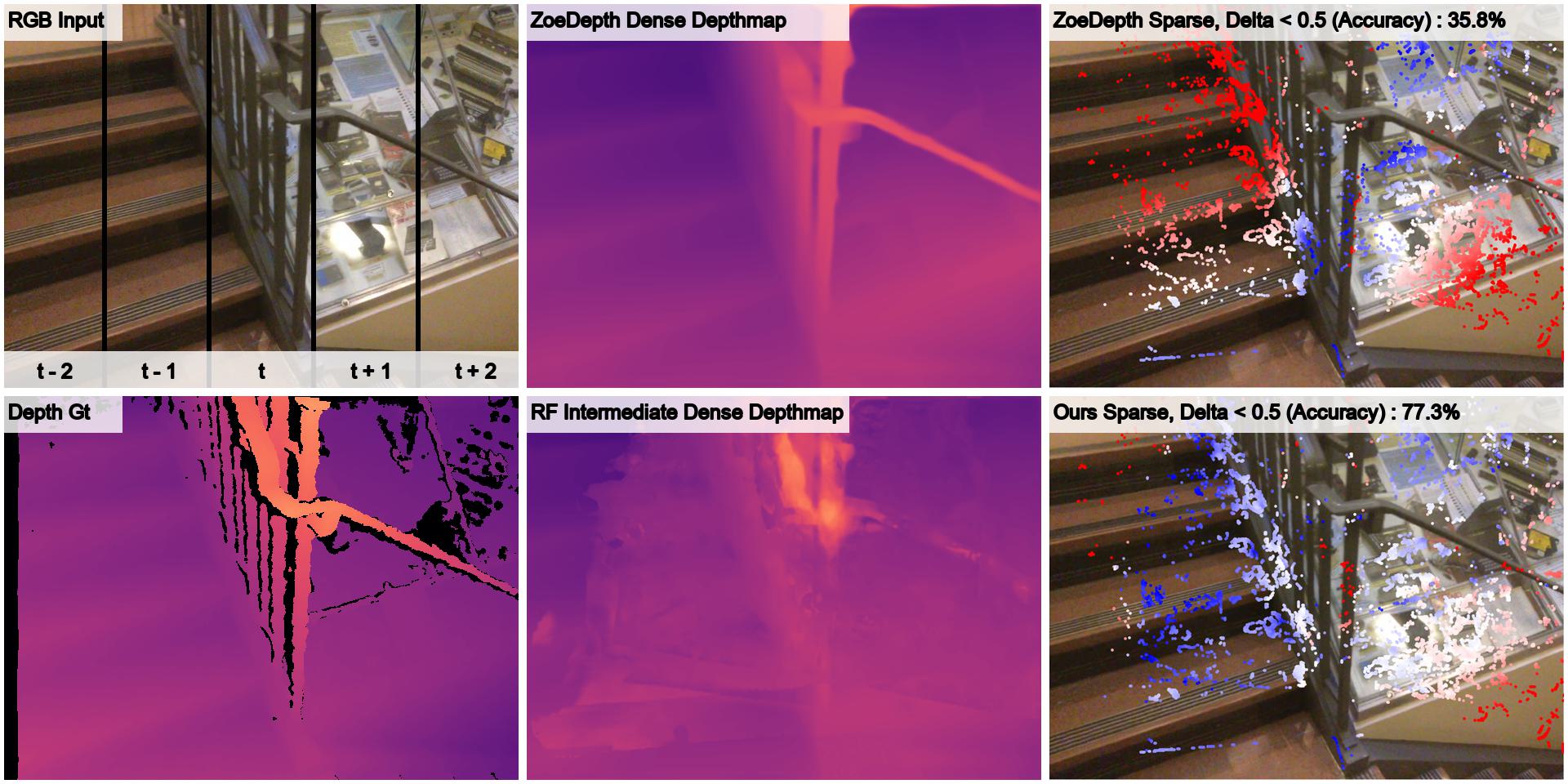}} \\
    \subfloat[]{\includegraphics[width=0.46\linewidth]{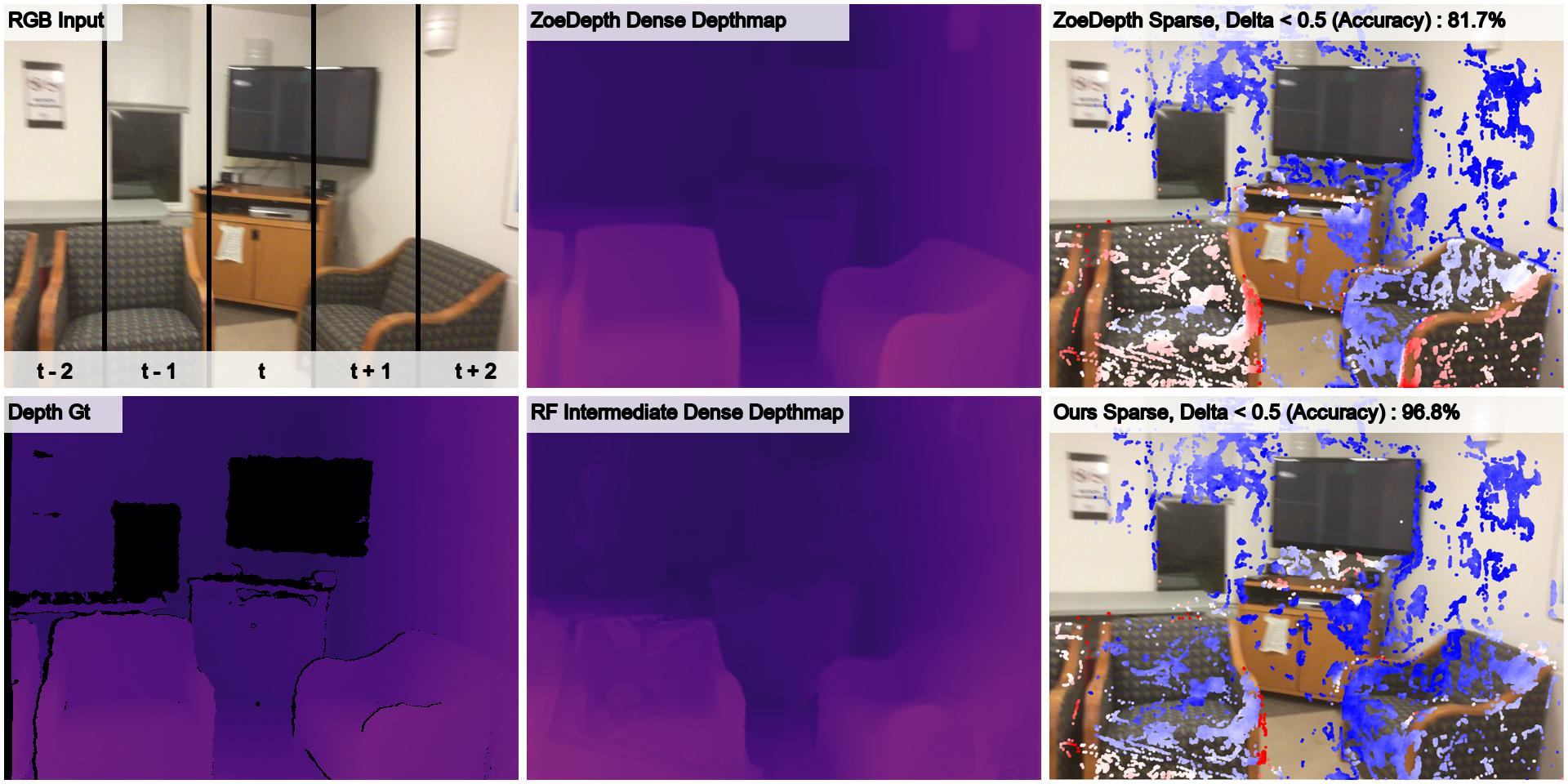}} \, 
    \subfloat[]{\includegraphics[width=0.46\linewidth]{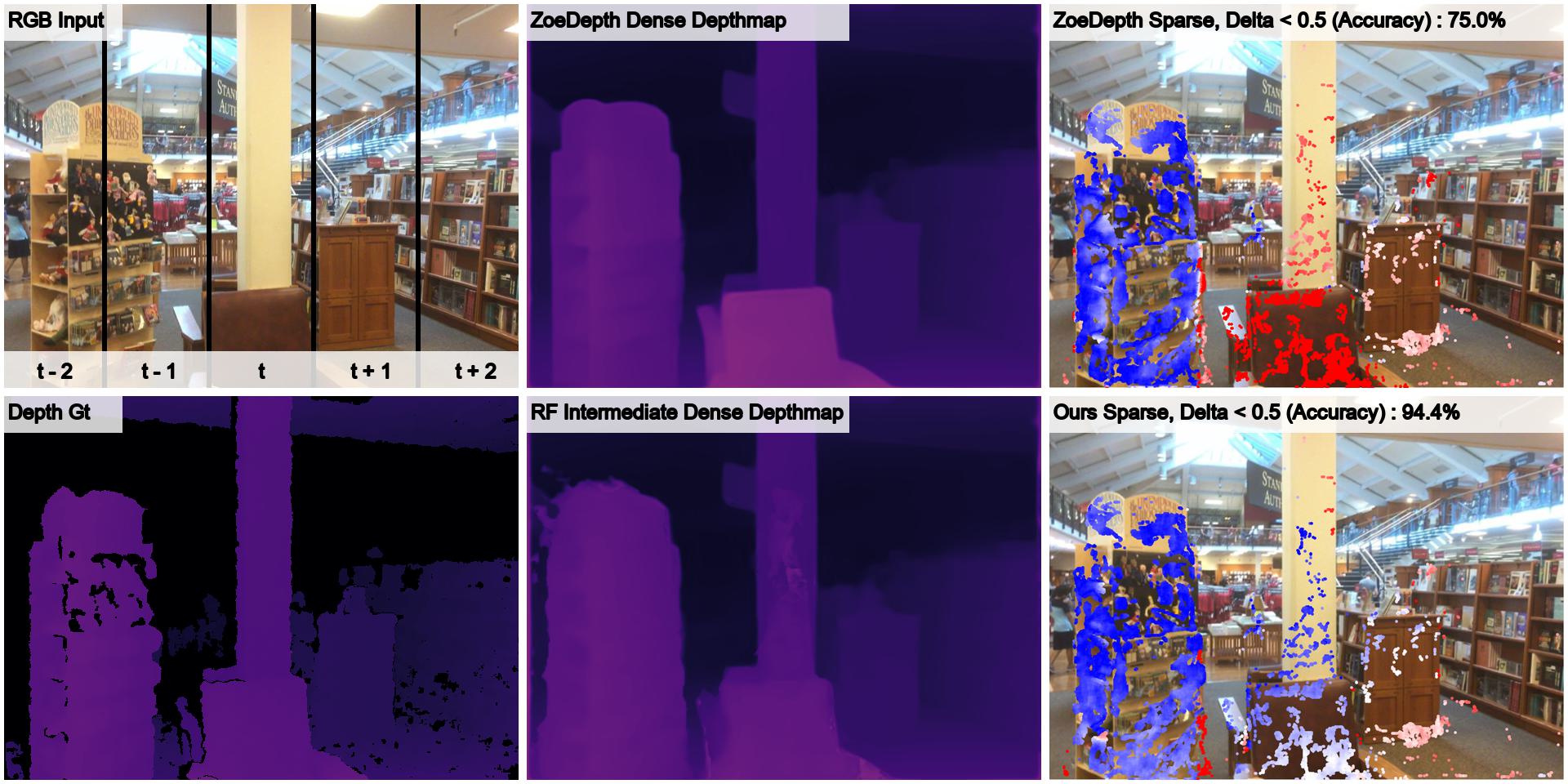}} \\
    \subfloat[]{\includegraphics[width=0.46\linewidth]{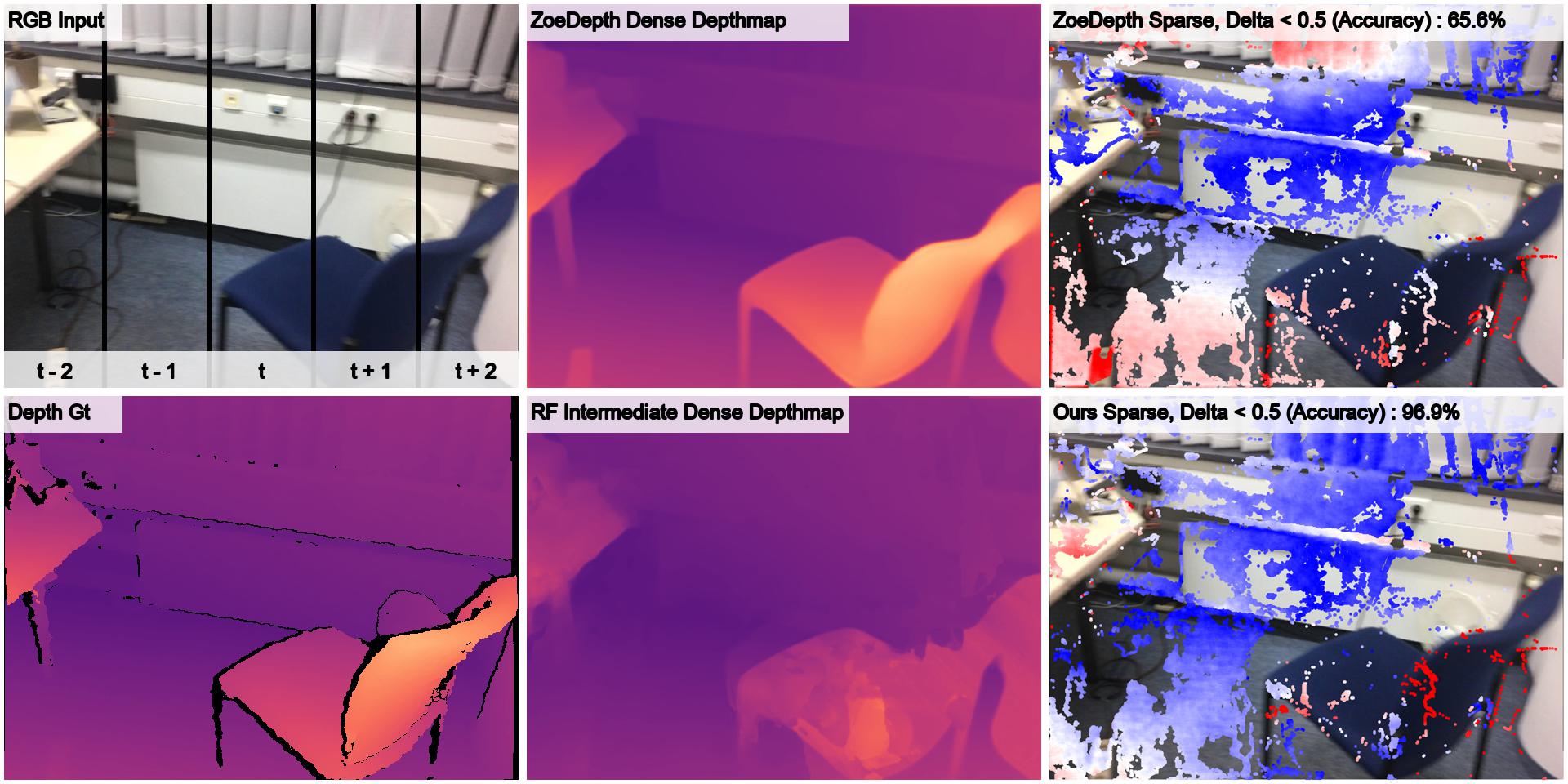}} \, 
    \subfloat[]{\includegraphics[width=0.46\linewidth]{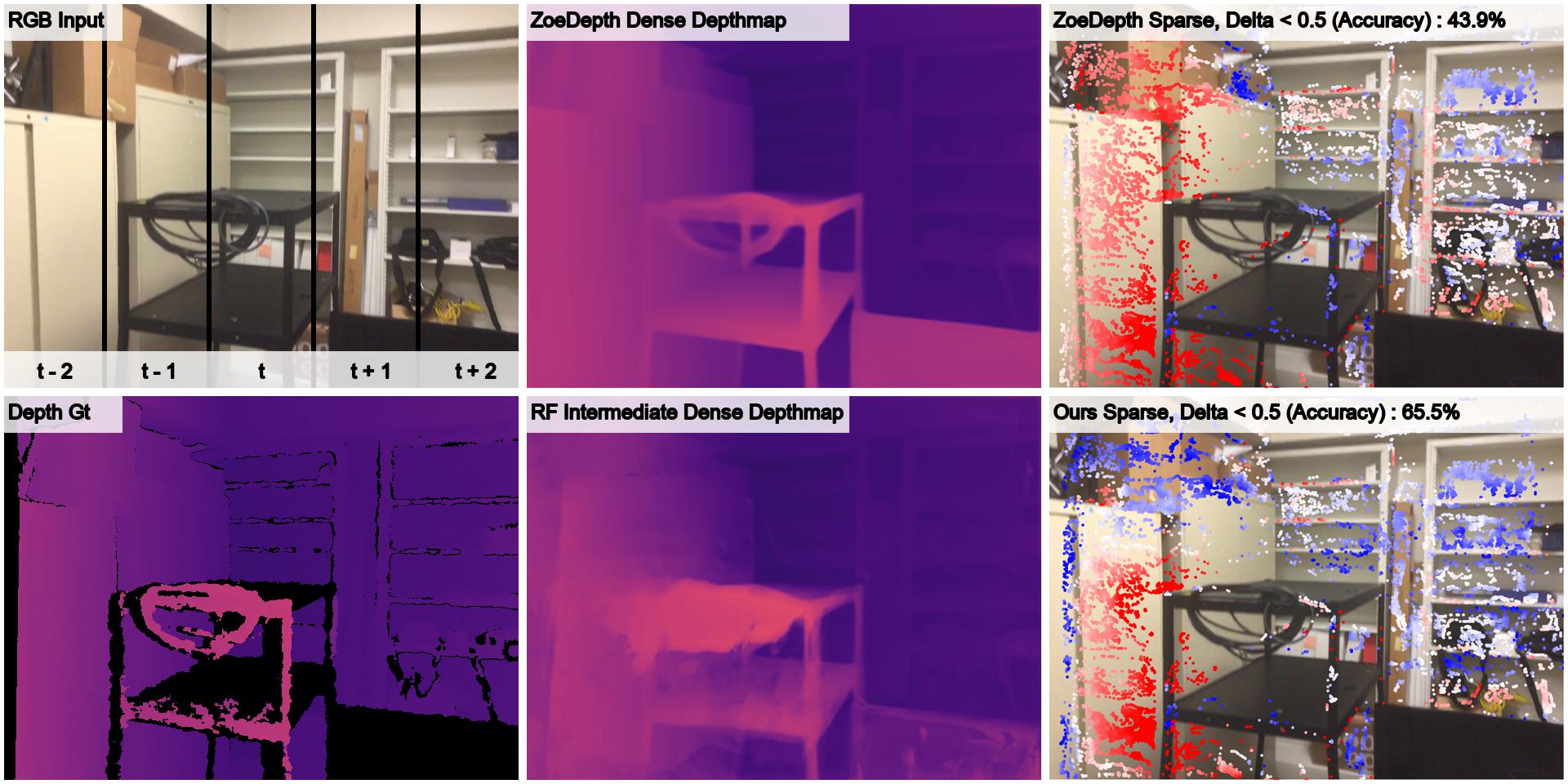}} \\
    \vspace{-2mm}
    \caption{\small 
    \textbf{Self-supervised Depth Estimation.} 
    Qualitative comparison with supervised model ZoeDepth~\cite{bhat2023zoedepth}.
    We represent the depth quality of each pixel with the metric \textbf{A.Rel}.
    Blue to red scatter indicates small to large errors, ranging between $0.0$ and $0.2$.
    }
    \label{fig:supp_sdepth}
\end{figure*}

\begin{figure*}[t!]
    \vspace{-1mm}
    \captionsetup{font=small}
    \centering
    \subfloat[]{\includegraphics[width=0.48\linewidth]{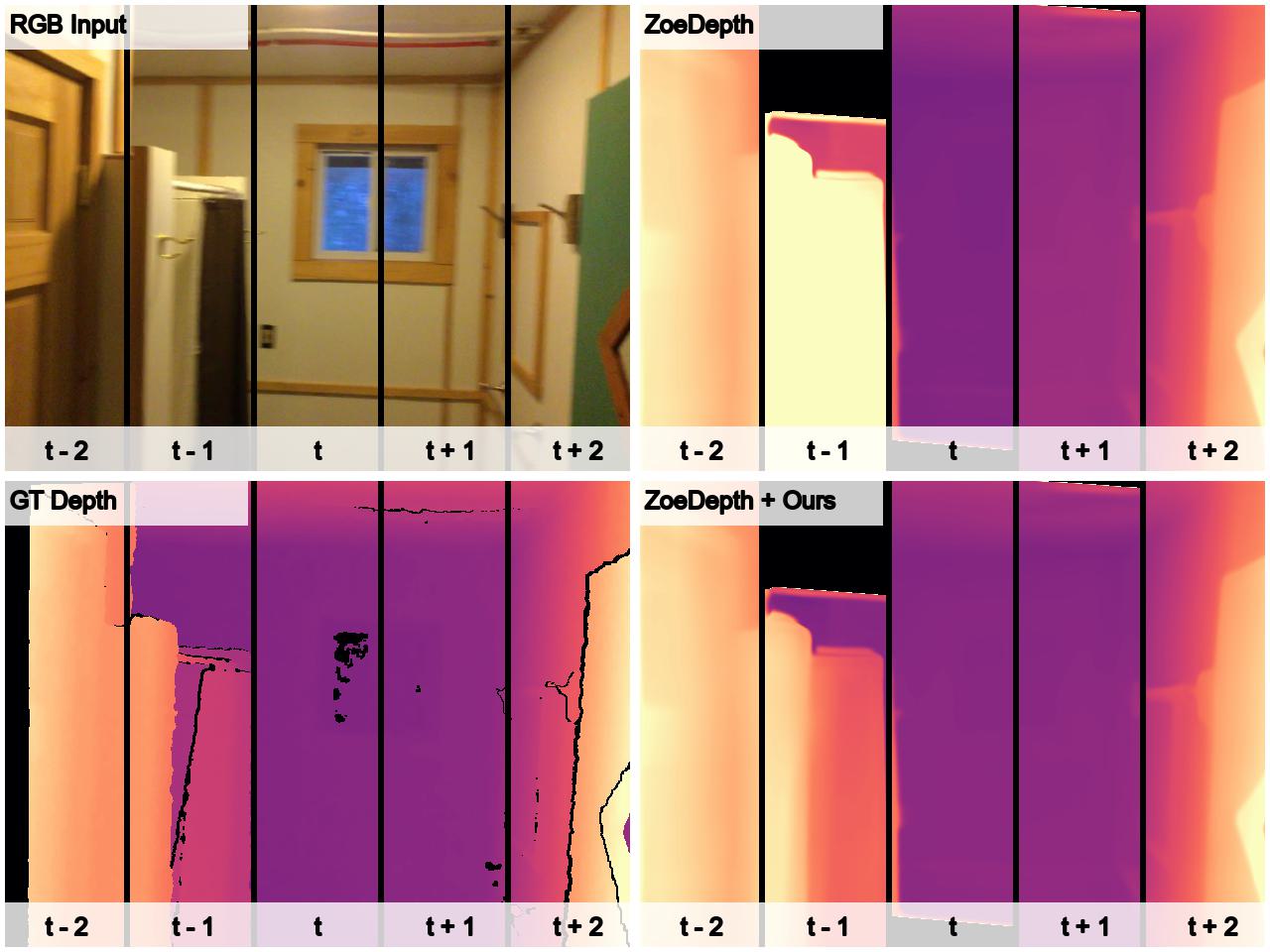}} \, 
    \subfloat[]{\includegraphics[width=0.48\linewidth]{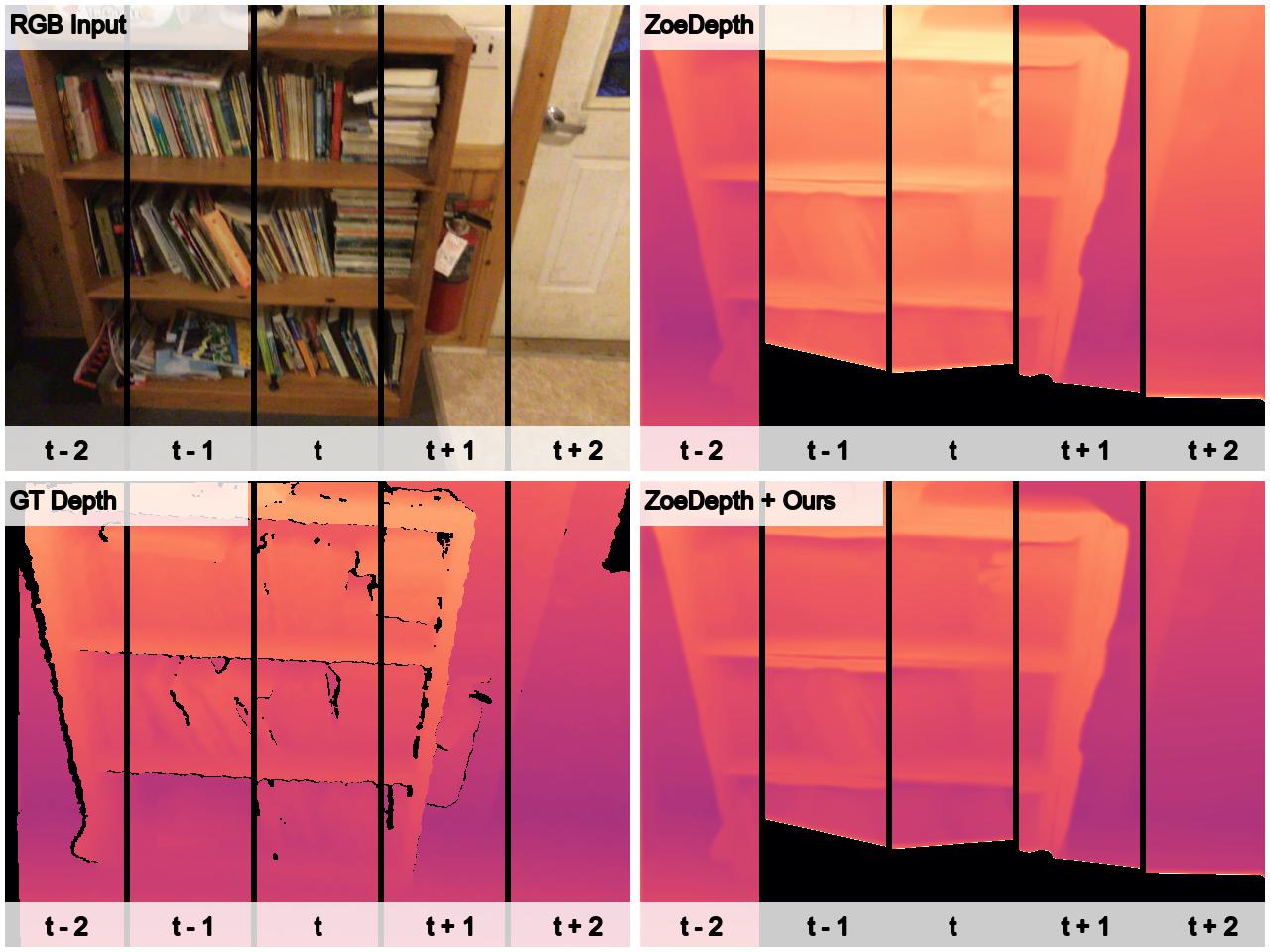}} \\
    \subfloat[]{\includegraphics[width=0.48\linewidth]{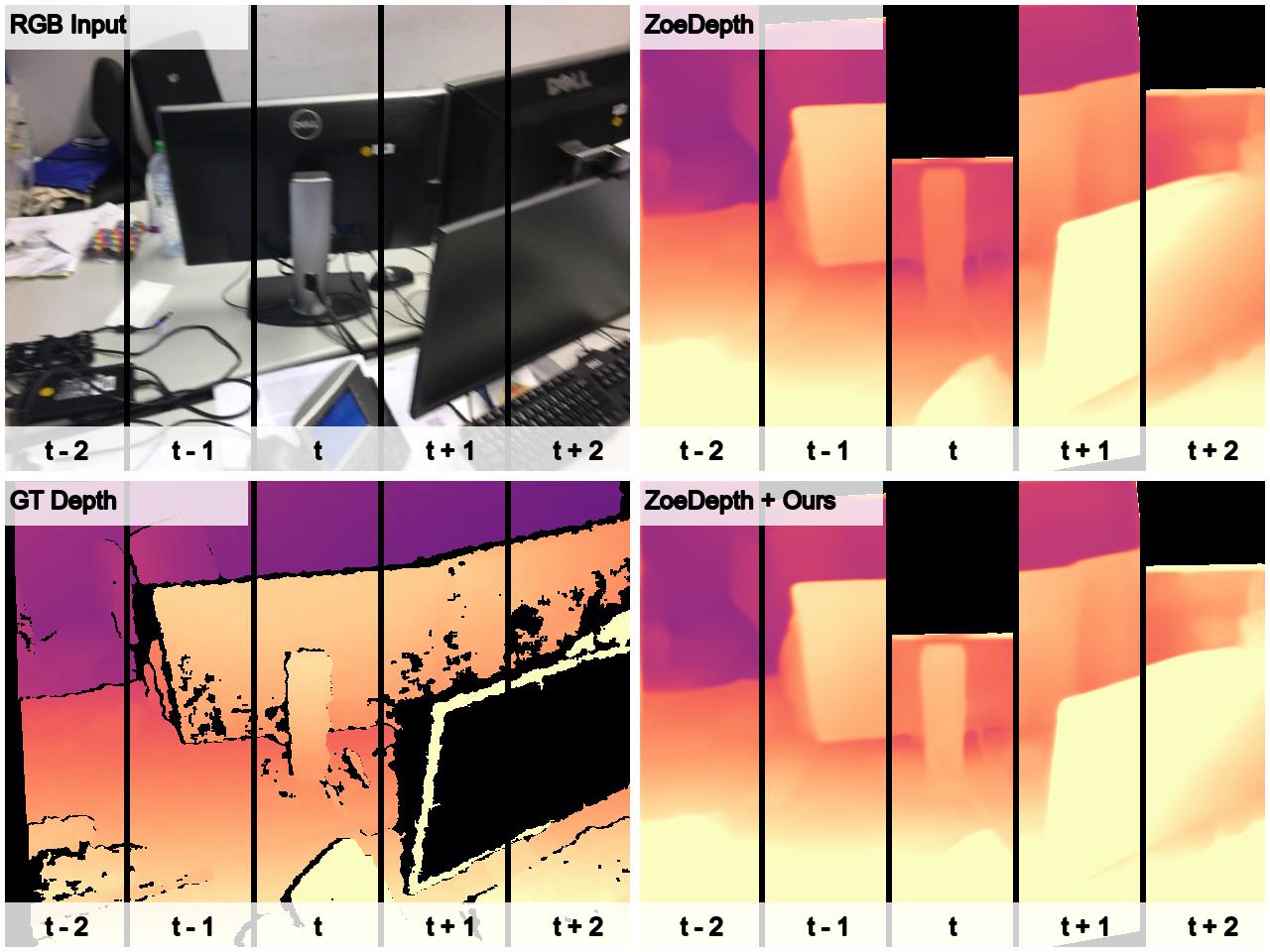}} \, 
    \subfloat[]{\includegraphics[width=0.48\linewidth]{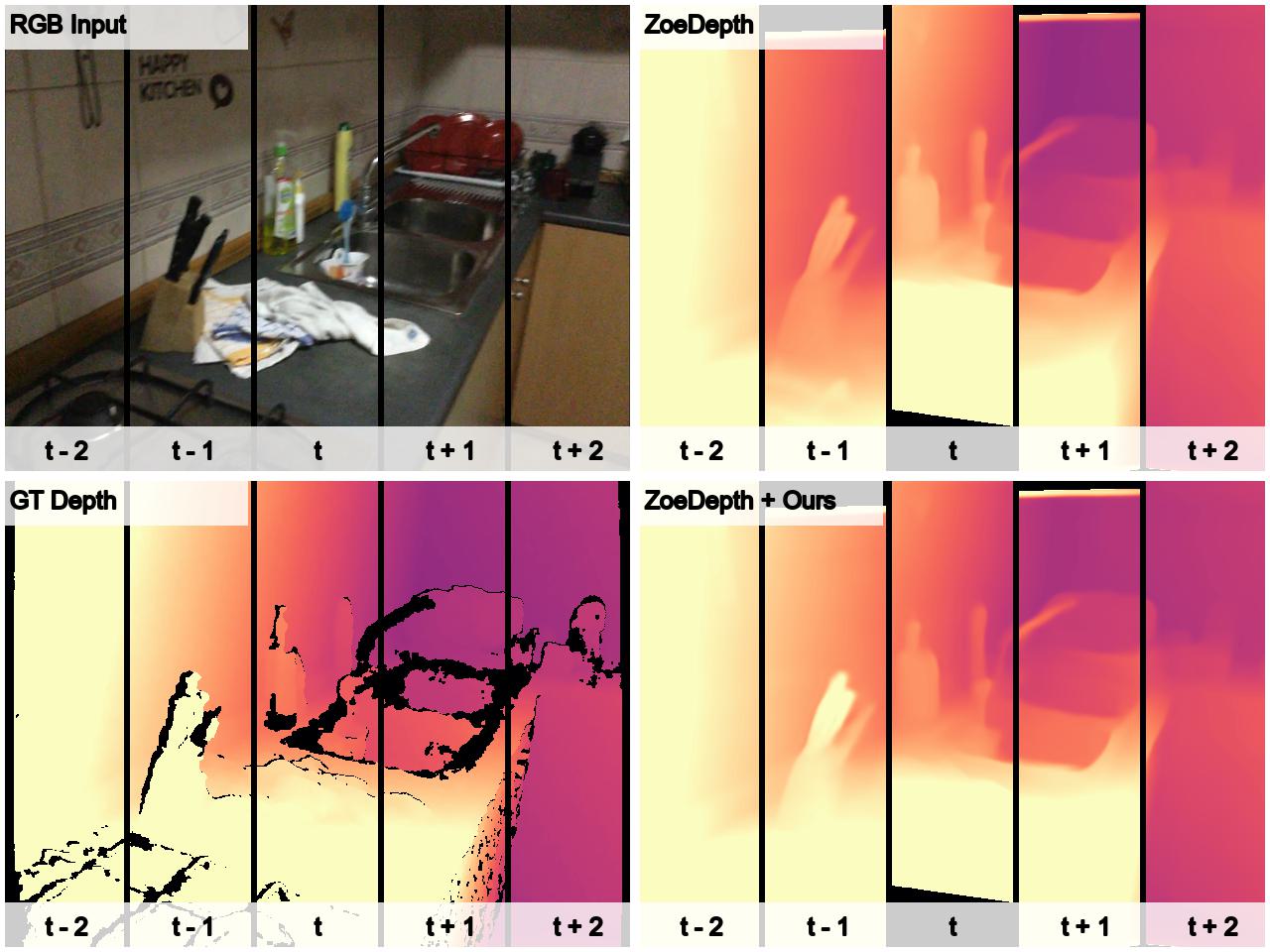}} \\
    \subfloat[]{\includegraphics[width=0.48\linewidth]{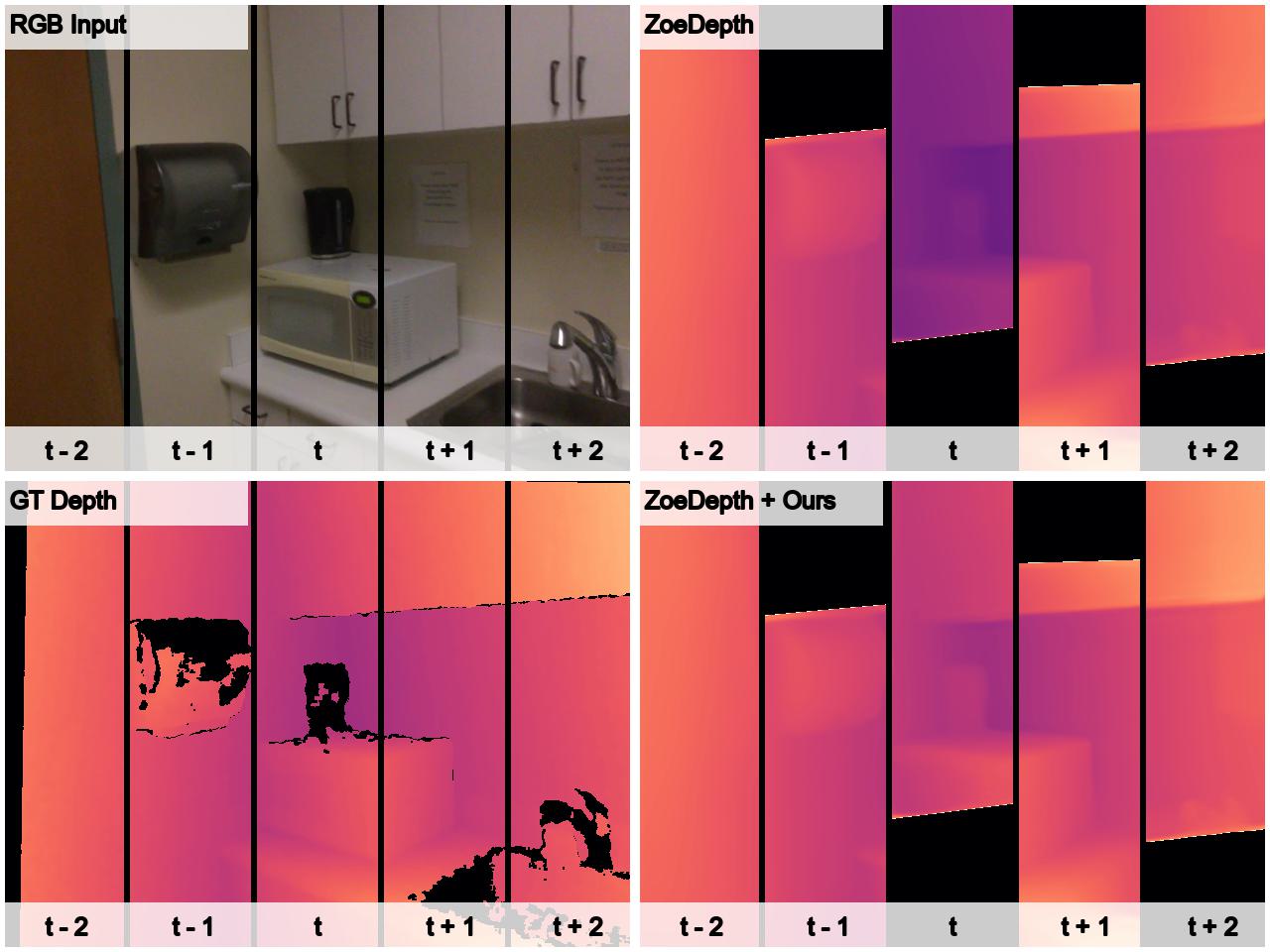}} \, 
    \subfloat[]{\includegraphics[width=0.48\linewidth]{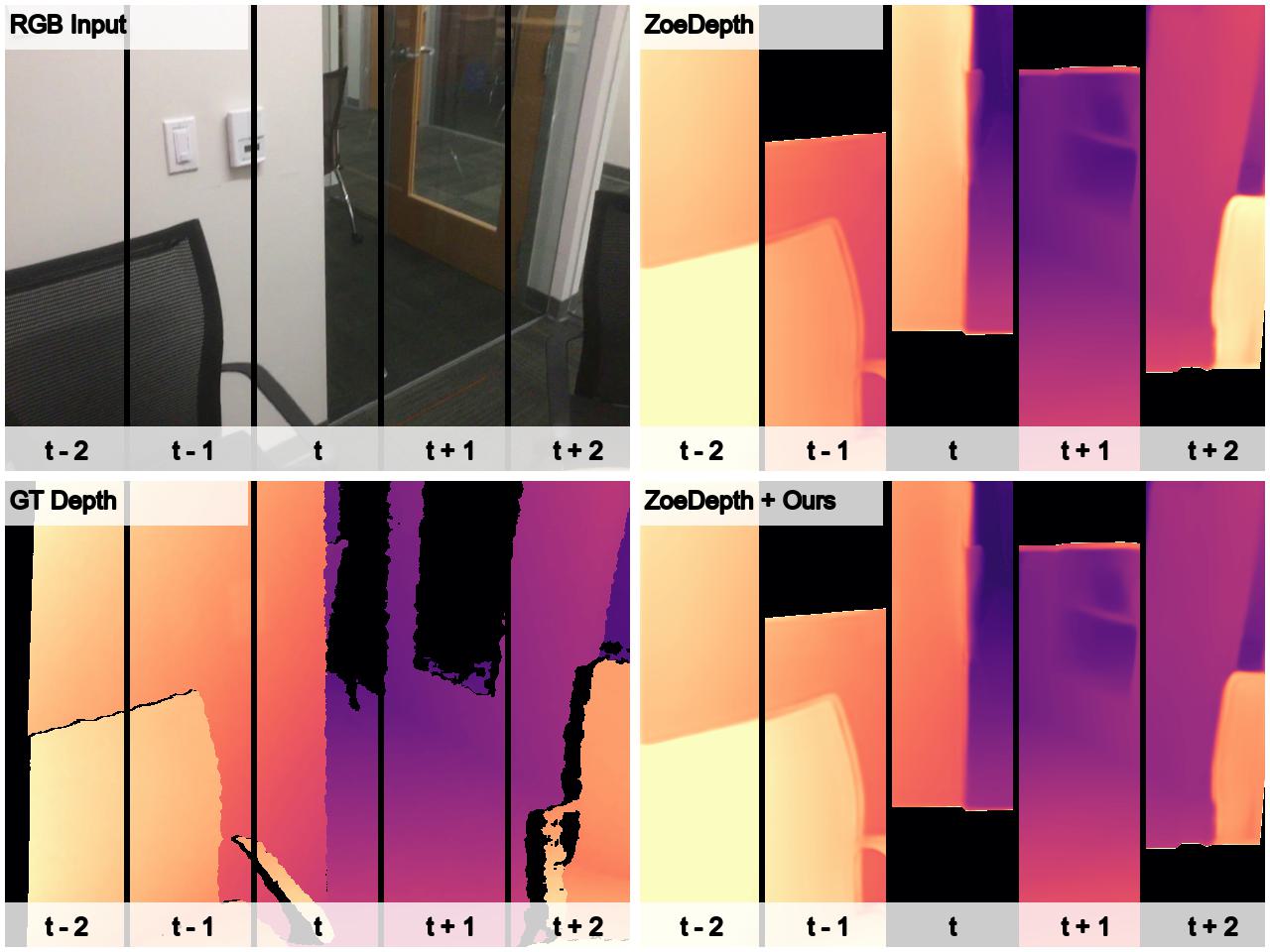}} \\
    \vspace{-2mm}
    \caption{\small 
    \textbf{Consistent Depth Estimation.} 
    Qualitative comparison with input SoTA monocular depth estimator ZoeDepth~\cite{bhat2023zoedepth}.
    }
    \label{fig:supp_cdepth}
\end{figure*}

\begin{figure*}[t!]
    \vspace{-1mm}
    \captionsetup{font=small}
    \centering
    \subfloat[]{\includegraphics[width=0.48\linewidth]{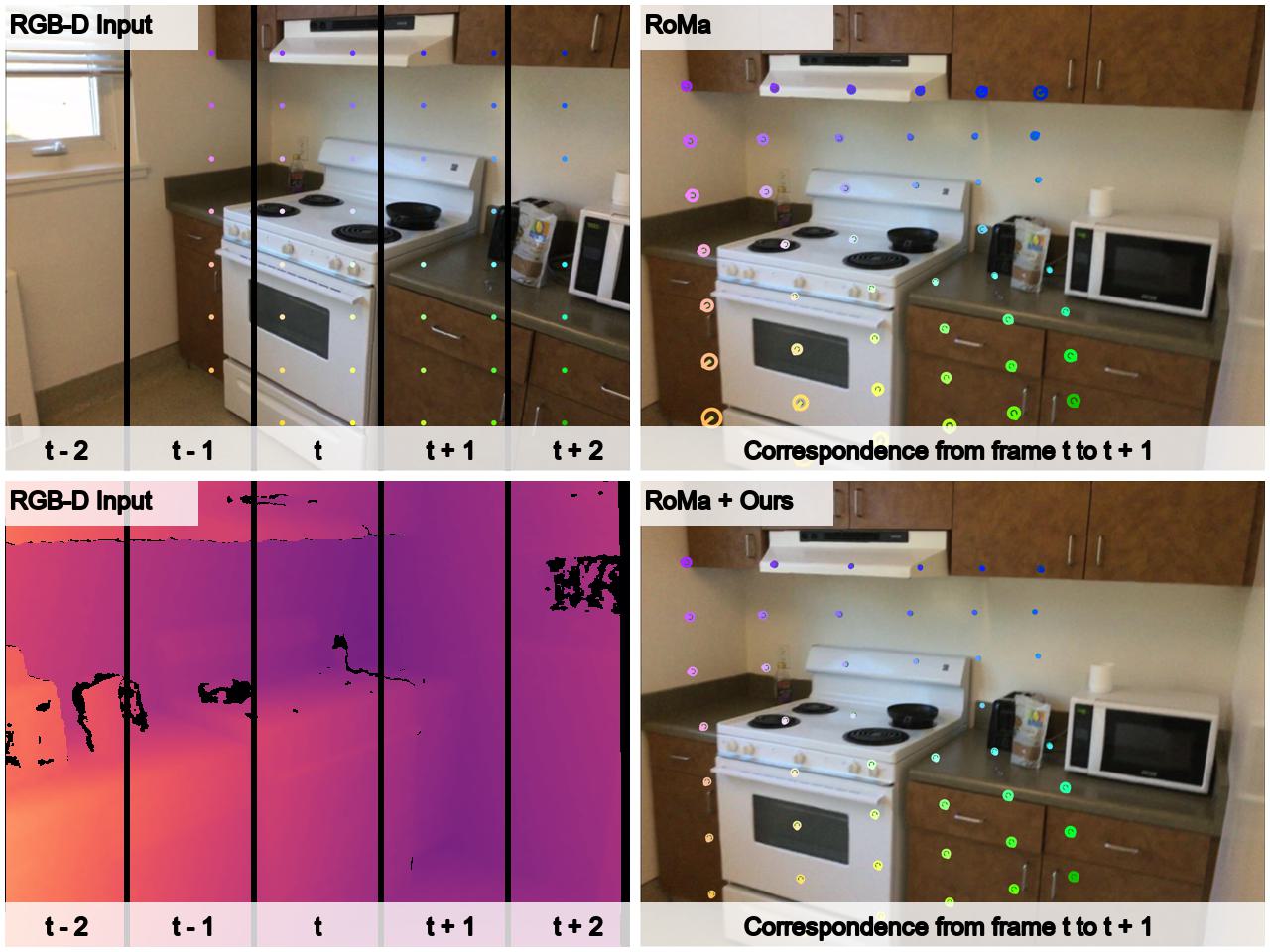}} \, 
    \subfloat[]{\includegraphics[width=0.48\linewidth]{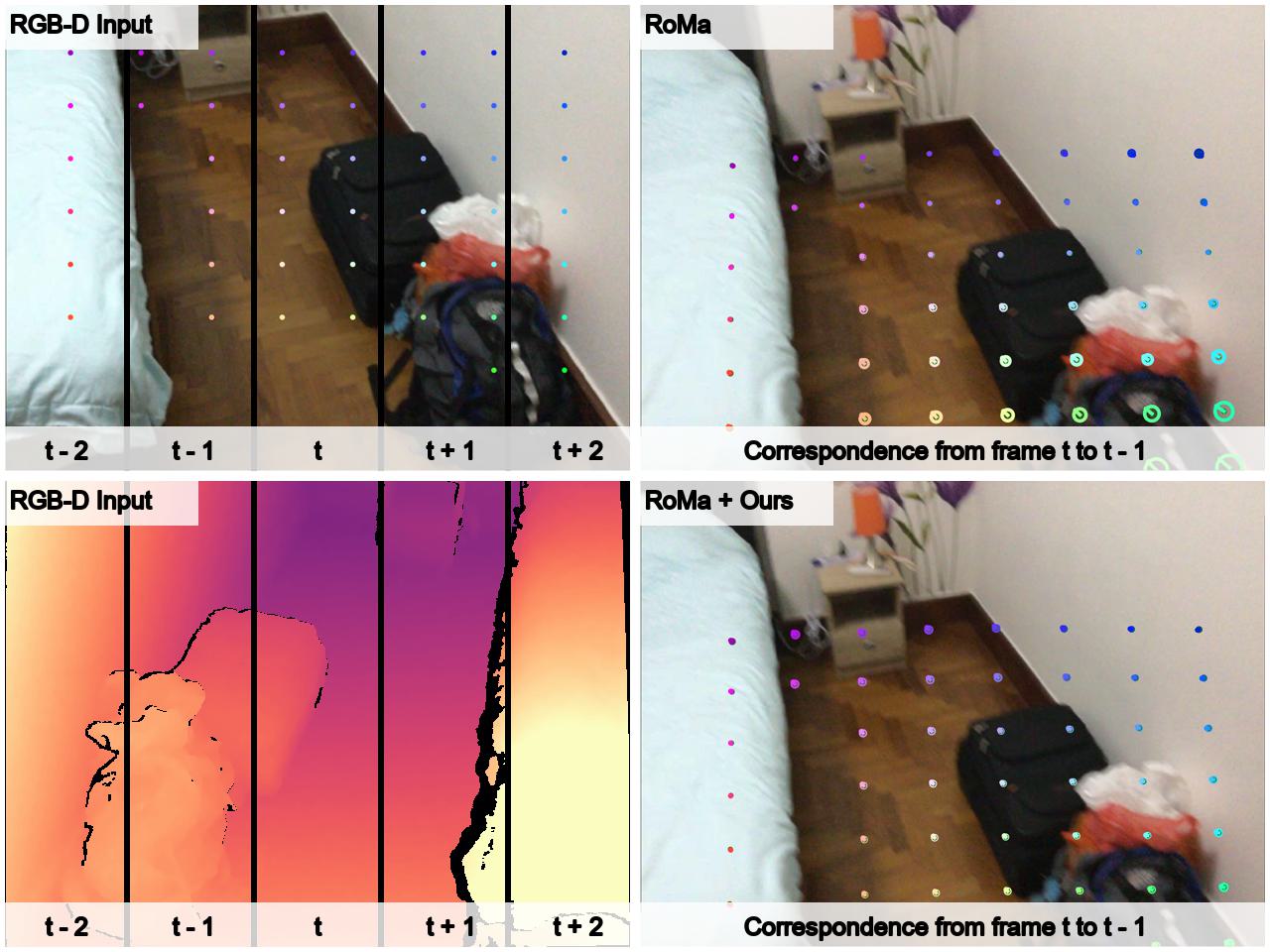}} \\
    \subfloat[]{\includegraphics[width=0.48\linewidth]{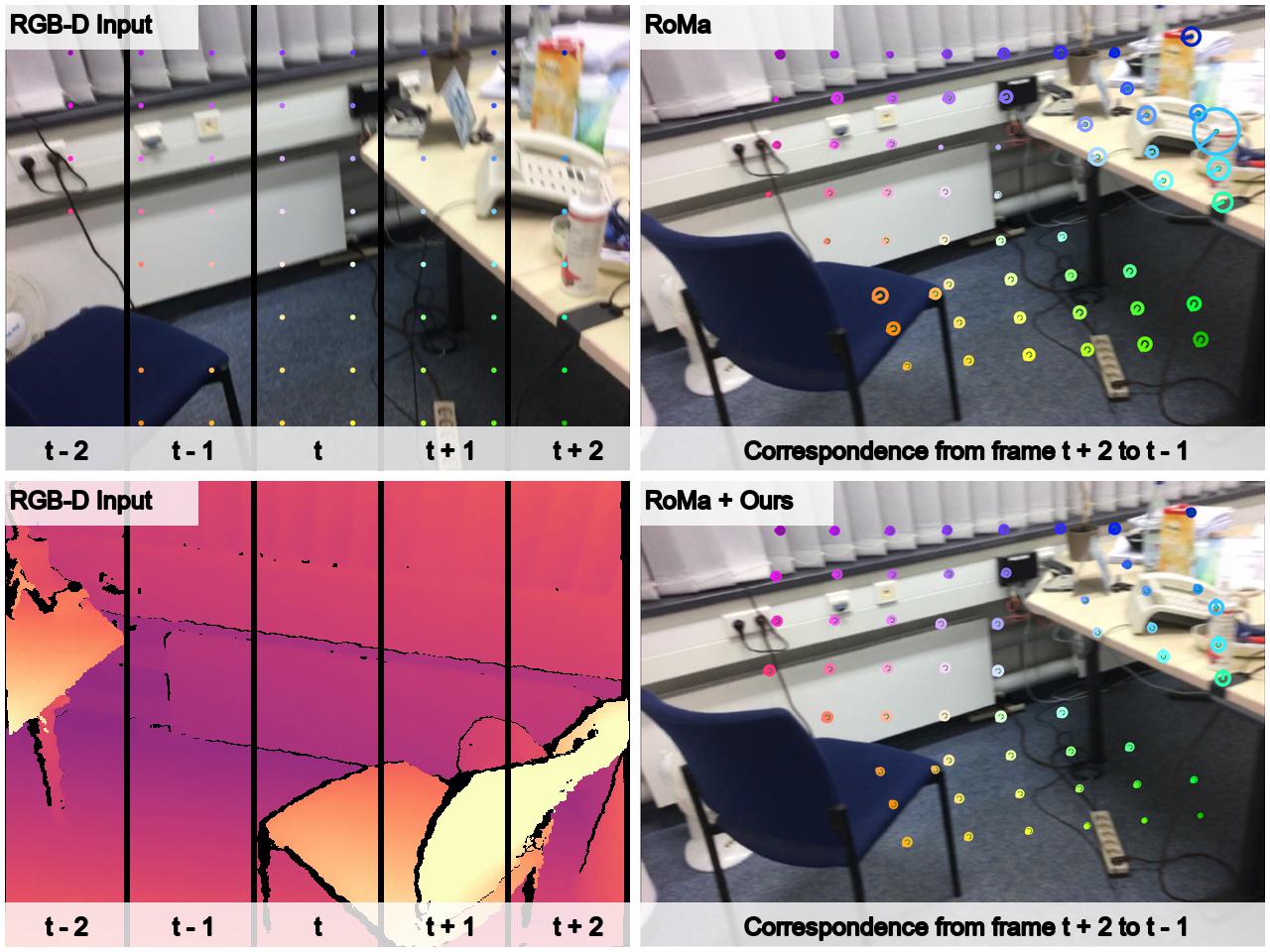}} \, 
    \subfloat[]{\includegraphics[width=0.48\linewidth]{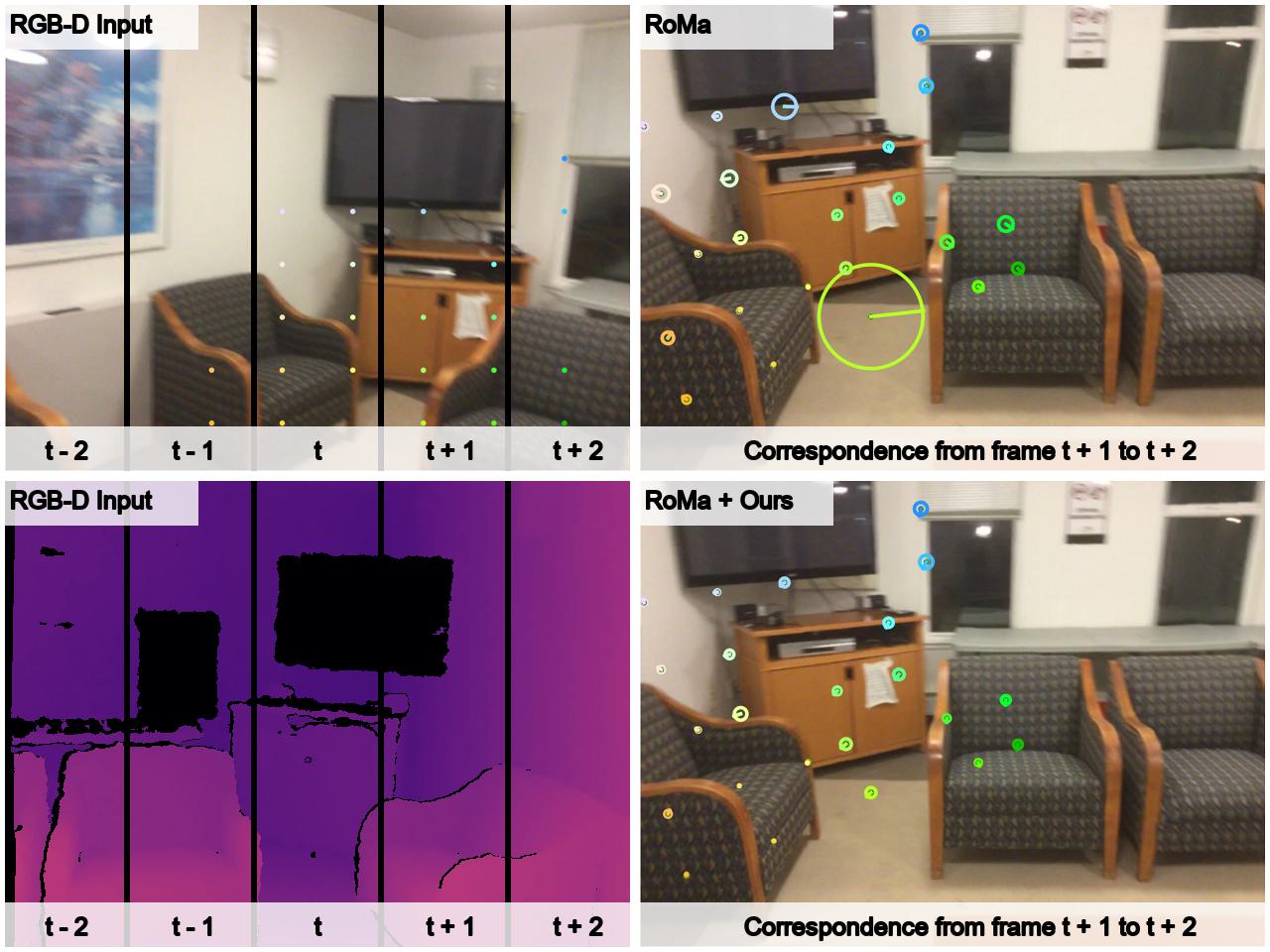}} \\
    \subfloat[]{\includegraphics[width=0.48\linewidth]{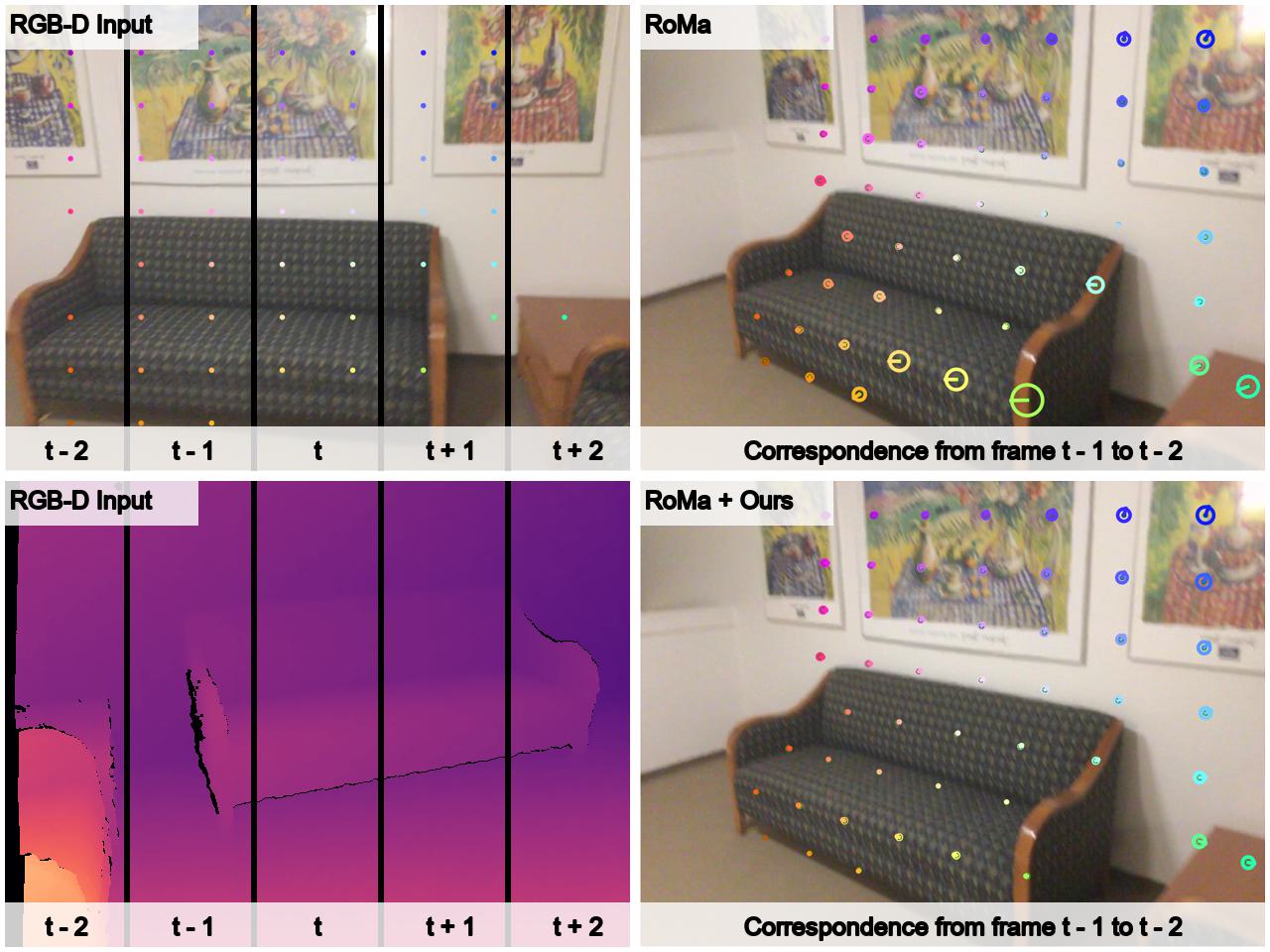}} \, 
    \subfloat[]{\includegraphics[width=0.48\linewidth]{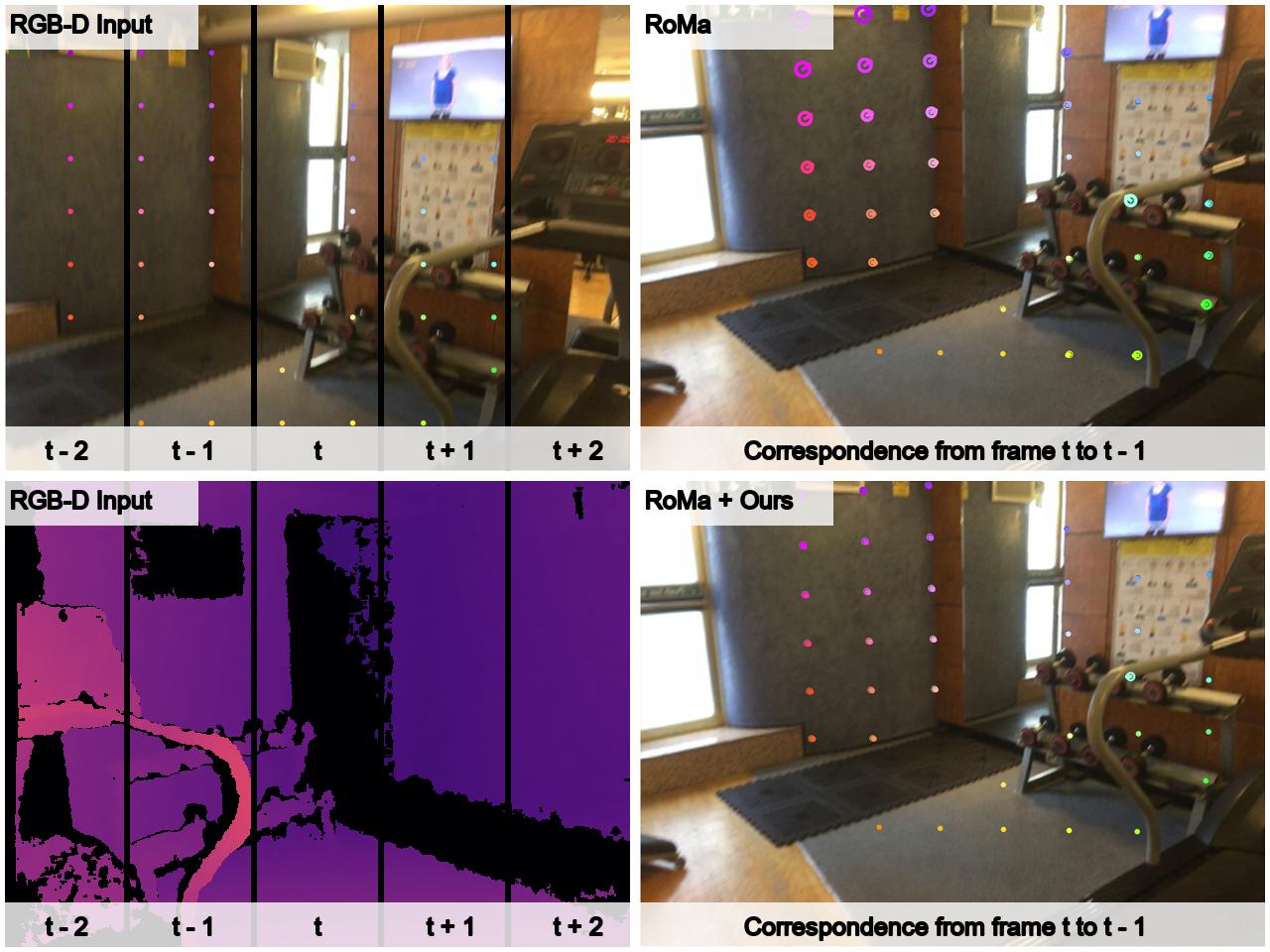}} \\
    \vspace{-2mm}
    \caption{\small 
    \textbf{Self-supervised Correspondence Estimation.} 
    Qualitative comparison with input SoTA correspondence estimator RoMa~\cite{edstedt2023roma}.
    }
    \label{fig:supp_corr}
\end{figure*}

We additionally visualize the self-supervised depth estimation, consistent depth estimation, and self-supervised correspondence estimation in \cref{fig:supp_sdepth}, \cref{fig:supp_cdepth}, and \cref{fig:supp_corr}, respectively.
Please see the captions for detailed analysis.

\Paragraph{Self-supervised Depth Estimation.} 
In \cref{fig:supp_sdepth}, the Radiance Field (RF) rendered dense depthmap has poor visual quality, especially around object boundaries.
This is expected, as the RF essentially applies triangulation, relying on consistent multi-view depth and correspondence estimation.
Yet, both monocular depthmap and correspondence estimation exhibit lower quality around borders.
When the inconsistency is significant, the triangulated results become underdefined, leading to noisy artifacts.

\Paragraph{Self-supervised Consistent Depth.} 
When generating \cref{fig:supp_cdepth}, we produce groundtruth correspondence via groundtruth pose and depthmap. 
Using this correspondence, we composite multi-frame depthmaps with groundtruth correspondence 
depthmaps from various frames and subsequently backproject them to the root frame. 
Each column corresponds to sampling from a different frame's depthmap, and the final image is composed over all input frames.






\clearpage

{
    \small
    \bibliographystyle{splncs04}
    \bibliography{main}
}

\end{document}